\documentclass[lettersize,journal]{IEEEtran}
\usepackage{amsmath,amsfonts}
\usepackage{algorithmic}
\usepackage{algorithm}
\usepackage{array}
\usepackage{textcomp}
\usepackage{stfloats}
\usepackage{url}
\usepackage{array,multirow,graphicx}
\usepackage{subcaption}
\usepackage{float}
\usepackage{gensymb}
\usepackage{caption}
\usepackage{placeins}
\usepackage{multirow}
\usepackage{flushend}
\usepackage{verbatim}
\usepackage{hyperref}
\usepackage{graphicx}
\usepackage{cite}
\graphicspath{{./Fig/}}	
\begin{document}

\title{Low-Contact Grasping of Soft Tissue with Complex Geometry using a Vortex Gripper}

\author{Roman Mykhailyshyn,~\IEEEmembership{Member,~IEEE} and~Ann~Majewicz Fey,~\IEEEmembership{Member,~IEEE}
% <-this % stops a space

\thanks{Roman Mykhailyshyn is with the Department of Systems Innovation, Osaka University, Toyonaka, Osaka 560-8531, Japan, and with transfer to the Industrial Cyber-Physical Systems Research Center at National Institute of Advanced Industrial Science and Technology (AIST), Japan, and on leave from the Texas Robotics and Walker Department of Mechanical Engineering, The University of Texas at Austin, Austin, TX 78712, USA (Corresponding author e-mail: mykhailyshyn.roman.es@osaka-u.ac.jp).}%
\thanks{Ann Majewicz Fey is with the Department of Mechanical Engineering, The University of Texas at Austin, Austin, TX 78712, USA, and is also with the Department of Surgery, UT Southwestern Medical Center, Dallas, TX 75390, USA. e-mail: Ann.MajewiczFey@utexas.edu}} % stops a space

%\markboth{Journal of ,~Vol.~, No.~, August~2024}%


\maketitle

\begin{abstract}
Soft tissue manipulation is an integral aspect of most surgical procedures; however, the vast majority of surgical graspers used today are made of hard materials, such as metals or hard plastics. Furthermore, these graspers predominately function by pinching tissue between two hard objects as a method for tissue manipulation. As such, the potential to apply too much force during contact, and thus damage tissue, is inherently high. As an alternative approach, gaspers developed using a pneumatic vortex could potentially levitate soft tissue, enabling manipulation with low or even no contact force. In this paper, we present the design and well as a full factorial study of the force characteristics of the vortex gripper grasping soft surfaces with four common shapes, with convex and concave curvature, and ranging over 10 different radii of curvature, for a total of 40 unique surfaces. By changing the parameters of the nozzle elements in the design of the gripper, it was possible to investigate the influence of the mass flow parameters of the vortex gripper on the lifting force for all of these different soft surfaces. An $\pmb{ex}$ $\pmb{vivo}$ experiment was conducted on grasping biological tissues and soft balls of various shapes to show the advantages and disadvantages of the proposed technology. The obtained results allowed us to find limitations in the use of vortex technology and the following stages of its improvement for medical use. 
\end{abstract}

%Surgical procedures frequently involve the delicate manipulation of soft tissues, which is the most common in trauma. Existing medical robotic grasping systems for soft tissues often cause tissue damage due to their substantial contact area. To address these issues in soft tissue manipulation, the use of vortex technology is proposed. Vortex technology enables the secure grasping of materials from a distance and levitation, distinguishing it from traditional suction cups by avoiding the depressurization effect when dealing with complex surfaces. In this paper, we present a comprehensive factorial study of the force characteristics exhibited by the vortex gripper when interacting with soft surfaces of four common shapes, each with varying radii of curvature. By adjusting the parameters of the nozzle elements in the gripper's design, we were able to explore how the mass flow parameters of the vortex gripper influence the lifting force on different soft surface shapes. To illustrate the advantages and disadvantages of this proposed technology, we conducted $\pmb{ex}$ $\pmb{vivo}$ experiments involving the manipulation of biological tissues and soft balls with various shapes. The results we obtained shed light on the limitations of using vortex technology and will guide us in refining and enhancing its applicability for medical purposes in subsequent stages of development.


% \fi

%=================================================
\section{Introduction}
%=================================================

%\IEEEPARstart{A}{pplied} contact force and its measurement during surgery is still a developing area of research~\cite{golahmadi2021tool}, but it is clear that experienced surgeons usually apply less force to the tissue. Minimizing force applied to tissue is important during surgery as studies have demonstrated that high tissue grasping forces lead to the destruction of endothelial and smooth muscle cells \cite{famaey2010vivo}, increase the risk of irreversible tissue changes \cite{barrie2018vivo, huan2020soft}, and elevate the likelihood of adhesion formation after surgery \cite{ellis2007postoperative}. Post-surgical adhesions can result in devastating long-term consequences for patients, including pain, bowel obstruction, and infertility ~\cite{diamond2001clinical}. Consequently, there has been significant interest in integrating force-sensing capabilities into surgical grippers ~\cite{kim2015force,wottawa2016evaluating}. Animal studies have demonstrated that incorporating a tactile haptic feedback device, intended to convey grasping forces to the user, can substantially reduce both grasping force and tissue damage~\cite{wottawa2016evaluating}. Despite its promise, this technique relies on custom sensors that are not yet widely available commercially. An alternative approach could involve designing novel tissue grippers that cannot exert such high, damaging forces.

\IEEEPARstart{A}{pplied} contact force and its measurement during surgery is still an evolving area of research~\cite{golahmadi2021tool}. Nevertheless, it is evident that experienced surgeons typically apply less force to the tissue. Minimizing force applied to the tissue holds significant importance in surgery, as studies have shown that high tissue grasping forces can lead to the destruction of endothelial and smooth muscle cells~\cite{famaey2010vivo}, increase the risk of irreversible tissue changes \cite{barrie2018vivo, huan2020soft}, and elevate the likelihood of adhesion formation after surgery \cite{ellis2007postoperative}. Post-surgical adhesions can result in severe long-term consequences for patients, including pain, bowel obstruction, and infertility~\cite{diamond2001clinical}. As a result, there has been substantial interest in integrating force-sensing capabilities into surgical grippers~\cite{kim2015force,wottawa2016evaluating, kinnicutt2024soft}. 

We propose a novel approach to grasping soft tissues, aimed at minimizing contact while ensuring sufficient force to hold soft objects of various shapes. A preliminary study~\cite{mykhailyshyn2024low} highlighted the critical influence of surface radius in gripping force, leading to key insights and necessary adjustments to the nozzle parameters.

\begin{figure}[t]
\centering
\vspace{1mm}
\includegraphics[width=0.92\linewidth,clip ,trim=0pt 20pt 0pt 10pt]{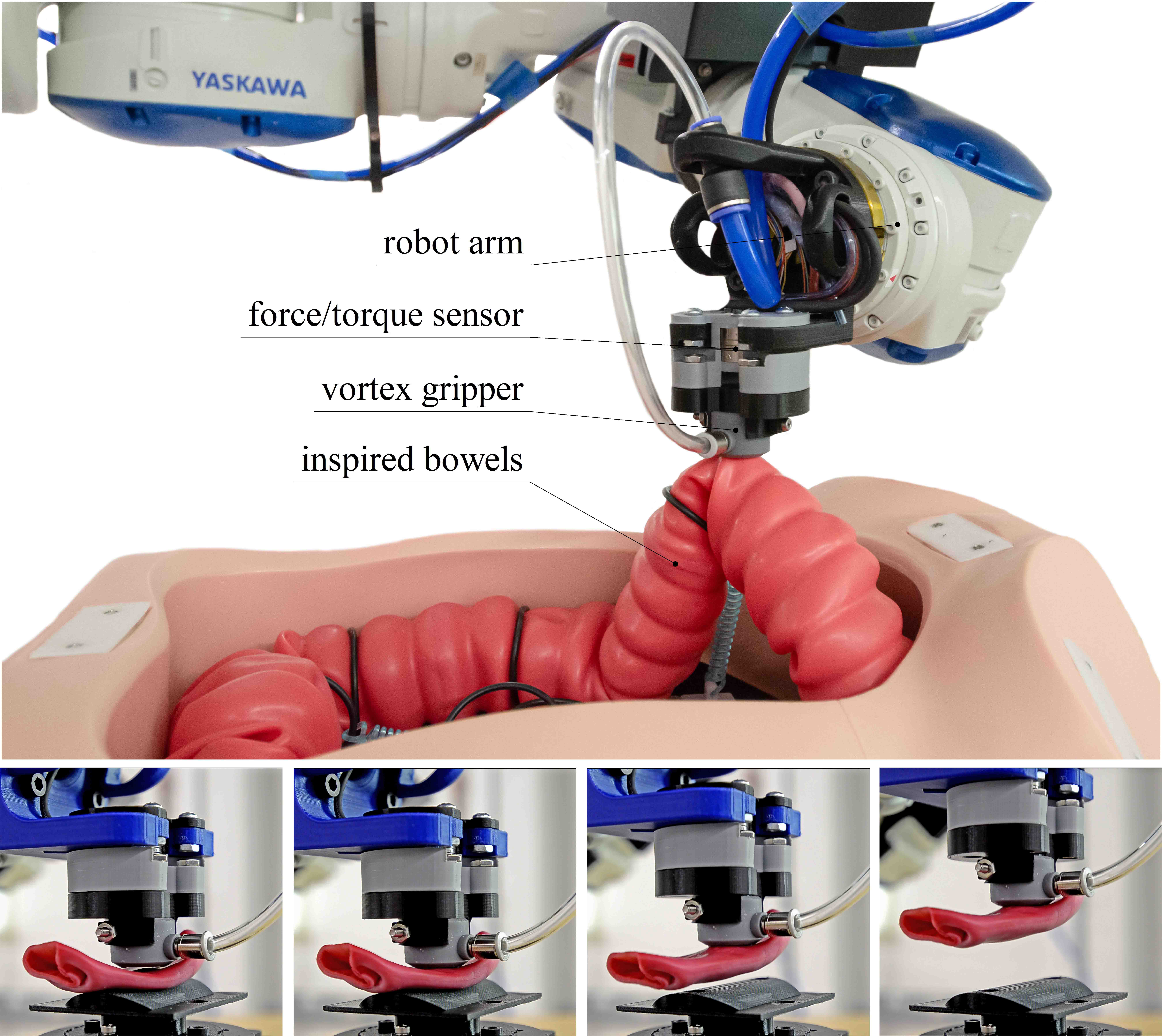}
 \caption{Practice low-contact grasping techniques on the bio-inspired intestine using the Colonoscope Training Simulator (Kyoto Kagaku).}
\label{fig-1}
\end{figure}

In this study, we used vortex gripper technology, which effectively addresses the challenges associated with unnecessary soft tissue contact. Using additive SLA 3D printing technologies for fabrication, we have reduced the size and weight of the gripper while overcoming issues related to printing small nozzles. Our results demonstrate that the vortex gripper can successfully grasp soft objects of various shapes (Fig.~\ref{fig-1}), showing both sufficient robustness and enhanced gripping force. Specifically, it achieves a lifting force of up to 4.8 N at 400 kPa supply pressure, which is more than 48 times the weight of the gripper.

This paper highlights the following key contributions: 
\begin{itemize}
  \item A step by step methodology for fabricating and calculating the parameters of 3D-printed tangential nozzles in a vortex gripper.
  \item Comprehensive characterization of gripping performance (changing the nozzle diameter and supply pressure) with soft objects of different shapes and radii.
  \item Application of machine learning to predict the effects of gripper and object parameters on the force characteristics of the gripper using the generated dataset.
  \item Evaluation of gripper performance with various soft objects, bio-inspired tissues, and $ex$ $vivo$ specimens.
\end{itemize}

%This paper highlights the following contributions: (1) a simple design of a vortex gripper consisting of tangentially directed nozzles to a cylindrical cavity; (2) three different diameters of nozzle elements; and (3) characterization of grasping performance using both soft objects of different shapes and radii, and a set of test objects with different balls geometries, as well as $ex$ $vivo$ objects.

%-------------------------------------------------
\subsection{Motivation for Jet Grasping Technology}
%-------------------------------------------------

Animal studies have demonstrated that incorporating a tactile haptic feedback device intended to convey grasping forces to the user can significantly reduce both grasping force and tissue damage~\cite{wottawa2016evaluating}. Despite its promising outcomes, this technique relies on custom sensors that are not yet widely available commercially. An alternative approach could involve designing novel tissue grippers that are incapable of exerting such high, damaging forces.

%\IEEEPARstart{T}{he} use of robotic systems during medical applications has many advantages. In particular, a reduction in conversion rates was demonstrated over a 6-year period associated with the use of robotic platforms in surgery in the United States \cite{abd2022trends}. Most minimally invasive surgery is performed using robotic systems with rigid links and end effectors \cite{haidegger2022robot}. However new continuum robots and effectors were developed to ensure less trauma and accessibility in minimally invasive surgery \cite{burgner2015continuum, kwok2022soft}. They significantly minimize the robot's contact with the body, but the process of grasping and manipulating objects in the body cavity can still lead to undesirable consequences. The further development of systems of tactile sensations for medical robotic systems \cite{okamura2019haptics}, allowed to control of the movements of continuous robots more precisely and minimized the contact of robot links with tissues and organs, which made it possible to reduce the manifestations of adhesion. However, grasping systems continue to interact with tissues with significant force during surgery. It has been proven that high tissue grasping forces lead to the destruction of endothelial and smooth muscle cells \cite{famaey2010vivo}, increasing the risk of irreversible tissue change \cite{barrie2018vivo, huan2020soft}, and increasing the appearance of adhesions after surgery \cite{ellis2007postoperative}. 

Currently, only a few different types of grippers are used in surgical applications: mechanical (with different types of drive) \cite{george2022design}, vacuum (suction cups of different designs) \cite{vonck2012grasping}. However, it is mechanical grippers that are used most often, which differ in the type of drive: compliant mechanisms \cite{10.1115/1.2056561}, elastic actuator \cite{gerboni2016novel} and others. During the grasping of tissues, mechanical grippers can easily damage them, the authors \cite{ullrich2015magnetically} demonstrate the detachment of the liver remaining on the micro-gripper.

Therefore, soft mechanical grippers that can minimize tissue damage are becoming widely popular: for object coverage \cite{dunn2022thermomagnetic}, for thread and needle grasping \cite{low2014customizable}, and three-finger grippers for tissue grasping \cite{rateni2015design, liu2020musha}. Soft grippers reduce the force of excess clamping but do not allow it to be effectively controlled. Because of this, researchers began integrating clamp force sensors into surgical grippers \cite{kim2015force}. 

The authors of the article \cite{abiri2019multi} investigated that the grasping force of a person and a classic mechanical gripper (in the robot DaVinci) is on average 3 times less. Such results became the basis for conducting research in the direction of minimization of tissue damage while proposing methods of improvement: controlling the clamping force using touch sensors \cite{barrie2018vivo} and combining this method with soft covers \cite{huan2020soft}. However, the grasping force in the case of clamping the tissue between the gripper fingers will be greater than the minimum required for holding. This proves that in order to minimize the force of contact and tissue damage, it is necessary to develop other approaches and methods. Suction cups \cite{patronik2009miniature, bamotra2018fabrication} are also used for suction to the surface and can move along it with the help of a cable drive. Such grasping systems produce less tissue damage, but still have disadvantages: the use of suction cups provides a large contact area, which can lead to surgical adhesion; the suction cup is limited in its ability to grasp other surfaces in the body cavity and can easily clog during surgery.

\begin{figure}[t]
\centering
\subfloat[]{\includegraphics[width=0.46\linewidth,clip ,trim=0pt 0pt 0pt 0pt]{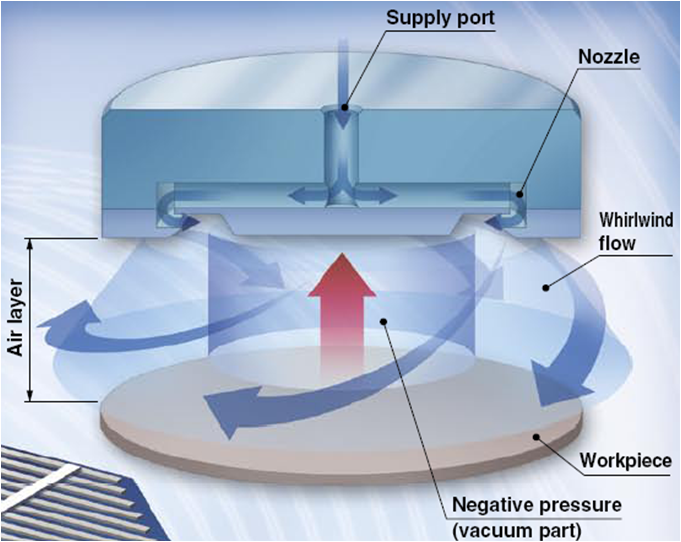}%
\label{fig2_1}}
\hfil
\subfloat[]{\includegraphics[width=0.5\linewidth,clip ,trim=0pt 0pt 0pt 0pt]{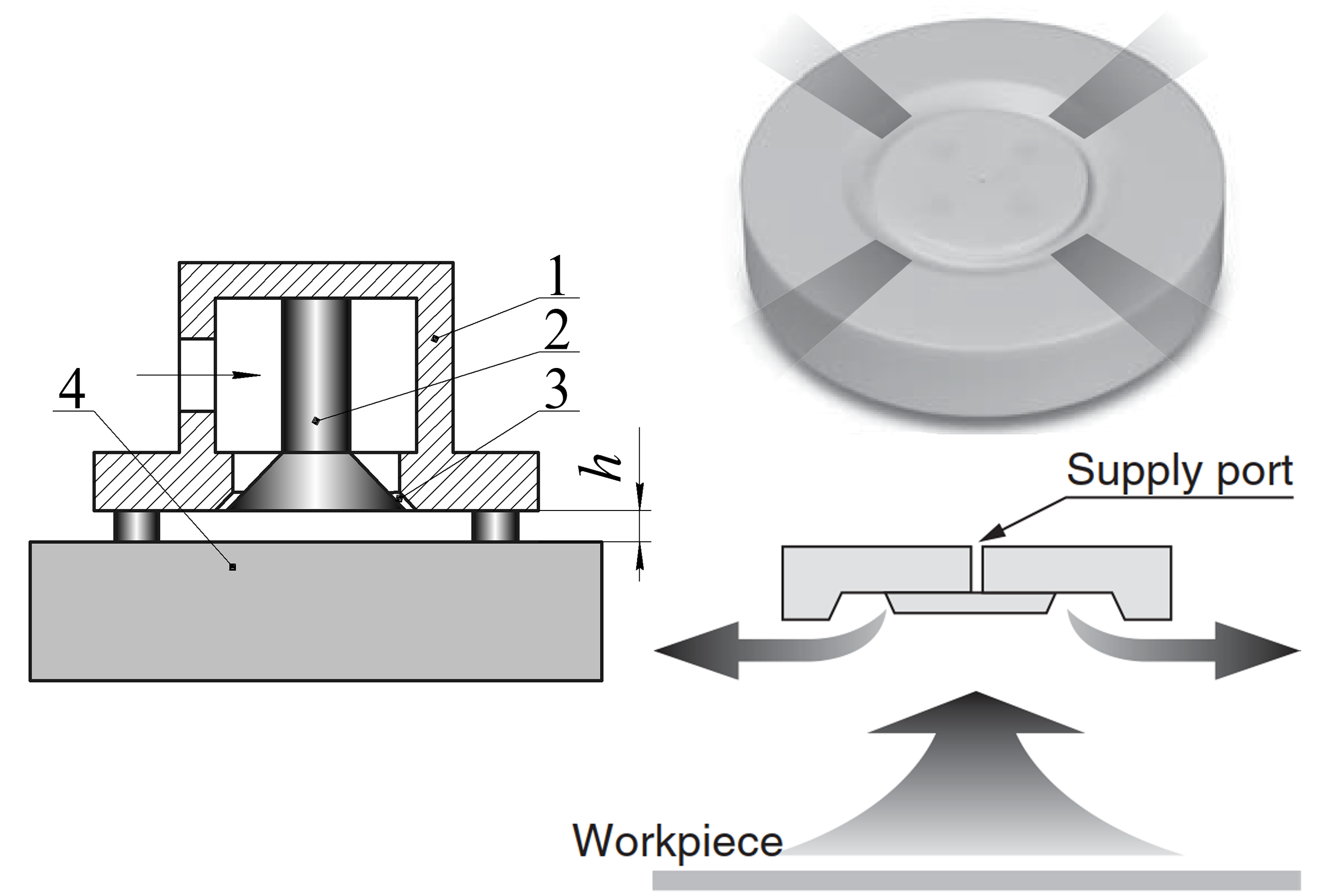}%
\label{fig2_2}}
\hfil
 \caption{The two most common jet gripping technologies~\cite{mykhailyshyn2022systematic}: (a) - Vortex grasping process; (b) - Bernoulli grasping process (1 - body, 2 - conical insert, 3 - circular nozzle, 4 - object).}
\label{fig2}
\end{figure}

However, other pneumatic jet technologies may be used for medical grasping applications. Among all pneumatic jet gripping devices \cite{mykhailyshyn2022systematic, li2015experimental}, vortex (Fig.~\ref{fig2}(a)) and Bernoulli grippers (Fig.~\ref{fig2}(b)) are distinguished by their unique properties. They are characterized by high reliability, long service life, and low manufacturing cost. An important characteristic is the possibility of contact and non-contact grasping of manipulation objects, regardless of their material, mechanical characteristics, structure of the surface layer and temperature.

Most often, the use jet gripping devices can be found when grasping fragile (solar panels \cite{zhao2019gap}, silicon plates \cite{wang2019effect, liu2018soft}, glass \cite{kim2015configuration}, etc.), flexible (packaging \cite{ozcelik2005examination}, textiles \cite{Mykhailyshyn2022gripper, mykhailyshyn2023fem, Mykhailyshyn2024toward}, etc.) n.) and objects with a complex shape (fruit \cite{liu2020design}, leather \cite{dini2009grasping}, etc.) and holes (electronic boards \cite{xin2014development}). Among the new directions of application of Bernoulli grasping during laparoscopy \cite{erturk2019design, trommelen2011development} as a tool, however, they have not gained much popularity.

With the development of mobile robotics, among all the tasks facing developers, it is possible to single out the provision of robot movement on various surfaces \cite{rubio2019review}. One of the solutions to this problem was the use of jet gripping devices as elements of holding the mobile robot on the surface \cite{journee2011investigation}. Among the existing options from the point of view of minimal contact with the surface is the use of pneumatic jet gripping devices \cite{zhao2018experimental}. Unlike suction cups, jet grippers have a much smaller influence of the surface of the grasping object on the force of attraction.

Among all jet gripping devices, vortex gripping devices have the least effect of the surface roughness and shape of the object on the holding force \cite{mykhailyshyn2024low}. Unlike Bernoulli grippers in which the air flow is directed at the object, in vortex grippers the nozzles are located tangentially to the gripper cavity and due to the centrifugal force created in the vortex, the air flow practically does not interact with the surface of the object . Due to this effect, these grippers are often used when gripping electrical boards on which various components are located  and when it is necessary to attach to objects with great roughness \cite{shi2020vacuum}.

%=================================================
\section{Methodology and Material}
%=================================================

%-------------------------------------------------
\subsection{Design Vortex Gripper}
%-------------------------------------------------

\begin{figure}[t]
\centering
\includegraphics[width=0.95\linewidth,clip ,trim=0pt 0pt 0pt 40pt]{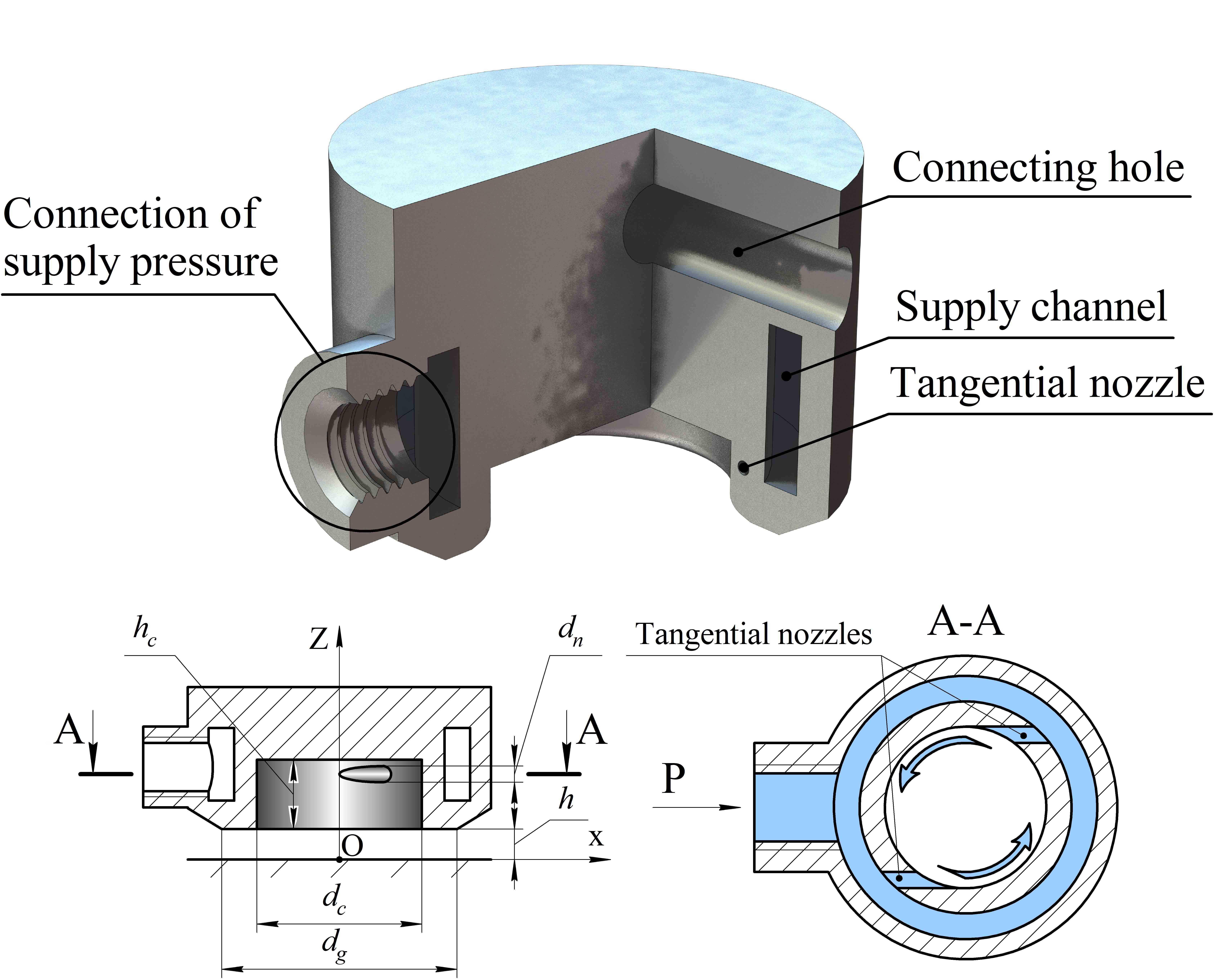}
 \caption{The proposed design of the gripper (with parameters indicated: $d_g$ - gripper diameter, $d_n$ - nozzle diameter, $d_c$ - cavity diameter, $h_c$ - cavity height, $h$ - the gap between the gripper and the object).}
\label{fig-3}
\end{figure}

The vortex gripper has a well-studied design as it is used to grab fragile objects like silicon and solar panels in manufacturing. However, designing such a gripper for medical applications has many limitations. The biggest challenge is to design it small enough to be used, and with the ability to provide enough lifting force to manipulate soft tissue. We selected as an initial condition, the parameters of the bowls (common for manipulation in trauma), which have an average statistical minimum radius of 15 mm \cite{doi:10.1148/radiographics.20.2.g00mc15399}. Therefore, it was decided that the body of the gripper should not exceed 26 mm. For the operation of the vortex gripper, compressed air (or liquid) $P$ is supplied through the connector in the gripper body (Fig.~\ref{fig-3}). After that, the compressed air enters the supply channel (to ensure maximum force characteristics \cite{mykhailyshyn2022influence}), which is connected to two cylindrical nozzles (nozzle diameter $d_n$) tangentially directed to the cylindrical cavity of the gripper with diameter $d_c$ and height $h_c$. Compressed air exits through these nozzles, and due to the cylindrical shape of the gripper cavity and the direction of the nozzles, the air begins to swirl (Fig.~\ref{fig-3}). As a result of swirling, the air flow acquires angular velocity and centrifugal force, which pushes it through the gap $h$ between the object and the active surface of the vortex gripper. Thanks to this effect, we get a negative pressure on the surface of the object in zone $d_c$ opposite the cavity, and in the zone of the object opposite $d_g - d_c$ (the active surface of the gripper) we get a positive pressure. Since the area of the negative pressure zone is much larger than the area of the positive pressure zone, and the resulting negative pressure has a greater difference with atmospheric pressure than with positive pressure, lifting force will be generated. A simplified derivation of the equation calculating the lifting force for vortex grippers is presented in \cite{wang2019effect}:

\begin{table}[t]
\begin{center}
\caption{Parameters of vortex grippers (unit: mm).}
\label{tab1}
\begin{tabular}{| c | c | c | c | c |}
\hline
\multirow{2}{*}{Gripper} & Nozzle & Gripper & Cavity & Cavity \\
 & diameter $d_n$ & diameter $d_g$ & diameter $d_c$ & height $h_c$ \\
\hline
G$_1$ & 0.6 & \multirow{3}{*}{20} & \multirow{3}{*}{14} & \multirow{3}{*}{4}\\ \cline{1-2}
G$_2$ & 0.8 &  &  & \\ \cline{1-2}
G$_3$ & 1.0 &  &  & \\
\hline
\end{tabular}
\end{center}
\end{table}

\begin{equation}
\label{deqn_ex1}
F_l = \frac{1}{4} \rho \pi \omega^{2}  {\bigg( \frac{d_c}{2} \bigg) }^{4},
\end{equation}

\noindent where $\rho$ is the air density (kg/m$^3$), $\omega= 2 u_{\alpha} / d_c  $ is the air angular velocity (1/s), and $u_{\alpha}$ is the circumferential velocity (m/s).

It is obvious that the same air pressure is maintained in the supply channel as is supplied to the gripping device (Fig.~\ref{fig-3}). Since the gripper device can be operated at high pressures up to 600 kPa, it is important to ensure safety and prevent the gripper wall from breaking. Therefore, it is accepted that the walls bordering the channel should not be less than 2 mm, and the channel itself must provide sufficient mass flow and also have a width of 2 mm. Based on the existing limitation of the gripper body diameter of 26 mm and channel parameters, the maximum possible gripper cavity diameter of 14 mm was selected. Based on the recommendations given in article \cite{wang2019effect}, the height of the cavity $h_c$ was chosen to be 4 mm. At the same time, in order to ensure a wide range of lifting force and efficiency of the gripper, we suggest changing the diameter of the nozzles $d_n = 0.6$, 0.8, and 1 mm, which will allow us to have three grippers with different force parameters (Table~\ref{tab1}).

%\begin{figure}[t]
%\centering
%\subfloat[]{\includegraphics[width=0.68\linewidth,clip ,trim=0pt 0pt 0pt 0pt]{Fig/Fig4_2.png}%
%\label{fig4_1}}
%\hfil
%\subfloat[]{\includegraphics[width=0.55\linewidth,clip ,trim=0pt 0pt 0pt 0pt]{Fig/Fig4_3.png}%
%\label{fig4_2}}
%\hfil
%\subfloat[]{\includegraphics[width=0.34\linewidth,clip ,trim=0pt 0pt 0pt 0pt]{Fig/Fig5_3.pdf}%
%\label{fig4_3}}
%\hfil
% \caption{SLA 3D printing of nozzle elements of different diameters of the vortex gripper: (a) - effect of orientation on gray resin nozzle printing; (b) - reducing the diameter of the nozzle for gray and transparent resin at a 90-degree orientation; (c) - graph of nozzle diameter reduction.}
%\label{fig4}
%\end{figure}

%-------------------------------------------------
\subsection{Fabrication Vortex Gripper}
%-------------------------------------------------

%FFFFFFFFFFFFFFFFFFFFFFFFFFFFFFFFFFFFFFFFFFFFFF
\begin{figure}[t]
\newcommand{\mywidth}{1}
\centering
	\begin{subfigure}[b]{0.65\linewidth}
         \centering
         \includegraphics[width=\mywidth\linewidth, clip, trim=0pt 0pt 0pt 0pt]{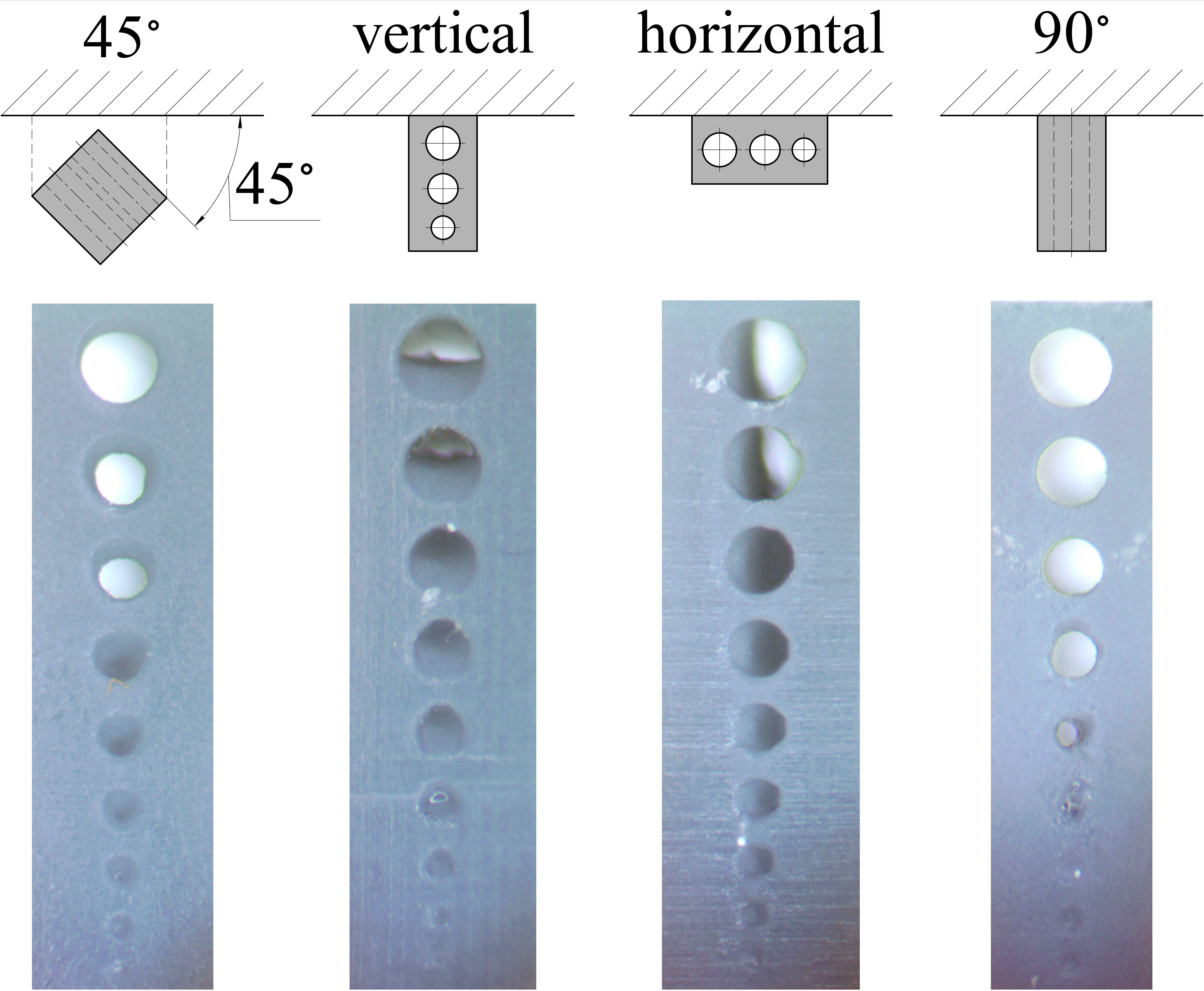}
        \caption{\label{fig4_1}}
	\end{subfigure}
	\\
	\begin{subfigure}[b]{0.7\linewidth}
         \centering
        \includegraphics[width=\mywidth\linewidth, clip, trim=0pt 0pt 0pt 0pt]{Fig/Fig4_3.jpg}
         \caption{ \label{fig4_2}}
	\end{subfigure}\\
 \caption{SLA 3D printing of nozzle elements of different diameters of the vortex gripper: (a) - effect of orientation on gray resin nozzle printing; (b) - reducing the diameter of the nozzle for gray and transparent resin at a 90-degree orientation. \label{fig4}}
\end{figure}
%FFFFFFFFFFFFFFFFFFFFFFFFFFFFFFFFFFFFFFFFFFFFFF

SLA 3D printing (Formlabs Form 3+) was chosen for the fabrication of vortex gripping devices, thanks to which it is possible to achieve the highest accuracy (print height 25 $\mu$m) of the reproduction of small parts of the prototype. However, during the printing of the vortex gripping device, it turned out that the shape and diameter of the nozzle elements change greatly for $d_n \leq 1$ mm. The first parameter that affects the change in the shape of the nozzle is the orientation of the nozzle during printing. We use resin materials for printing: Grey V4 \cite{GreyV4} and Transparent V4 \cite{ClearV4}. After printing the nozzle elements from Grey V4 (from 0.3 to 1 mm in steps of 0.1 mm) with horizontal, vertical, 45$^{\circ}$, and 90$^{\circ}$ orientation to the platform (Fig.~\ref{fig4_1}), it can be seen that due to the density of the resin and the small radius, there is a capillary effect and for $d_n \leq 0.5$ mm resin hardens in the nozzle. 

\begin{figure}[t]
\centering
\includegraphics[width=0.86\linewidth,clip ,trim=0pt 0pt 0pt 0pt]{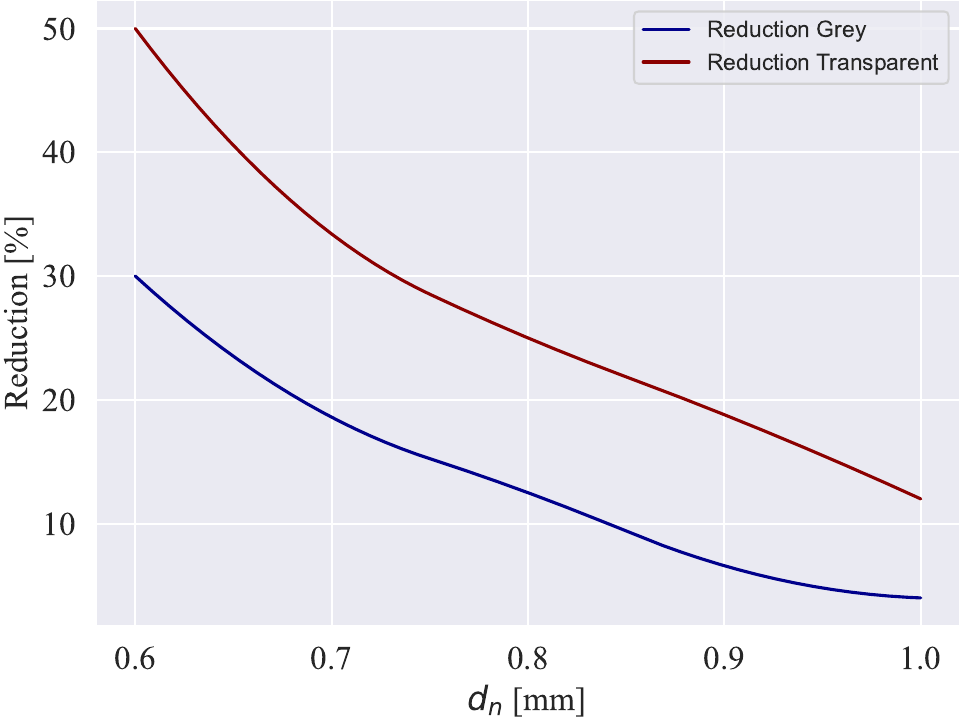}
 \caption{Reduction of the diameter of the nozzle relative to the original size that was specified during printing.}
\label{fig4_3}
\end{figure}

For $d_n \geq 0.6$ mm (Fig.~\ref{fig4_1}), excessive hardening of the resin occurs in places where it flows into the tank under the influence of gravity forces. This leads to the deformation of the shape of the nozzle for horizontal, vertical, and 45$^{\circ}$ orientation. However, the deformation of the nozzle does not occur when it is oriented at 90$^{\circ}$ to the platform, as the resin flows evenly over the entire cylindrical surface of the nozzle. Therefore, when 3D printing a nozzle with an orientation of 90$^{\circ}$, only the diameter of the nozzle is reduced (Fig.~\ref{fig4_2}). 

From Fig.~\ref{fig4_2}, it is obvious that the reduction of the ratio of the nozzle diameter for the Transparent V4 material is much greater than for the Grey V4, which is related to the parasitic illumination by ultraviolet rays of the previous layers through the transparent material during 3D printing. Therefore, for further prototyping, the Grey V4 resin will be used for more accurate printing. The FDM 3D printing method of nozzle elements of Bernoulli grippers and reduction effects has already been studied by the authors \cite{mykhailyshyn2022three}. However, for SLA 3D printing, it is enough to determine the nozzle diameter reduction coefficient ($k_d$) for the material (Fig.~\ref{fig4_3}), which will allow determining the necessary changes in the predefined diameter 3D model of the nozzle ($d_n^{\text{CAD}}$):

\begin{equation}
\label{deqn_ex2}
d_n = d_n^{\text{CAD}} (1 - k_d).
\end{equation}

Because the value of the coefficient of diameter reduction cannot be less than zero, we use the exponential regression equation obtained from experimental data, which is equal to Gray material (coefficient of determination $R^2 = 0.9957$ in the range from $d_n = 0.4$ to 1.2 mm):

\begin{equation}
\label{deqn_ex3}
k_d  = 6.5028 e^{-5.066 d_n^{\text{CAD}}},
\end{equation}

\noindent to Transparent material (coefficient of determination $R^2 = 0.9946$ in the range from $d_n = 0.4$ to 1.2 mm):

\begin{equation}
\label{deqn_ex4}
k_d  = 3.856 e^{-3.429 d_n^{\text{CAD}}}.
\end{equation}

Since for prototyping we will use Grey V4 material to provide a nozzle with a radius of $d_n = 0.6$ mm, from (2) and (3) we can determine the required nozzle diameter of the 3D model $d_n^{\text{CAD}} = 0.722$ mm, for $d_n = 0.8$ mm we will get $d_n^{\text{CAD}} = 0.869$ mm, for $d_n = 1.0$ mm we will get $d_n^{\text{CAD}} = 1.036$ mm.

%For Grey V4 material $d_n = 0.6$ mm ($d_n^{\text{CAD}} = 0.722$), $d_n = 0.8$ mm ($d_n^{\text{CAD}} = 0.869$), $d_n = 1.0$ mm ($d_n^{\text{CAD}} = 1.036$).

%\begin{split}
%d_r  = 34.167 \big( d_n^{\text{CAD}} \big) ^4 -111 \big( d_n^{\text{CAD}} \big) ^3+ \\
%+ 134.96 \big( d_n^{\text{CAD}} \big) ^2 - 73.37 d_n^{\text{CAD}} + 15.285.
%\end{split}

%-------------------------------------------------
\subsection{Description of the Experimental Setup}
%-------------------------------------------------

\begin{figure}[t]
\centering
\subfloat[]{\includegraphics[width=0.86\linewidth,clip ,trim=0pt 0pt 0pt 0pt]{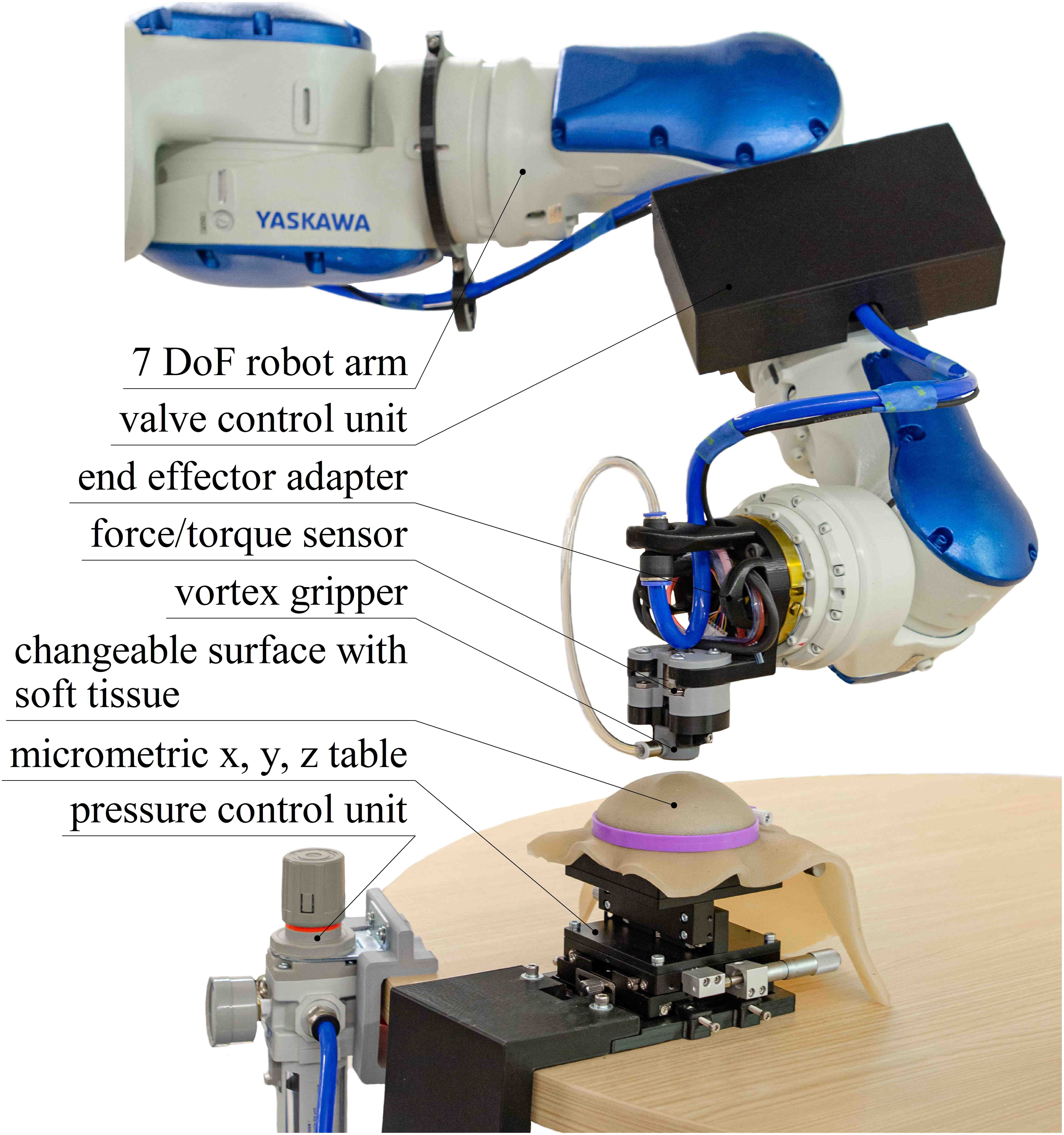}%
\label{fig7_1}}
\hfil
\subfloat[]{\includegraphics[width=0.92\linewidth,clip ,trim=0pt 0pt 0pt 0pt]{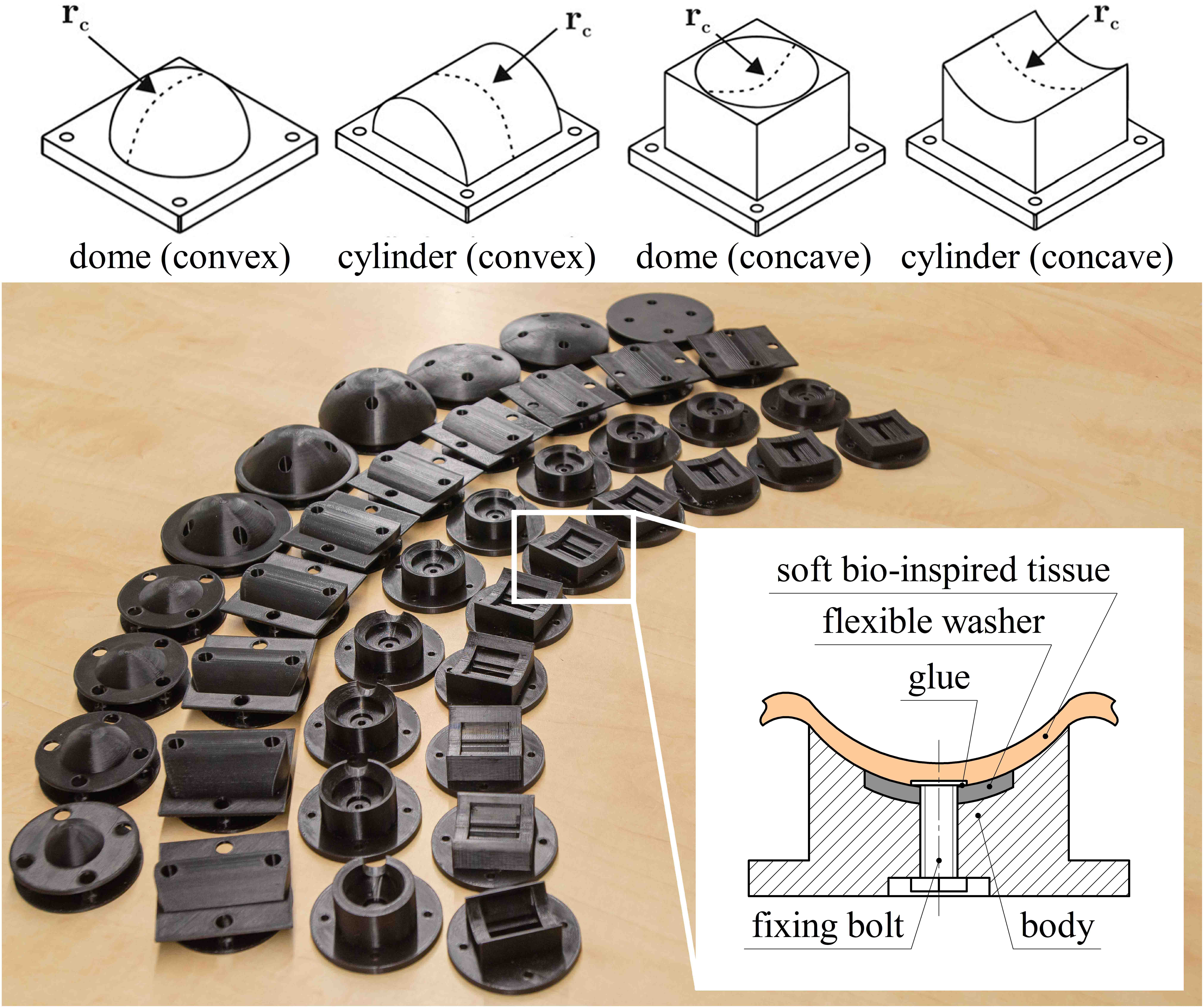}%
\label{fig7_2}}
\hfil
 \caption{Experimental setup for studying the force characteristics of a vortex gripper: (a) -  general view of the installation; (b) – mechanism of changing the shape of the object.}
\label{fig-7}
\end{figure}

In order to determine the influence of the parameters of the shape of soft objects and tissues on the lifting force of vortex grippers, it is necessary to determine with which types and sizes of surfaces will be interacted grippers during surgical applications. For example, let's take the most frequent operation for trauma surgery, which is lifting and manipulating the bowels. The radius of the bowels averages from 15 mm for the smallest intestine and less than 45 mm for the largest cecum \cite{doi:10.1148/radiographics.20.2.g00mc15399}. Depending on the filling of the bowels, their position in space, and interaction with other organs and tissues, in general, they can have 4 different types of surface shape: dome (convex), cylinder (convex), dome (concave), and cylinder (concave). In order to obtain a more general characteristic of vortex grippers for grasping soft tissues in medical applications, it is suggested to carry out studies for 4 types of surfaces and when changing the radius of the surfaces $r_c$ = 15, 20, 25, 30, 35, 40, 45 mm (which is the range for bowls) and 50, 75, 100 mm and flat (which is a range typical for other tissues).

To determine the force characteristics of the developed vortex gripping devices with soft objects of different surfaces, an experimental setup was developed (Fig.~\ref{fig-7}). The experimental setup (Fig.~\ref{fig7_1}) consists of a 7 DoF Yaskawa SDA10F robot arm to which a valve control unit and an end effector adapter are attached. A vortex gripper is attached to the end effector adapter via an ATI Nano17 force/torque sensor \cite{ati} with an accuracy of ±0.25$\%$. A compressed air pressure control unit and a micrometric $x, y, z$ table are attached to the table. Changeable surfaces are attached to the micrometric $x, y, z$ table on which soft bio-inspired tissue is fixed (Fig.~\ref{fig7_2}).

%The experimental setup (Fig.~\ref{fig7_1}) consists of a 7 DoF Yaskawa SDA10F robot arm, to which a gripping device \cite{} is attached via an adapter, a computer with ROS and ATI ----- \cite{} software. The robot controller and ATI ------ force measuring device with an accuracy of ±---- percent (Serial No. ------) are connected to the computer.

To imitate soft tissue, bio-inspired skin was used, which was fabricated from AB-type two platinum-catalyzed silicones. The two different layers, skin, and fat, were cured on top of each other at room temperature after 45-minute intervals. The layers were made using the measurements from the "Smooth On" instructions. The base of the skin pad is a double layer of powermesh that allows 4-way stretch and tear resistance. The skin layer uses Ecoflex-0030 \cite{SmoothOn_bng} in a 1A-1B ratio by volume and is poured in 3 separate batches to allow the powermesh to get fully soaked in the silicone mixture. Next, the fat layer uses Ecoflex GEL \cite{SmoothOn_gel} in a 1A:1B ratio by volume and is poured in 2 separate batches that cure into a super soft and tacky silicone layer.

%-------------------------------------------------
\subsection{Experiment Protocol}
%-------------------------------------------------

To determine the lifting force of the vortex gripping device, the appropriate value of compressed air pressure (100, 200, 300, or 400 kPa) is set on the pressure control unit. After that, calibration of the force/torque sensor is carried out with the gripper turned on (using the valve control unit) at a distance of more than 100 mm from the object (to avoid the interaction of air flows with the object) and after calibration, it is turned off. The lifting force of the vortex gripper will appear for different surfaces at different heights from them and it is difficult to determine analytically for soft tissues. It was decided to determine the force parameters of the gripper starting from the contact with the object. In our case, it is the $F_z$ data of the force/torque sensor that will determine the lifting force of the gripper. Therefore, with the help of the robot arm, the vortex gripper is brought vertically to the center of the studied surface until the moment when the $F_z$ value of the force/torque sensor does not reach -~2~Newtons. This means that the vortex gripper collided with the object and deformed the soft-bio-inspired tissue. 

After that, the valve that supplies compressed air to the vortex gripper is turned on. In response to the compressed air supply pressure, the $F_z$ indicator of the force/torque sensor decreases, which is because the gap between the object and the gripper is closed. After that, the vortex gripping device moves vertically upwards (velocity 0.01 m/s, acceleration 0.01 m/s$^2$) with the start of data collection of force characteristics ($F_z$) from the moment when $F_z = 0$ and until the gripper rises to a distance of more than 100 mm from the object. 

The experiment is repeated 10 times in a circle for the same surface, gripper, and pressure. This makes it possible to determine the average value of the lifting force of the vortex gripper for the given parameters of the system. Conducted experiments with changing the gripping device (3 variants), the supply pressure of the gripper (4 variants), 4 surface types with 10 different surface radii and flat (41 variants), and conducting 10 experiments with each variant allowed us to create a data set of 4,920 experiments.

%=================================================
\section{Results and Discusion}
%=================================================

\begin{figure}[t]
\centering
\includegraphics[width=0.8\linewidth,clip ,trim=10pt 10pt 20pt 20pt]{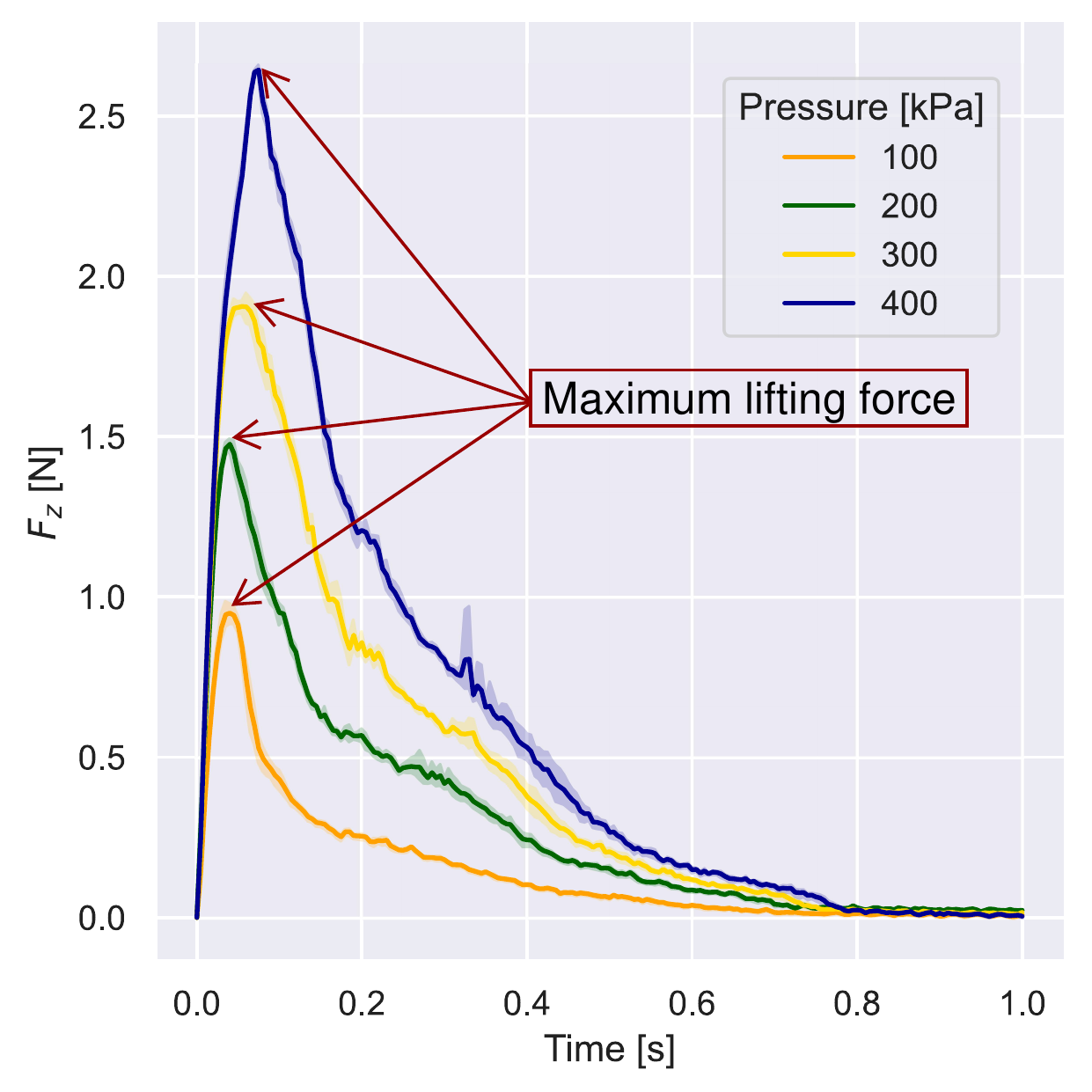}
 \caption{Average distribution lifting force $F_z$ during the vertical movement of the gripper for different supply pressure (Gripper 1, $d_n = 0.6$ mm, Flat surface).}
\label{fig-8}
\end{figure}

\begin{figure}[t]
\centering
\includegraphics[width=0.8\linewidth,clip ,trim=0pt 5pt 10pt 10pt]{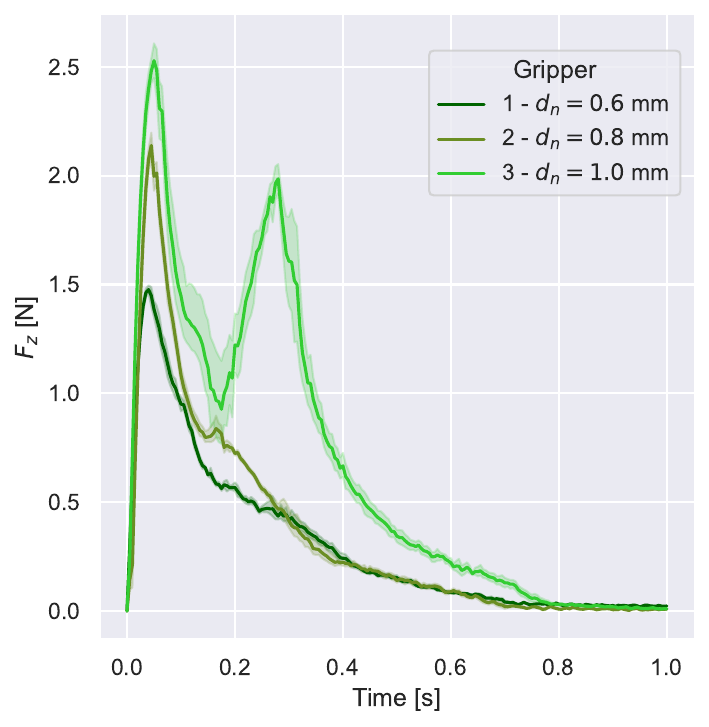}
 \caption{Average distribution lifting force $F_z$ during the vertical movement of the gripper for different grippers (Supply pressure 200~kPa, Flat surface).}
\label{fig-9}
\end{figure}

%FFFFFFFFFFFFFFFFFFFFFFFFFFFFFFFFFFFFFFFFFFFFFF
\begin{figure*}[tb]
\newcommand{\mywidth}{0.99}
\centering
	\begin{subfigure}[b]{0.23\linewidth}
         \centering
         \includegraphics[width=\mywidth\linewidth]{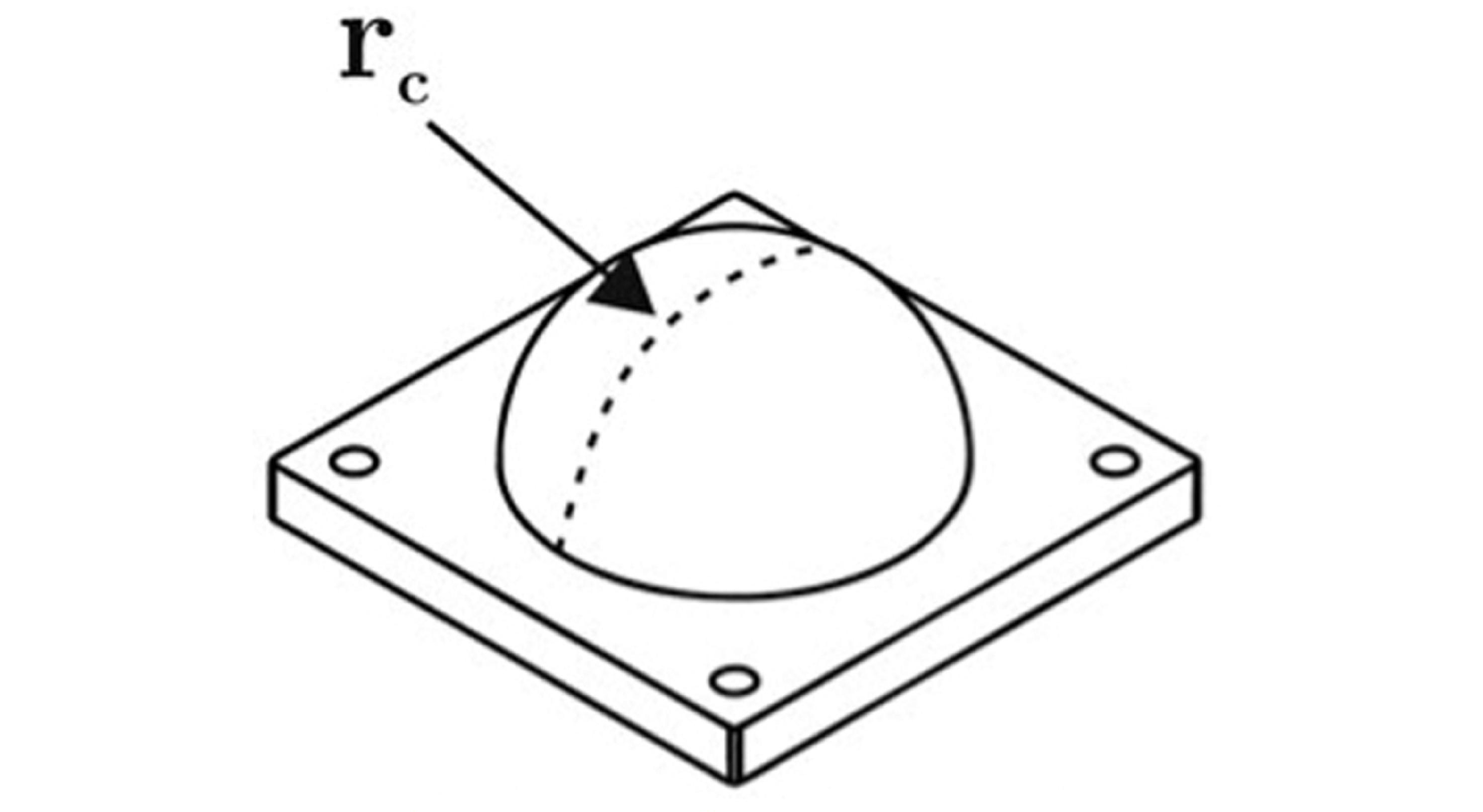}
         \caption{dome (convex) \label{subfig:tool_insertion_retraction}}
	\end{subfigure}
	~
	\begin{subfigure}[b]{0.23\linewidth}
         \centering
         \includegraphics[width=\mywidth\linewidth]{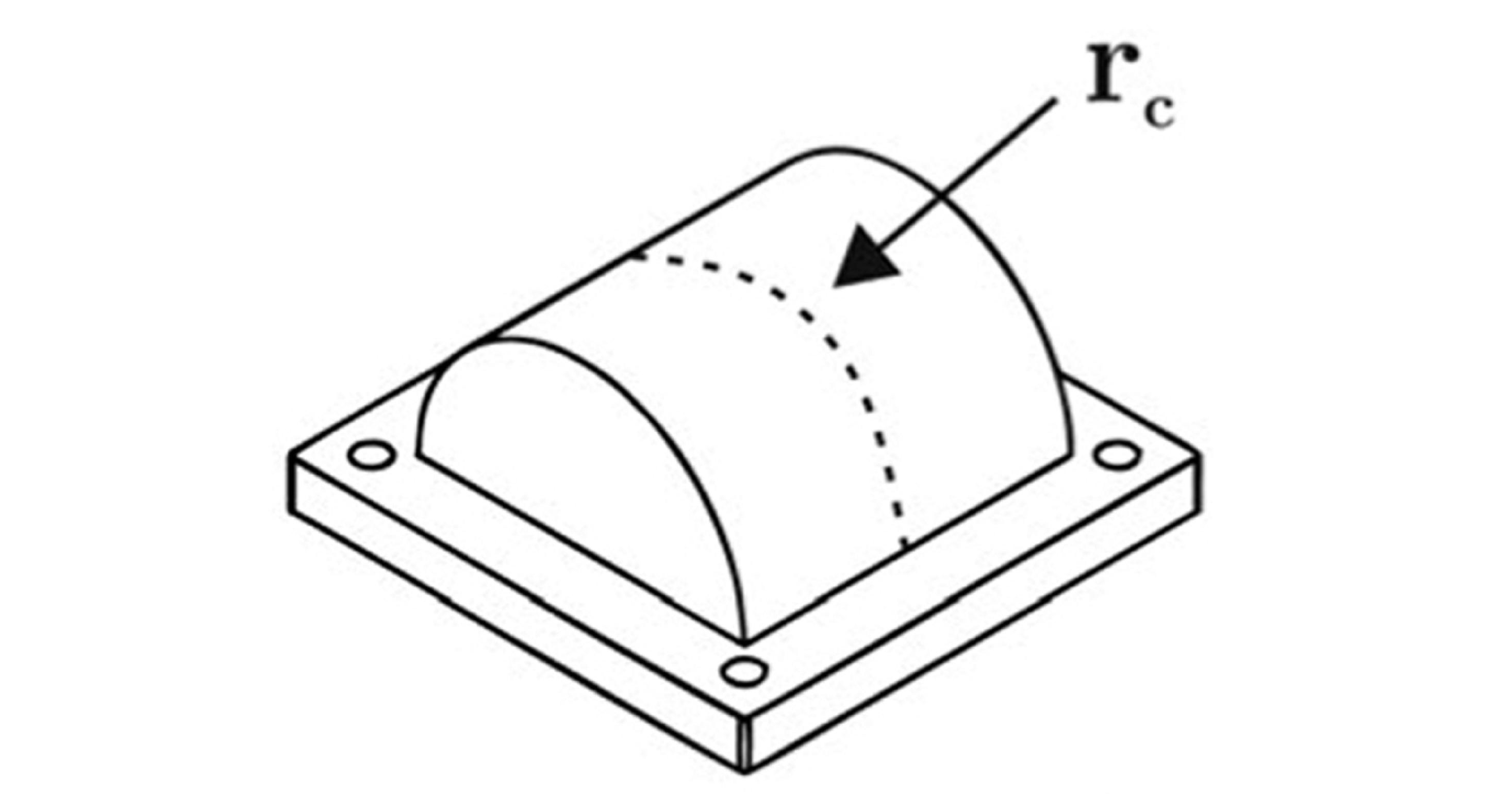}
         \caption{cylinder (convex) \label{subfig:tool_rotation}}
	\end{subfigure}
        ~
	\begin{subfigure}[b]{0.23\linewidth}
         \centering
         \includegraphics[width=\mywidth\linewidth]{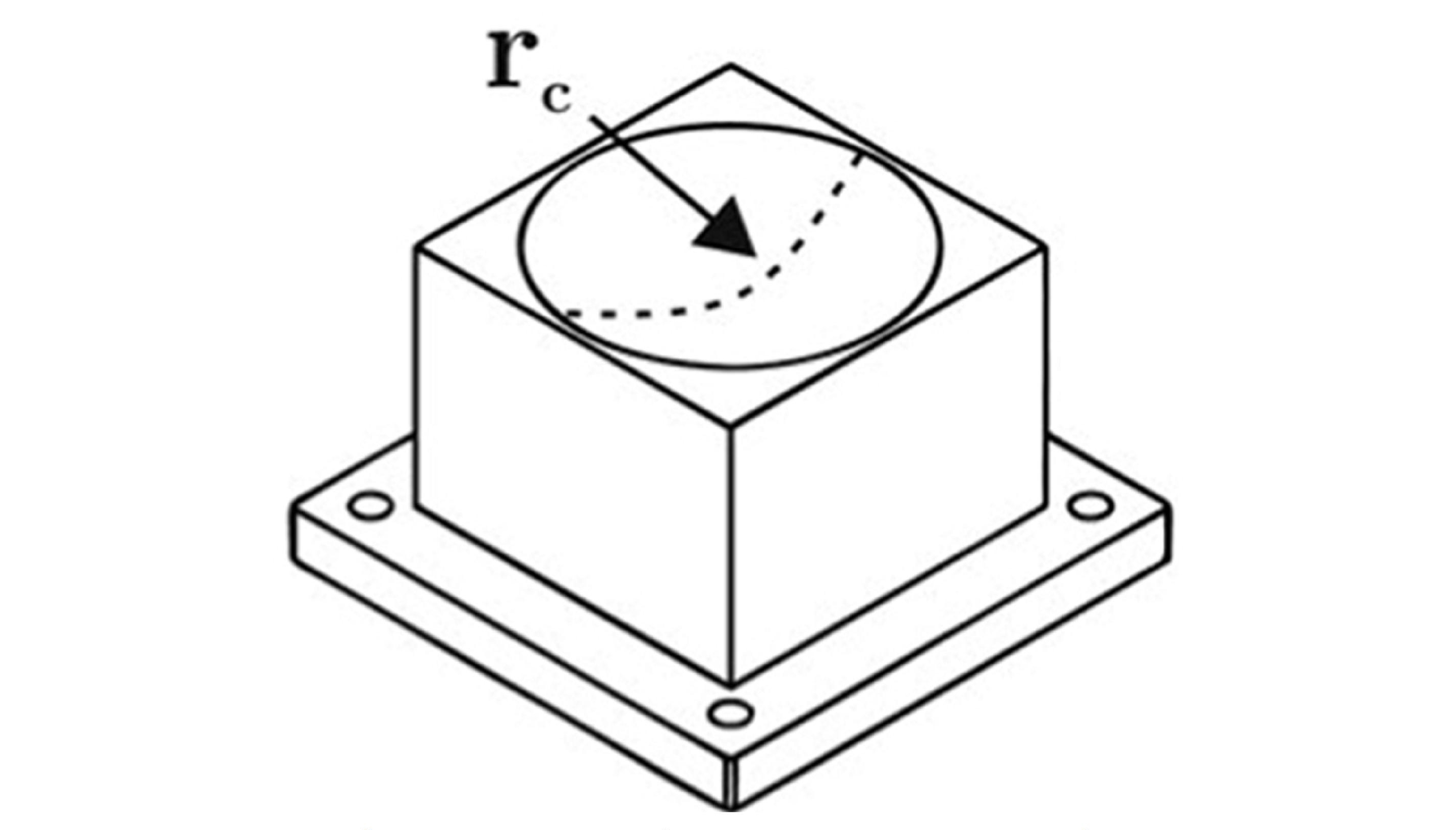}
         \caption{dome (concave) \label{subfig:tool_rotation1}}
	\end{subfigure}
	~
	\begin{subfigure}[b]{0.23\linewidth}
         \centering
         \includegraphics[width=\mywidth\linewidth]{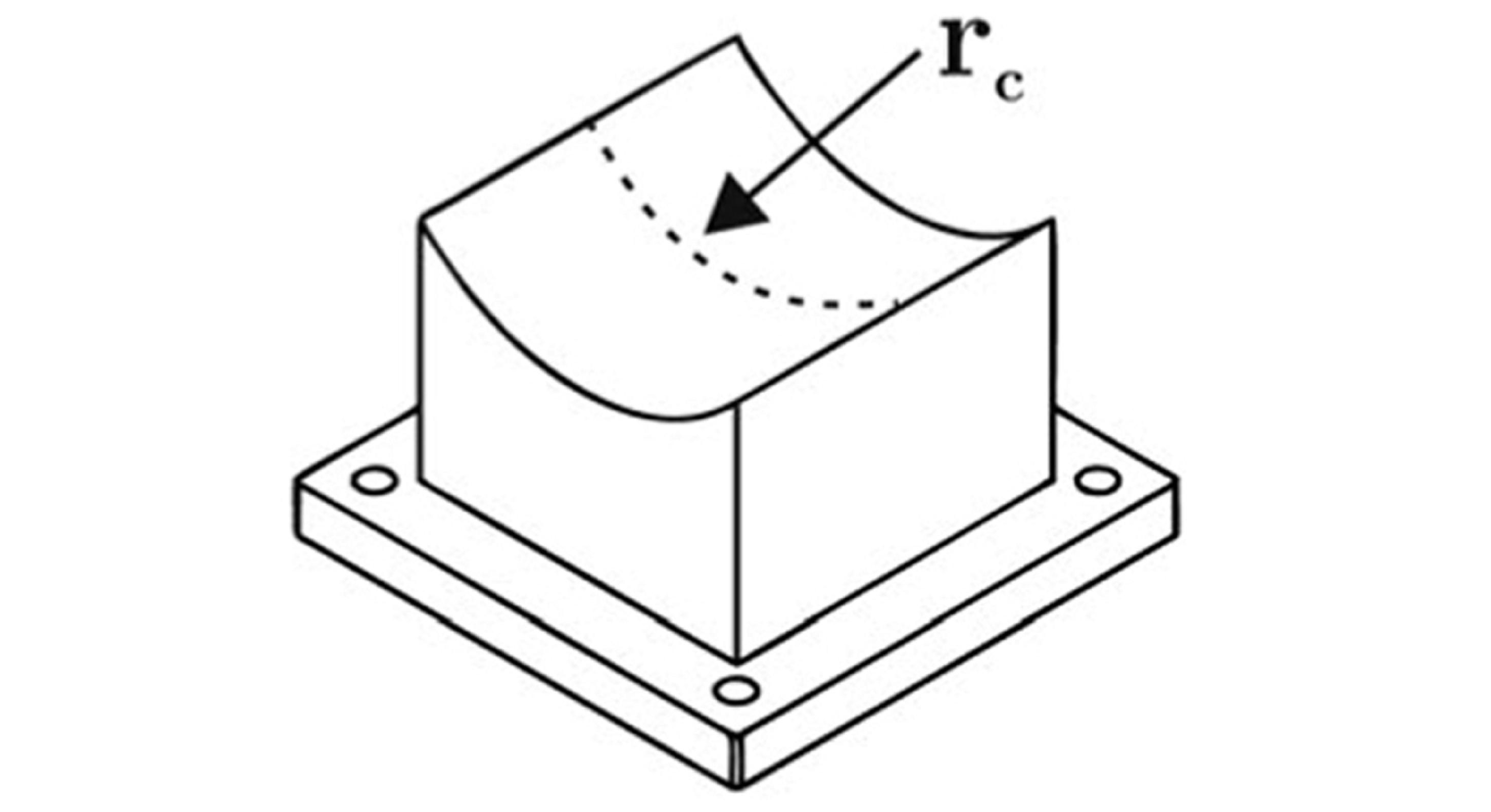}
         \vspace{-4mm}
         \caption{cylinder (concave) \label{subfig:tool_pitch}}
	\end{subfigure}\\
	%%%%%%%%%%%%%%%%%%%%%%%%%%%%%%%
 	\begin{subfigure}[b]{0.23\linewidth}
         \centering
         \includegraphics[width=\mywidth\linewidth]{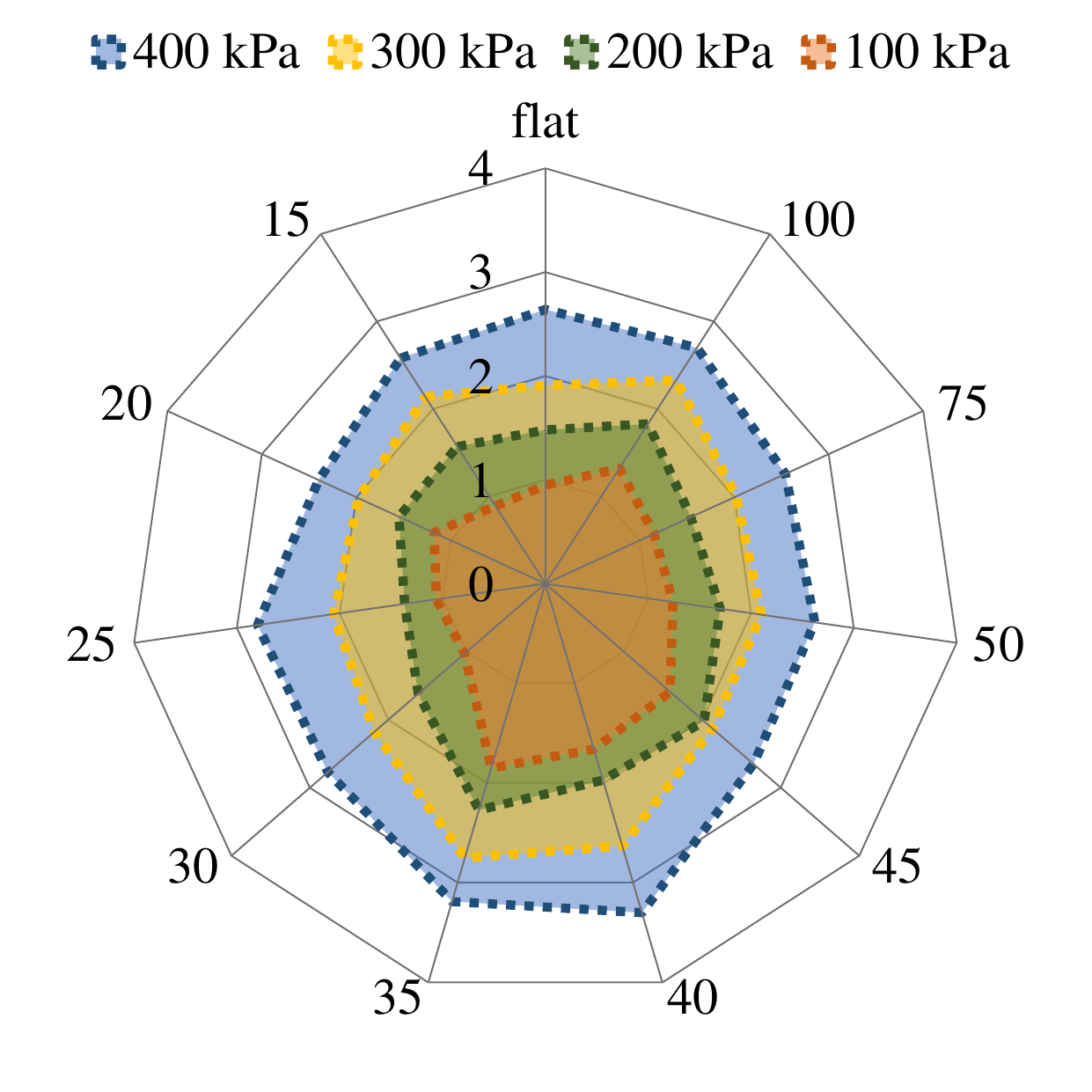}
         \vspace{-7mm}
        \caption{$d_n = 0.6$ mm \label{subfig:tool_insertion_retraction}}
	\end{subfigure}
	~
	\begin{subfigure}[b]{0.23\linewidth}
         \centering
         \includegraphics[width=\mywidth\linewidth]{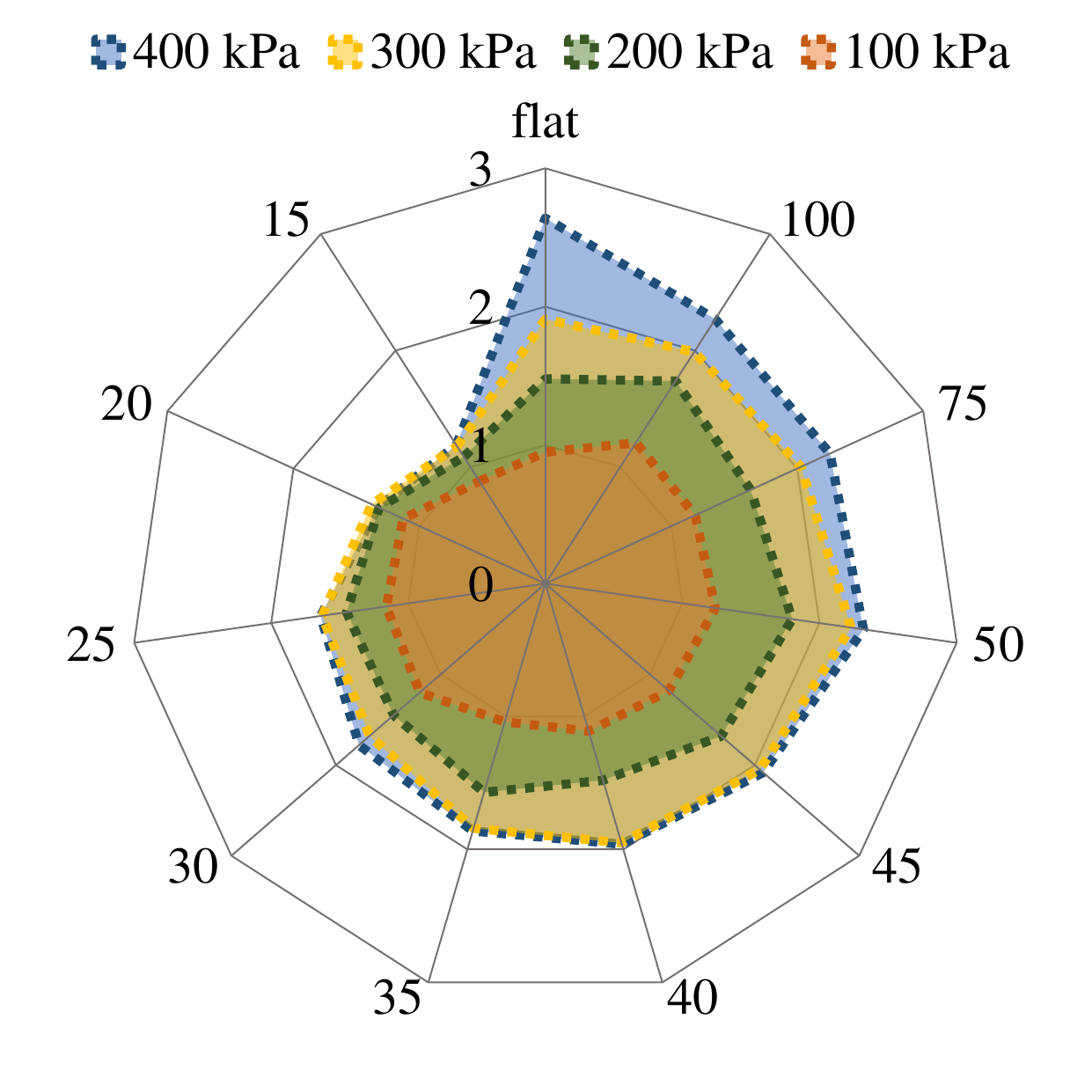}
         \vspace{-7mm}
         \caption{$d_n = 0.6$ mm \label{subfig:tool_rotation}}
	\end{subfigure}
        ~
	\begin{subfigure}[b]{0.23\linewidth}
         \centering
         \includegraphics[width=\mywidth\linewidth]{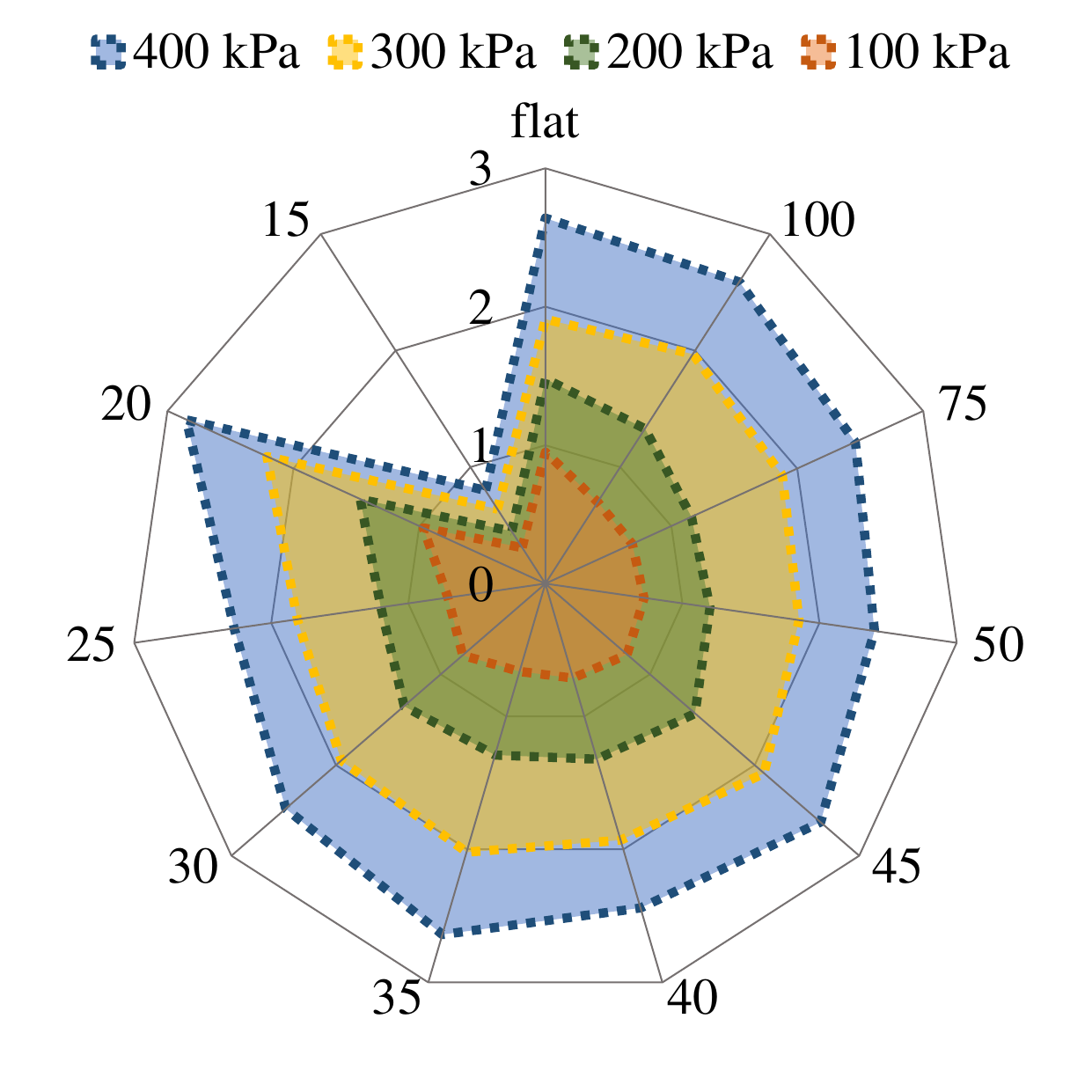}
         \vspace{-7mm}
         \caption{$d_n = 0.6$ mm \label{subfig:tool_rotation1}}
	\end{subfigure}
	~
	\begin{subfigure}[b]{0.23\linewidth}
         \centering
         \includegraphics[width=\mywidth\linewidth]{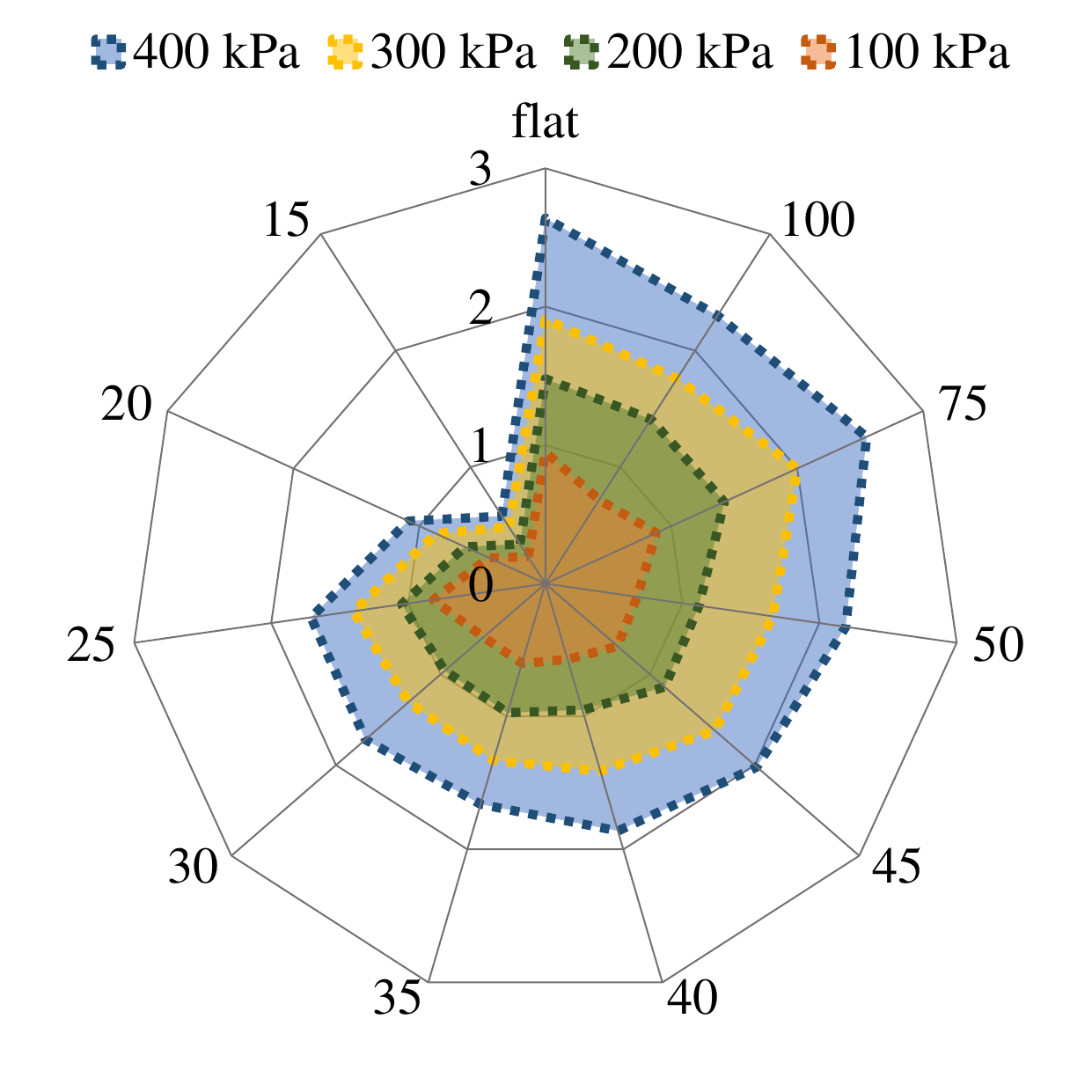}
         \vspace{-7mm}
         \caption{$d_n = 0.6$ mm \label{subfig:tool_pitch}}
	\end{subfigure}\\
	%%%%%%%%%%%%%%%%%%%%%%%%%%%%%%%
  	\begin{subfigure}[b]{0.23\linewidth}
         \centering
         \includegraphics[width=\mywidth\linewidth]{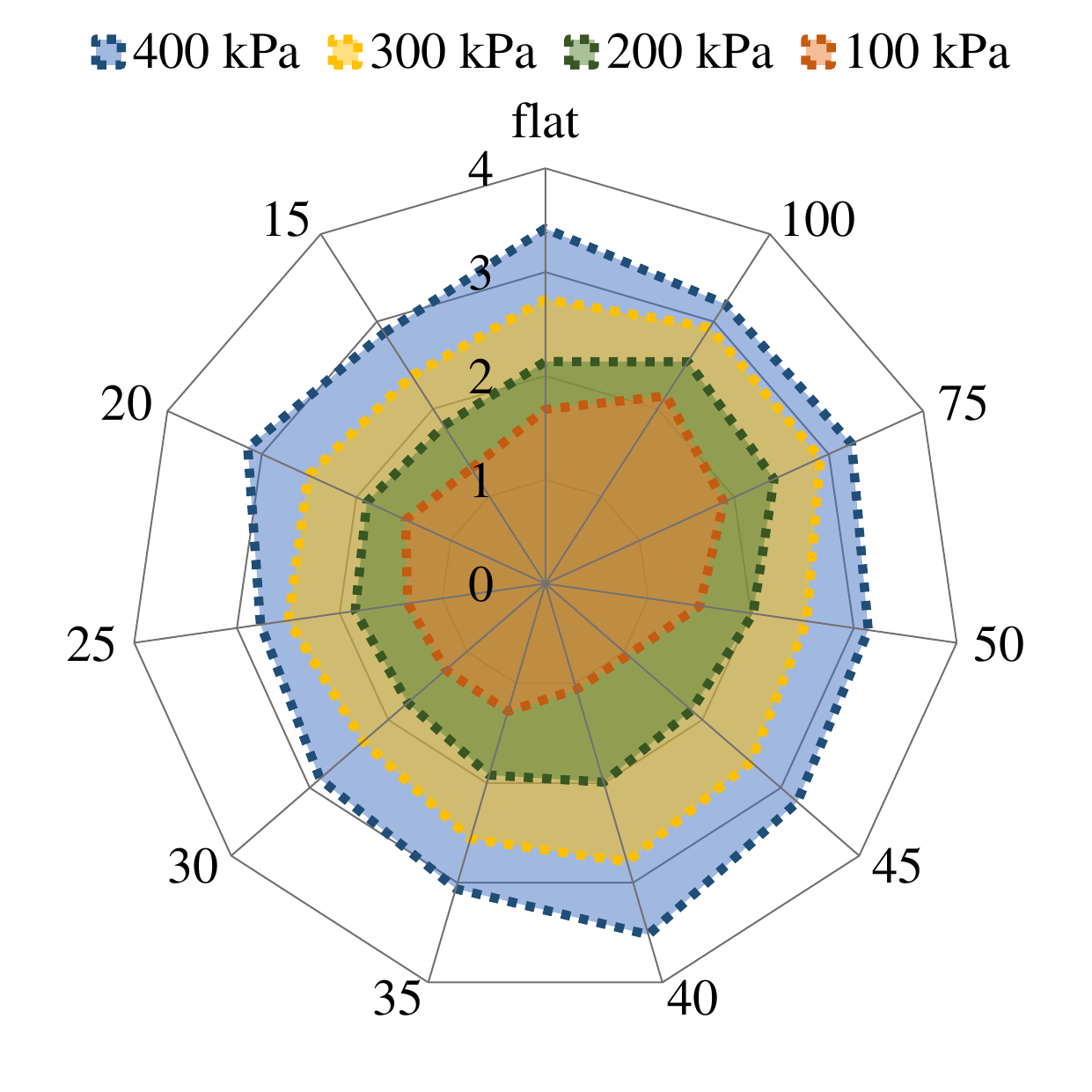}
         \vspace{-7mm}
         \caption{$d_n = 0.8$ mm \label{subfig:tool_insertion_retraction}}
	\end{subfigure}
	~
	\begin{subfigure}[b]{0.23\linewidth}
         \centering  \includegraphics[width=\mywidth\linewidth]{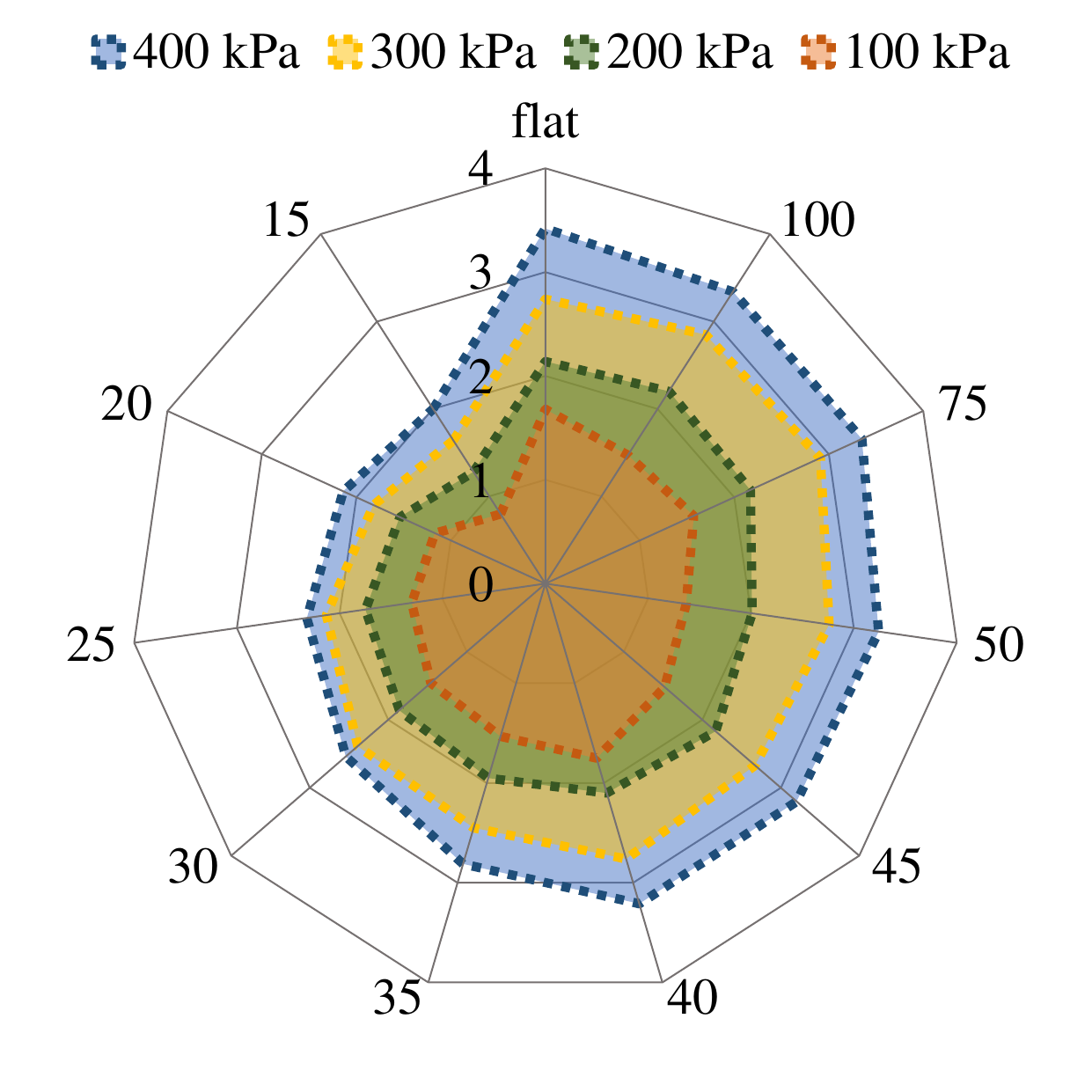}
         \vspace{-7mm}
         \caption{$d_n = 0.8$ mm \label{subfig:tool_rotation}}
	\end{subfigure}
        ~
	\begin{subfigure}[b]{0.23\linewidth}
         \centering
         \includegraphics[width=\mywidth\linewidth]{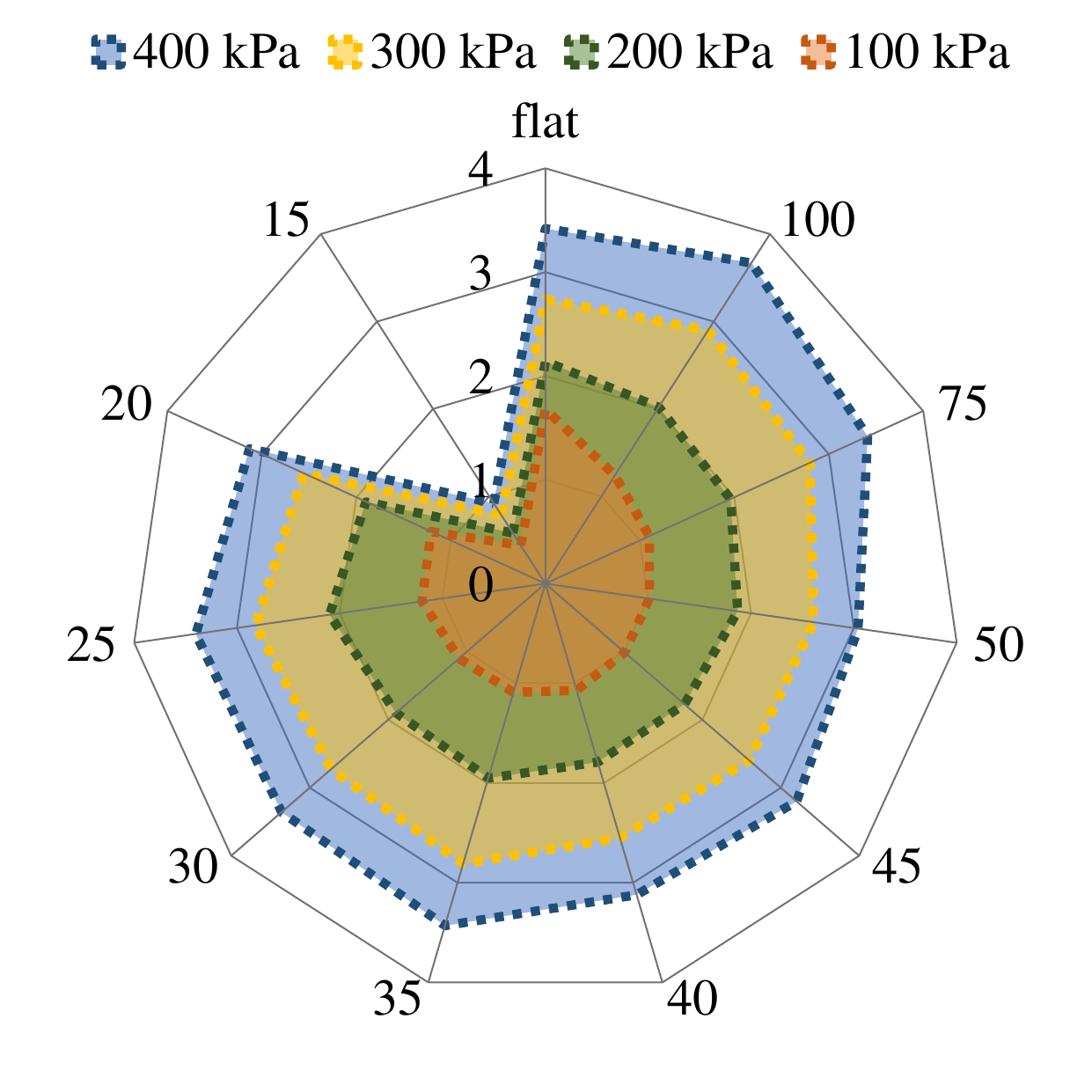}
         \vspace{-7mm}
         \caption{$d_n = 0.8$ mm \label{subfig:tool_rotation1}}
	\end{subfigure}
	~
	\begin{subfigure}[b]{0.23\linewidth}
         \centering
         \includegraphics[width=\mywidth\linewidth]{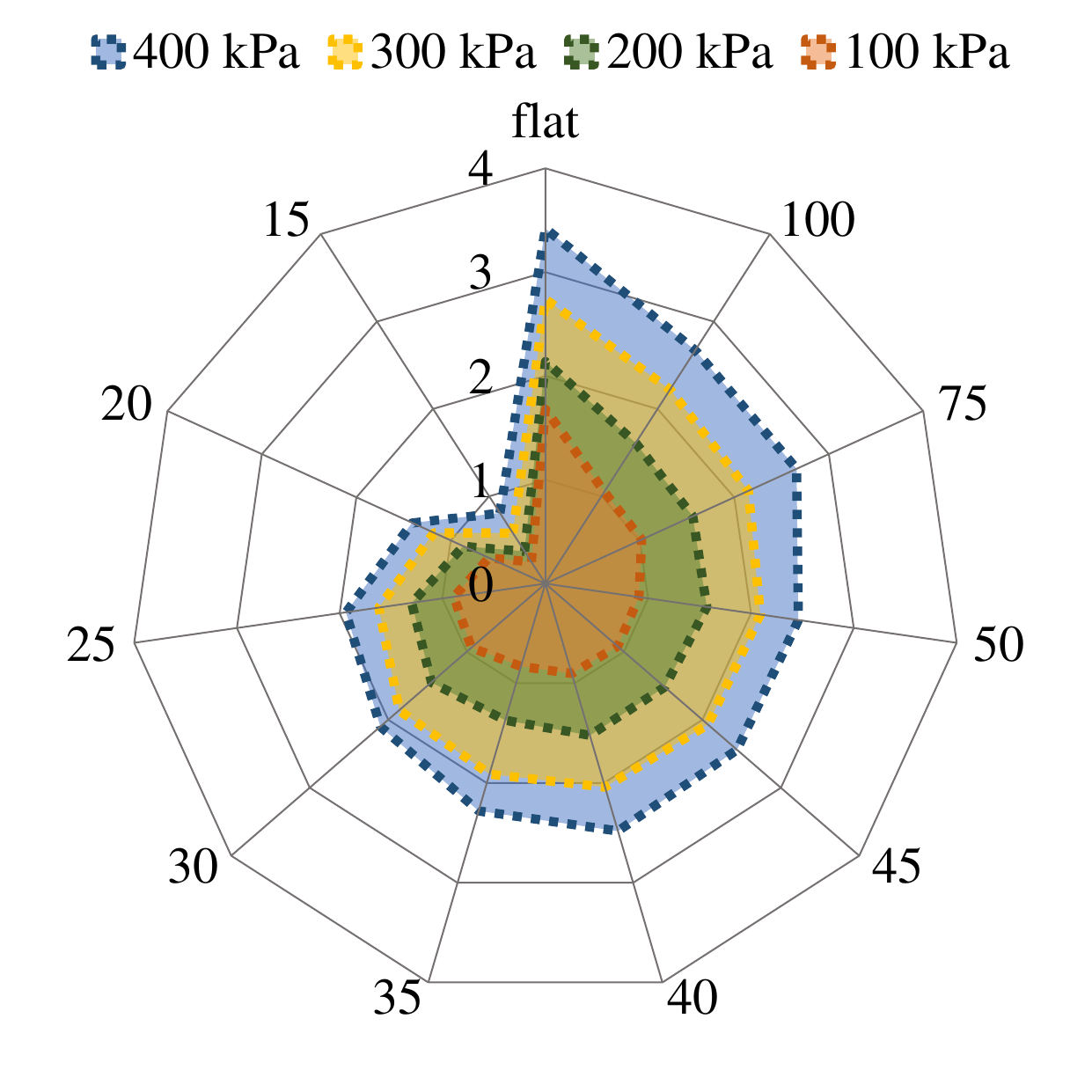}
         \vspace{-7mm}
         \caption{$d_n = 0.8$ mm \label{subfig:tool_pitch}}
	\end{subfigure}\\
	%%%%%%%%%%%%%%%%%%%%%%%%%%%%%%%
  	\begin{subfigure}[b]{0.23\linewidth}
         \centering
         \includegraphics[width=\mywidth\linewidth]{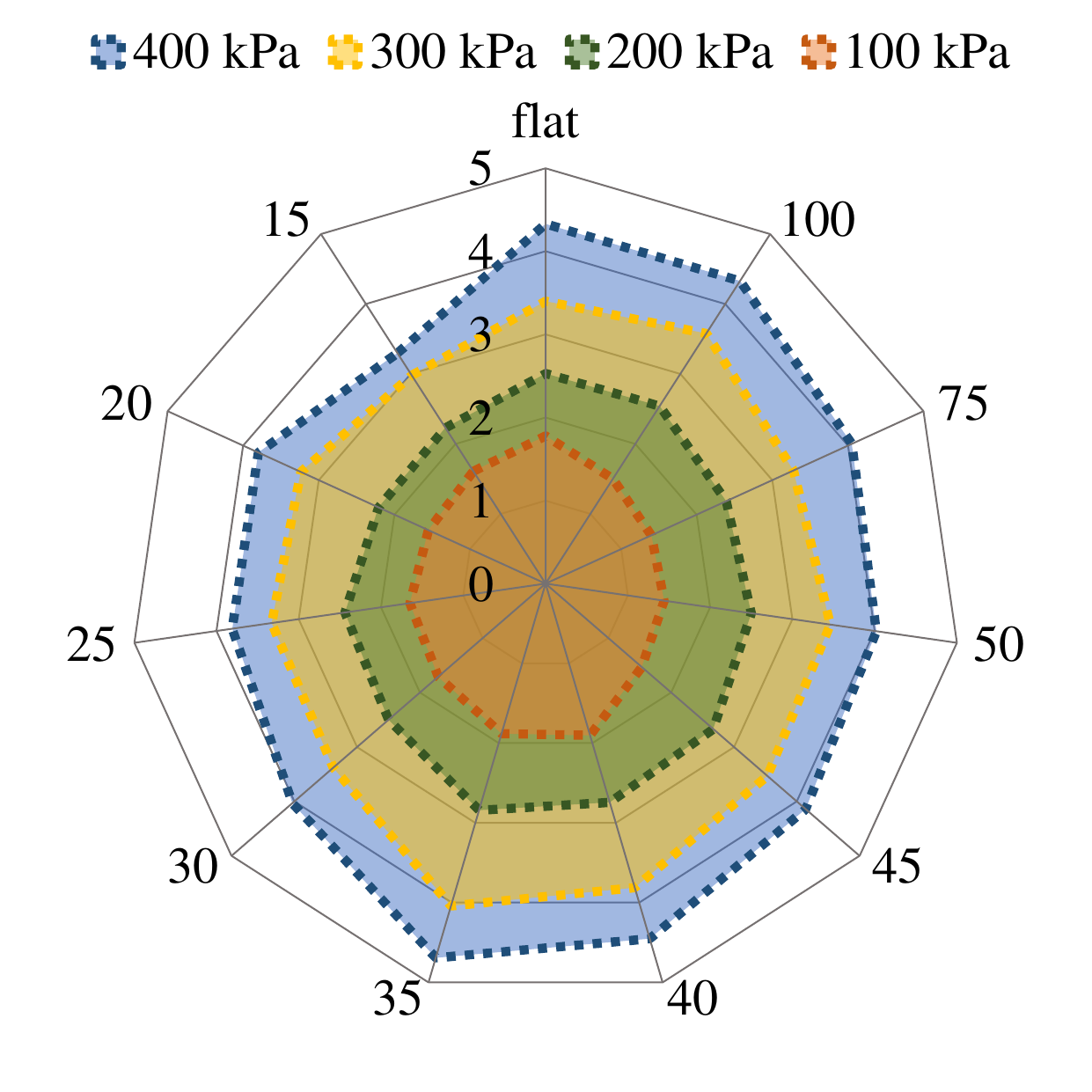}
         \vspace{-7mm}
         \caption{$d_n = 1$ mm \label{subfig:tool_insertion_retraction}}
	\end{subfigure}
	~
	\begin{subfigure}[b]{0.23\linewidth}
         \centering
         \includegraphics[width=\mywidth\linewidth]{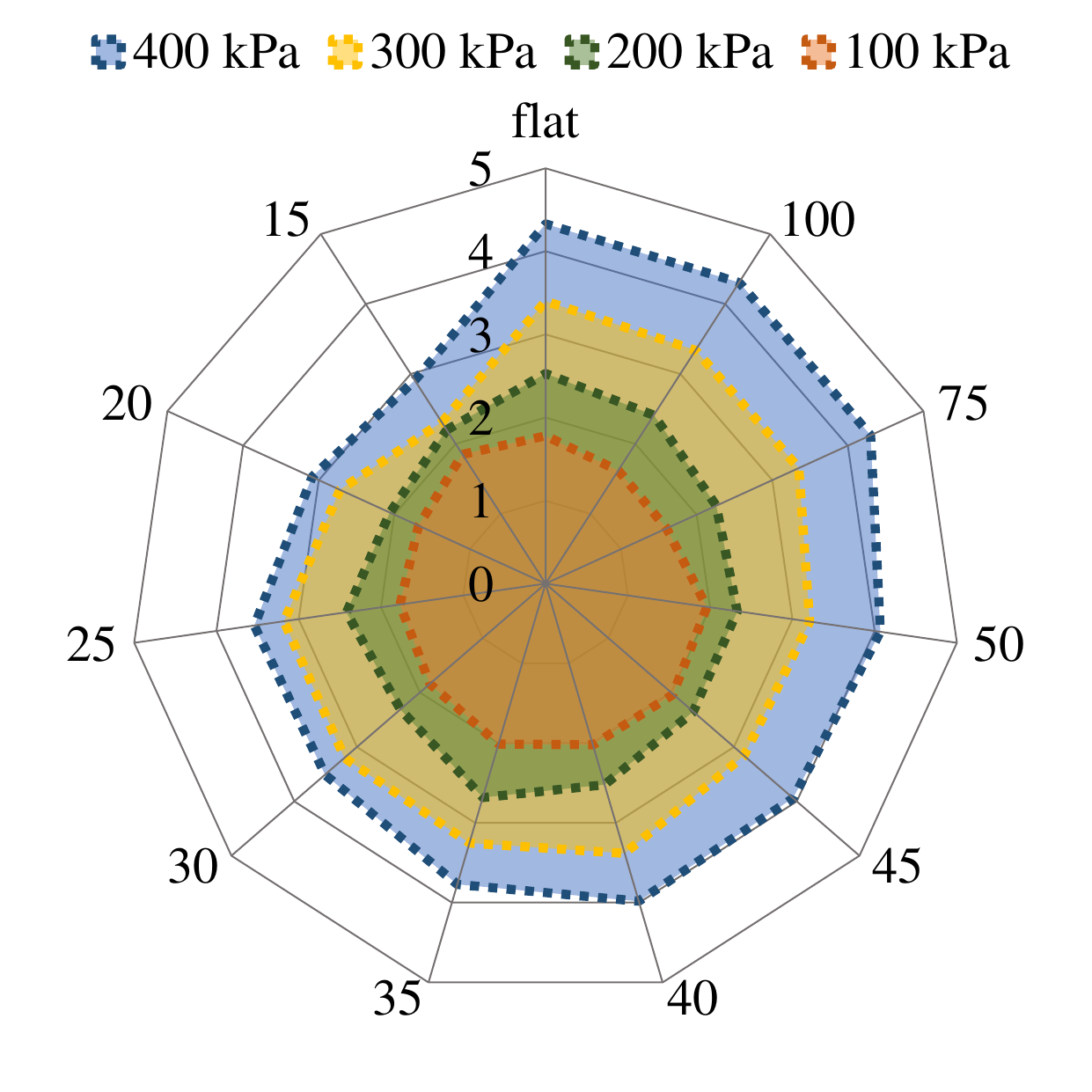}
         \vspace{-7mm}
         \caption{$d_n = 1$ mm \label{subfig:tool_rotation}}
	\end{subfigure}
        ~
	\begin{subfigure}[b]{0.23\linewidth}
         \centering
         \includegraphics[width=\mywidth\linewidth]{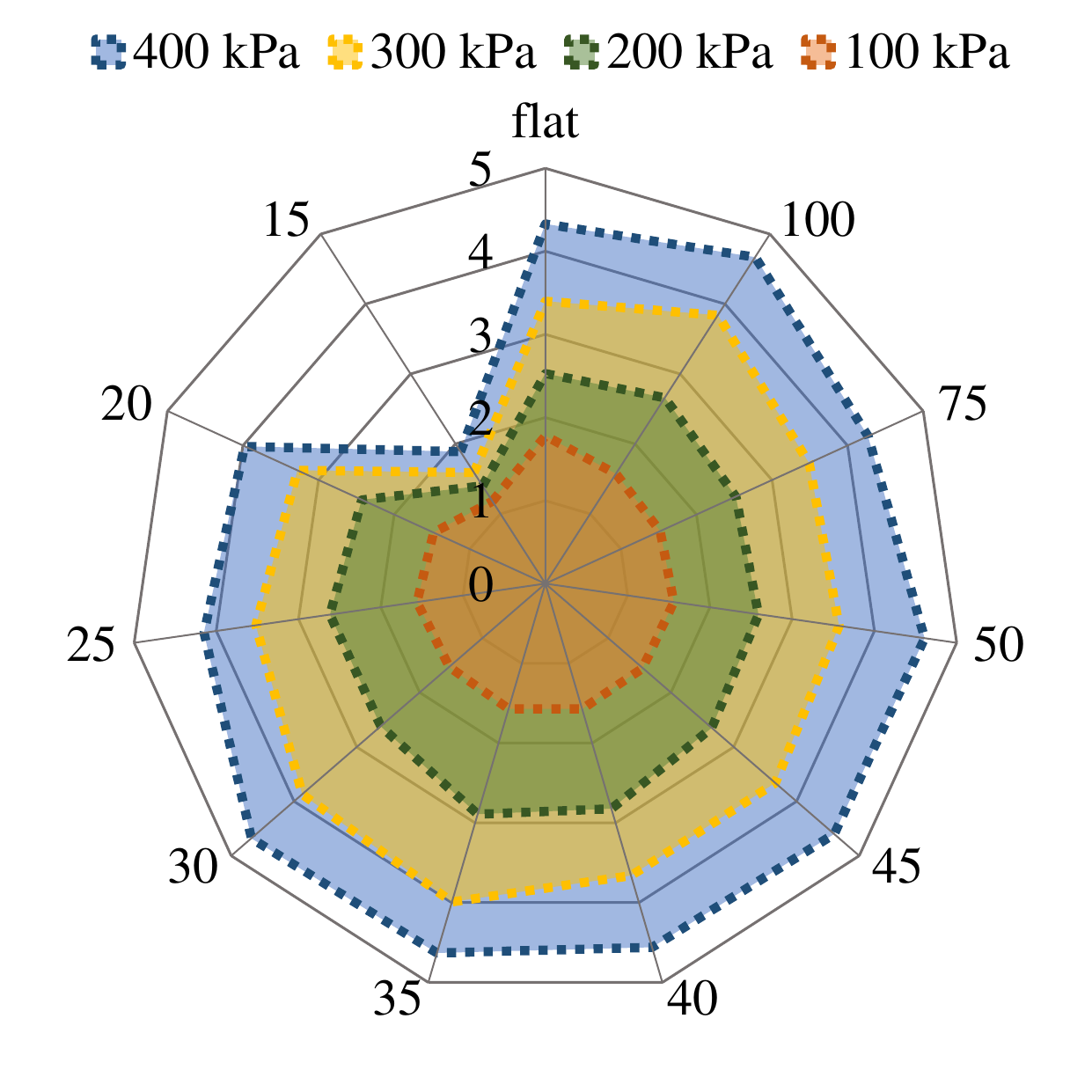}
         \vspace{-7mm}
         \caption{$d_n = 1$ mm \label{subfig:tool_rotation1}}
	\end{subfigure}
	~
	\begin{subfigure}[b]{0.23\linewidth}
         \centering
         \includegraphics[width=\mywidth\linewidth]{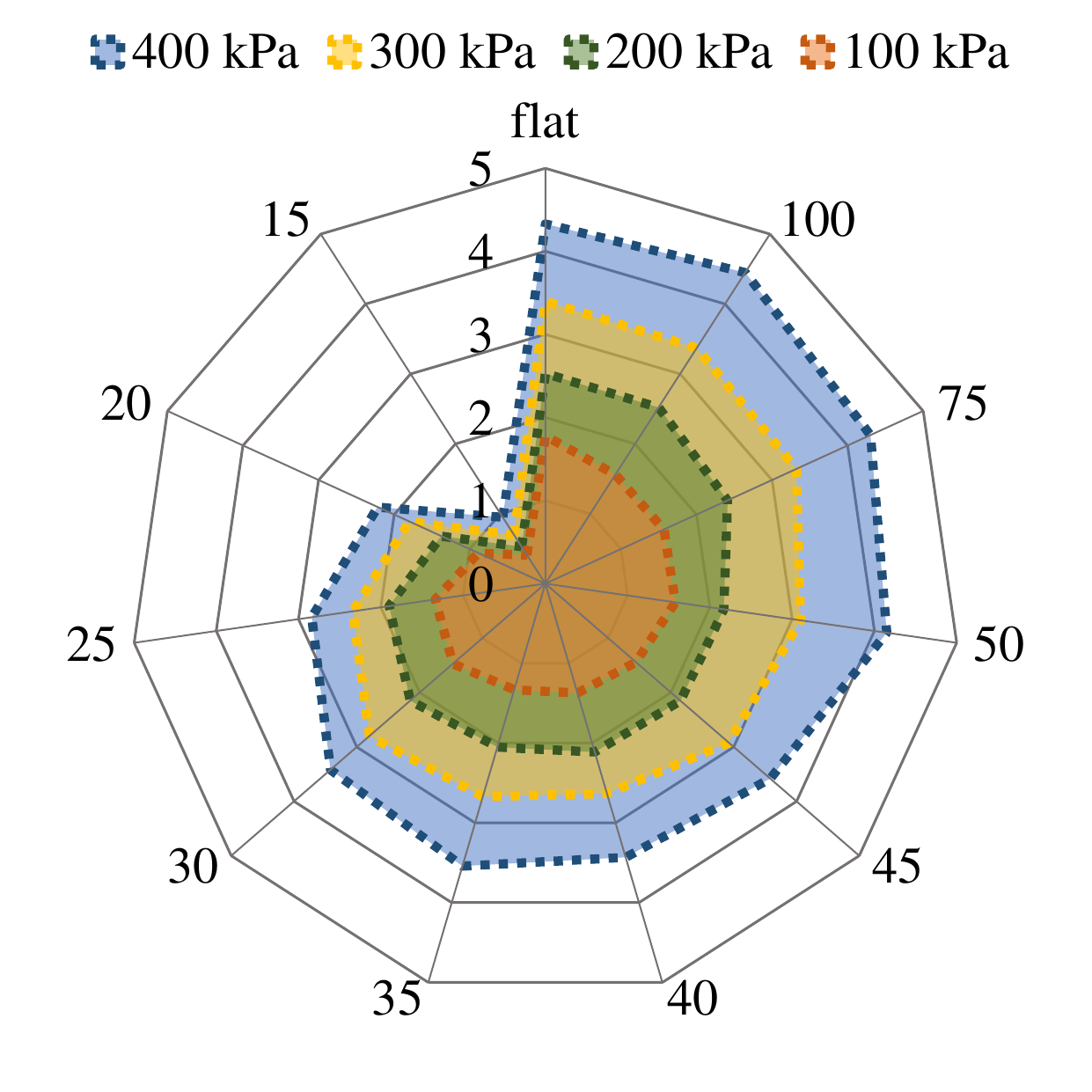}
         \vspace{-7mm}
         \caption{$d_n = 1$ mm \label{subfig:tool_pitch}}
	\end{subfigure}\\
 \caption{Characteristics of the lifting force (Newtons) of vortex grippers with soft tissues surfaces of different radius (mm): where horizontally grippers (Gripper 1 ($d_n = 0.6$ mm) - e, f, g, h; Gripper 2 ($d_n = 0.8$ mm) - i, j, k, l; Gripper 3 ($d_n = 1.0$ mm) - m, n, o, p), and vertically the type of surface (Dome (convex) - e, i, m; Cylinder (convex) - f, j, n; Dome (concave) - g, k, o; Cylinder (concave) - h, l, p).\label{fig-10-1}}
\end{figure*}
%FFFFFFFFFFFFFFFFFFFFFFFFFFFFFFFFFFFFFFFFFFFFFF

During experimental studies of the determination lifting force of a vortex gripper for soft objects of different shapes, we obtain the distribution of the $F_z$ force during the vertical movement of the gripper. Having 10 experiments with the same parameters (pressure, surface, gripper), allows us to obtain the average distribution lifting force $F_z$ during the vertical movement of the gripper (Fig.~\ref{fig-8}). Knowing this distribution, we can find the average maximum lifting force ($F_l^{max}$) that can create a vortex gripper with a suitable surface at a certain supply pressure (Fig.~\ref{fig-8}). Fig.~\ref{fig-8} shows that with an increase in the supply pressure of the gripper 1 by 100 kPa, the lifting force increases by an average of 0.5 N. This is obvious since an increase in the supply pressure leads to a uniform increase in the angular velocity of the air $\omega$ (1) in the gripper cavity. In addition, the achievement of the maximum lifting force $F_l^{max}$ shifts in time to the right (Fig.~\ref{fig-8}), which indicates an increase in the mass flow and the optimal height $h$ (Fig.~\ref{fig-3}) between the object and the gripper.

A similar situation with the growth of mass flows is observed when the diameter of the nozzle elements of the vortex gripper increases (Fig.~\ref{fig-9}). This, in turn, leads to a shift in reaching the maximum lifting force $F_l^{max}$ over time to the right, and it increases by an average of 0.5 N when the diameter of the nozzles $d_n$ increases by 0.2 mm (pressure supply 200 kPa, flat surface). Unlike the increase in supply pressure, when the diameter of the nozzles increases (increase in mass flow), a small jump in the lifting force at $d_n$ = 0.8 mm (0.16 $<$ Time $<$ 0.2) and a significant jump in the lifting force at $d_n$ = 1.0 mm (0.18 $<$ Time $<$ 0.4) begin to appear on the distribution of the lifting force over time. The reason for this is sufficient mass flow and lifting force to get deformation of the soft tissue at a significant height $h$ and re-lifting tissue. Which in turn leads to the manifestation of this jump in the lifting force distribution (Fig.~\ref{fig-9}). This feature of the vortex gripping device for a flat surface does not play a significant role. However, it can play an important role in the formation of the lifting force of other types of surfaces. Because of this, a multivariate experiment was conducted to find the maximum lifting force (Fig.~\ref{fig-10-1}) for grippers with three different nozzle diameters ($d_n$ = 0.6, 0.8, 1.0 mm), four supply pressures (100, 200, 300, and 400 kPa), and 41 soft surfaces of different radii and types.

%-------------------------------------------------
\subsection{Dome Convex Surface}
\begin{figure}[t]
\centering
\subfloat[Dome Convex Surface]{\includegraphics[width=0.9\linewidth,clip ,trim=0pt 0pt 0pt 0pt]{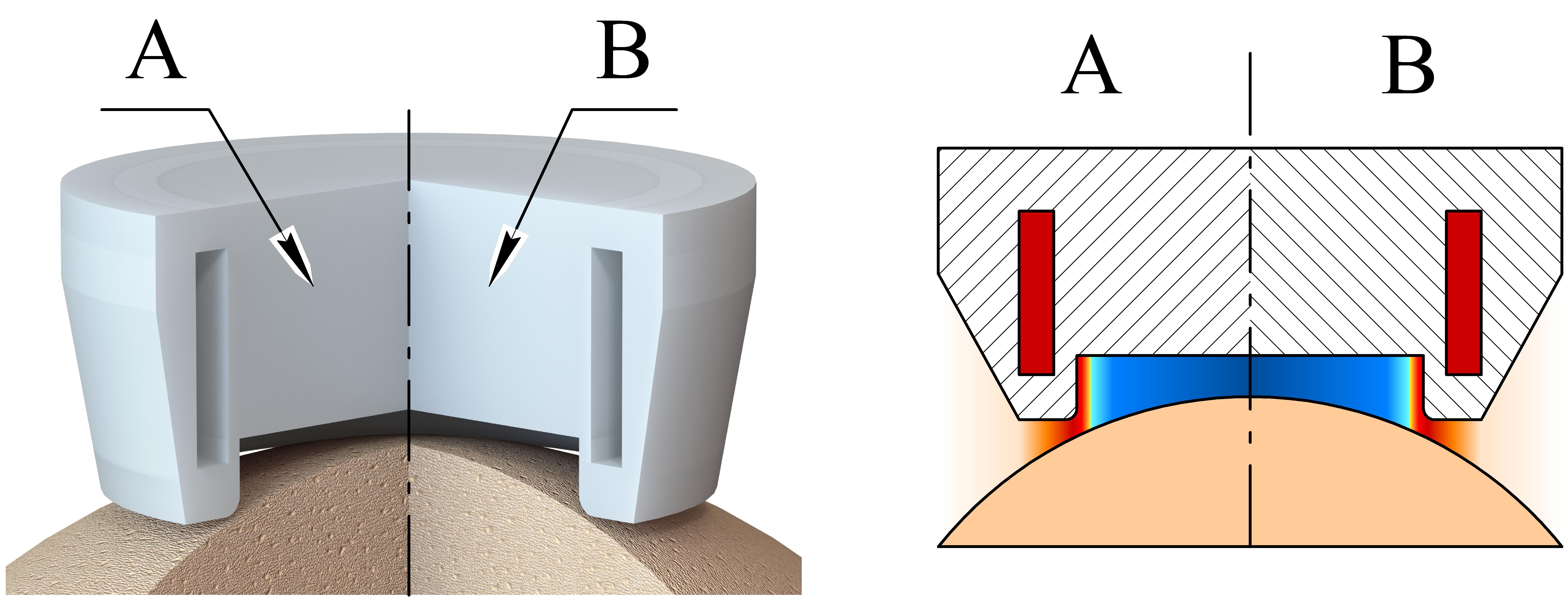}%
\label{fig11_1}}
\hfil
\\
\subfloat[Cylinder Convex Surface]{\includegraphics[width=0.9\linewidth,clip ,trim=0pt 0pt 0pt 0pt]{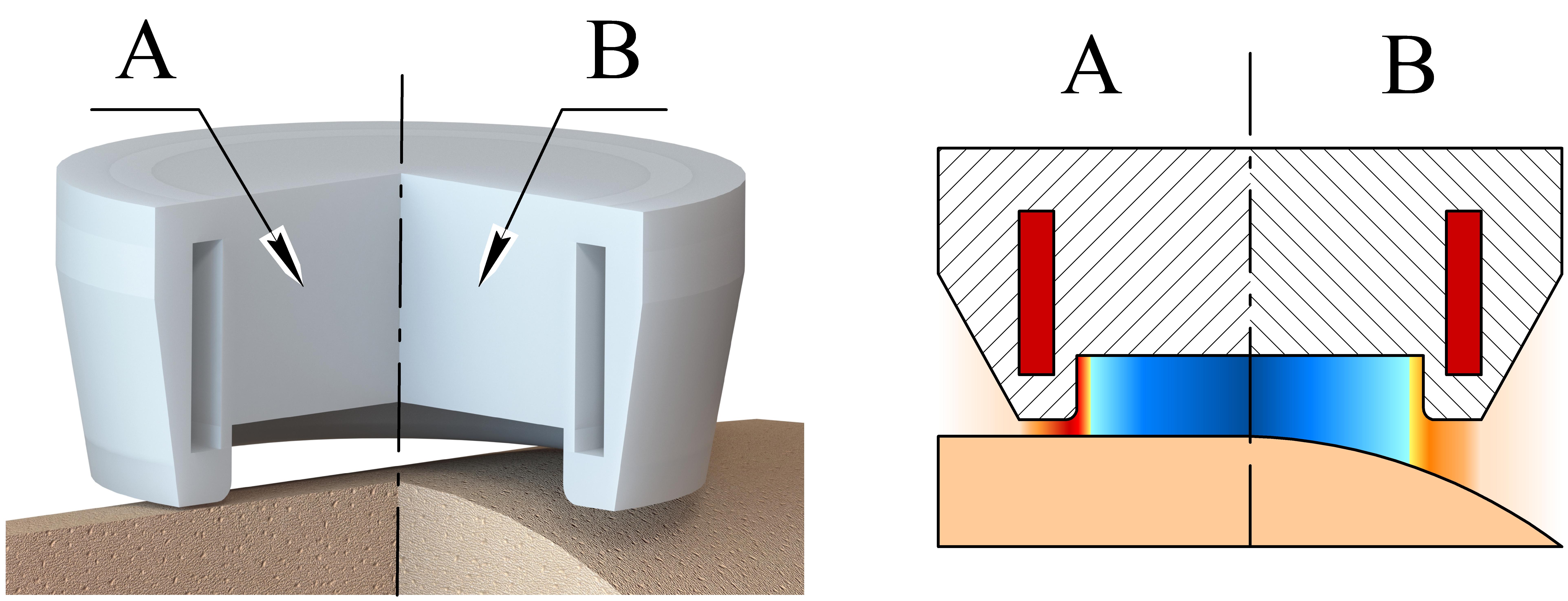}%
\label{fig11_2}}
\hfil
\\
\subfloat[Dome Concave Surface]{\includegraphics[width=0.9\linewidth,clip ,trim=0pt 0pt 0pt 0pt]{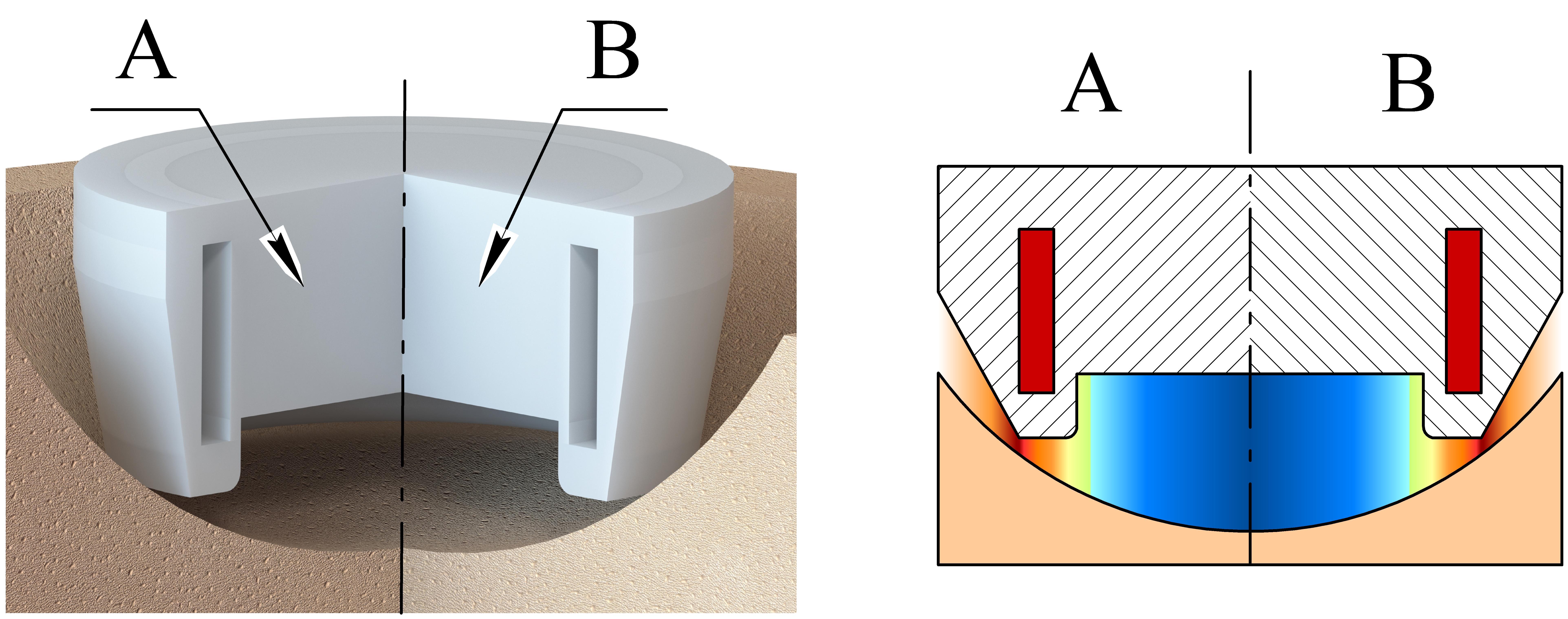}%
\label{fig11_3}}
\hfil
\\
\subfloat[Cylinder Concave Surface]{\includegraphics[width=0.9\linewidth,clip ,trim=0pt 0pt 0pt 0pt]{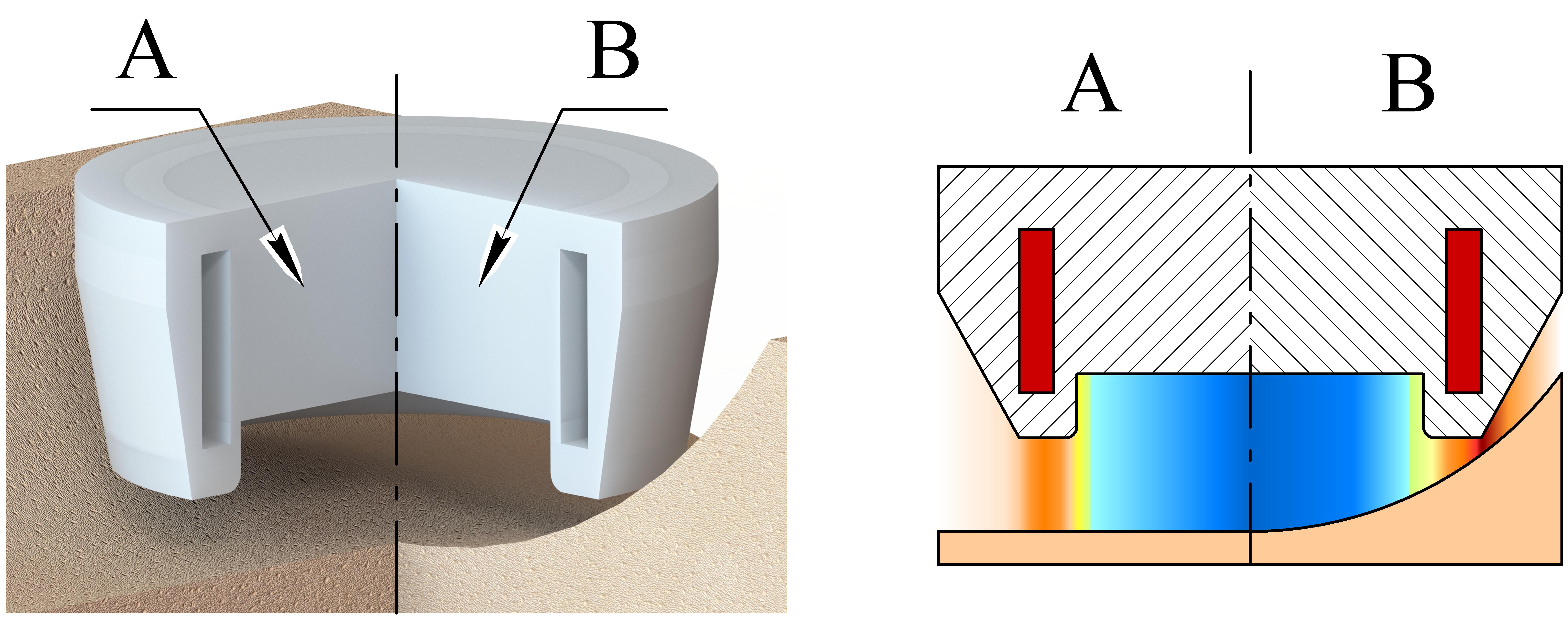}%
\label{fig11_4}}
\hfil
\caption{The gap between the active surface vortex gripper and soft tissue surfaces of different shapes, where A and B are their perpendicular cross sections along the axis of symmetry.}
\label{fig-11}
\end{figure}

%-------------------------------------------------
As can be seen from the distribution of lifting forces for all surfaces, the most stable force is preserved during the grasping of the dome's concave soft surface (Fig.~\ref{fig-10-1},~a,~e,~i,~m). This is evidenced by the practically uniform circle of the force characteristic for all three grippers. This is due to the fact that the vortex gripper, having a cavity and a rounded chamfer of the transition from the cavity to the flange of the gripper, allows the dome's convex surface during grasping to enter this cavity. Which practically maintains a uniform gap between the surface and the gripper (Fig.~\ref{fig11_1}). Such an effect even has a positive effect on the lifting force for the dome's convex surfaces at a certain radius. Thus, for surface rounding radii $r_c$ from 30 to 45 mm (for gripper 1 and 3), the lifting force increases in comparison with a flat surface by an average of 54\% for gripper 1 ($G_1$) and by 12\% for gripper 3 ($G_3$). For gripper 2 ($G_2$) and surface rounding radii $r_c$ from 35 to 45 mm, the lifting force on average remains at the same level as for a flat surface, but for other radii the lifting force decreases. 

%-------------------------------------------------
\subsection{Cylinder Convex Surface}
%-------------------------------------------------
Analyzing the force characteristics of the vortex gripper with a cylindrical convex surface (Fig.~\ref{fig-10-1}, b, f, j, n), it is obvious that the lifting force is reduced. It can also be seen from the graphs that there is a uniform decrease in the lifting force when the radius of the cylindrical surface decreases. For G$_1$, at a supply pressure of 100 kPa and a radius $r_c = 100$ mm, the lifting force is equal to 10 N and decreases to 5 N at a radius $r_c = 15$ mm. On average, for all supply pressures, the drop in the lifting force of the vortex gripper G$_1$ for a cylindrical convex surface (from $r_c = 100$ to 15 mm) is 37\%. This behavior is due to the fact that when grasping a cylindrical convex surface, it is not possible to create a uniform gap between the surface and the gripper (Fig.~\ref{fig11_2}). This leads to an uneven flow of air through the gap and increases the height $h$ when maximum force is reached. Which in turn minimizes the possibility of interaction of the airflow with part of the cylindrical surface. The tendency to reduce the lifting force when the radius of the cylindrical convex surface is reduced is also preserved for grippers G$_2$ (24\%) and G$_3$ (15\%). However, the decrease in the lifting is noticeably reduced, which is associated with an increase in the mass flow characteristics of the vortex grippers G$_2$ and G$_3$. That can easily be explained by the possibility of greater mass flow interacting with a cylindrical convex surface even with small radii of its rounding.

%-------------------------------------------------
\subsection{Dome Concave Surface}
%-------------------------------------------------
Discussing the results of the lifting force vortex gripping device with a dome concave surface (Fig.~\ref{fig-10-1}, c, g, k, o), it can be noted that the lifting force increases for almost all radii compared to a flat surface. This is due to the fact that due to the roundness of the dome's concave surface, the zone with negative pressure increases due to the shift of the narrowest gap to the edge of the gripper (Fig.~\ref{fig11_3}). However, for all grippers, when the radius of rounding of the concave surface is $r_c = 15$ mm, compared to $r_c = 20$ mm, the lifting force decreases by 70\% on average. This is explained by the reduction of the gap between the gripper body and the concave surface at $r_c = 15$ mm, which critically reduces the mass flow of air through the formed gap. However, even with a significant percentage reduction of the lifting force for a radius of $r_c = 15$ mm, the G$_3$ gripper at a supply pressure of 400 kPa can provide a lifting force of 2 N, which is 20 times greater than the self-weight of the vortex gripper.

%-------------------------------------------------
\subsection{Cylinder Concave Surface}
%-------------------------------------------------
One of the largest drops in the lifting force vortex gripping device is observed for a cylindrical concave surface (Fig.~\ref{fig-10-1}, d, h, l, p). Moreover, the lifting force declines uniformly with a decrease in the radius of the cylindrical concave surface for all grippers. This is caused by a non-uniform gap between the gripper and the cylindrical concave surface (Fig.~\ref{fig11_4}), as in the case of the cylindrical convex surface. However, for a cylindrical concave surface, the lifting force declines much faster. What can be explained by two negative effects at once, such as an increase in the gap of cross-section A (Fig.~\ref{fig11_4}) and the restriction of air leakage through the gap on the other side of cross-section B (Fig.~\ref{fig11_4}). However, as with a cylindrical convex surface, the reduction in the lifting force becomes not so great when the mass flow of the gripper increases, and thanks to this, the G$_3$ gripper achieves maximum force characteristics in comparison with G$_1$ and G$_2$.

%-------------------------------------------------
\subsection{General Evaluation}
%-------------------------------------------------

%FFFFFFFFFFFFFFFFFFFFFFFFFFFFFFFFFFFFFFFFFFFFFF
\begin{figure*}[tb]
\newcommand{\mywidth}{1}
\centering
	\begin{subfigure}[b]{0.236\linewidth}
         \centering
         \includegraphics[width=\mywidth\linewidth, clip, trim=0pt 0pt 0pt 0pt]{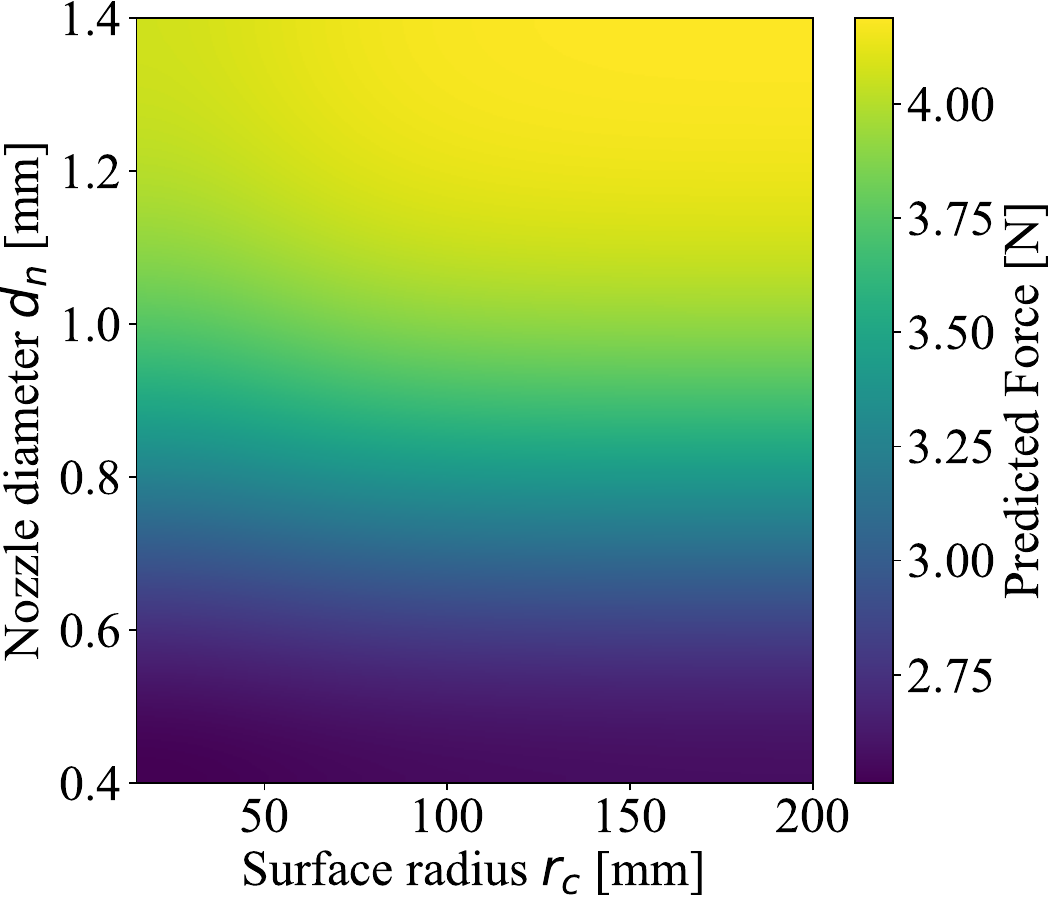}
        \caption{dome-convex \label{subfig12-1}}
	\end{subfigure}
	~
	\begin{subfigure}[b]{0.236\linewidth}
         \centering
         \includegraphics[width=\mywidth\linewidth, clip, trim=0pt 0pt 0pt 0pt]{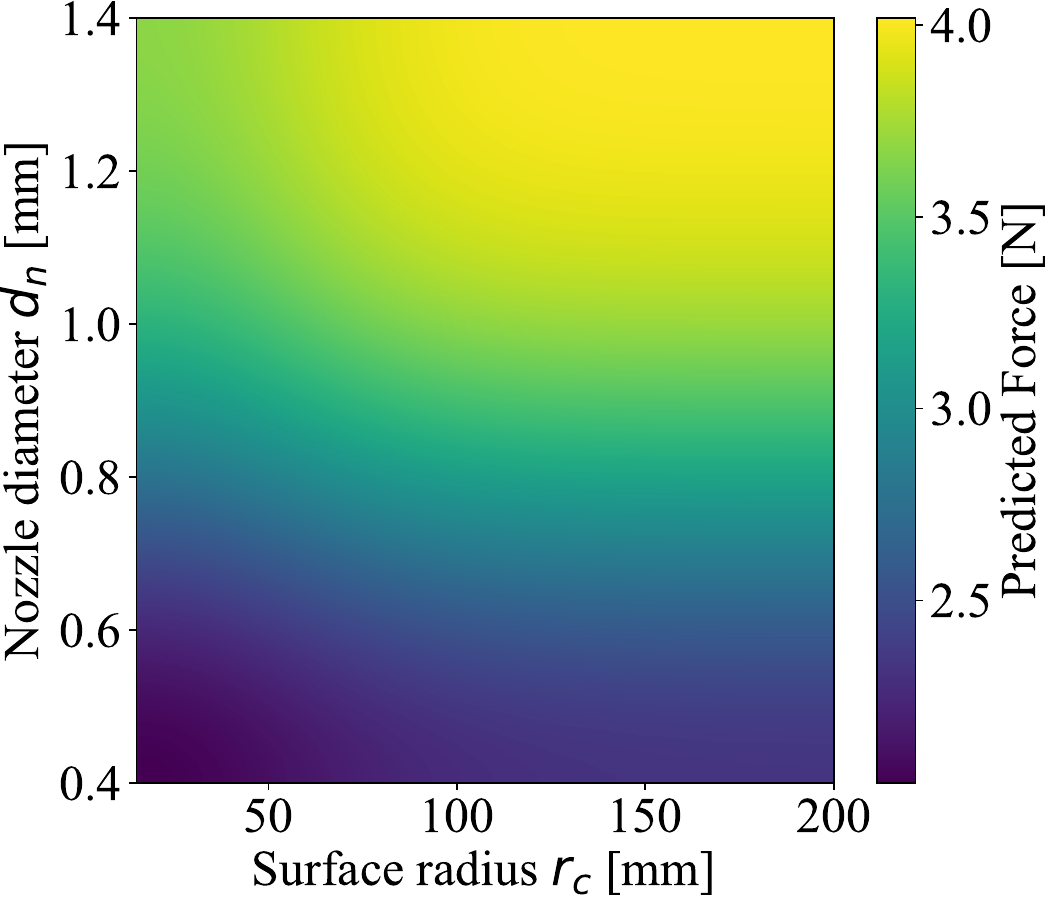}
         \caption{cylinder-convex \label{subfig12-2}}
	\end{subfigure}
        ~
	\begin{subfigure}[b]{0.236\linewidth}
         \centering
         \includegraphics[width=\mywidth\linewidth, clip, trim=0pt 0pt 0pt 0pt]{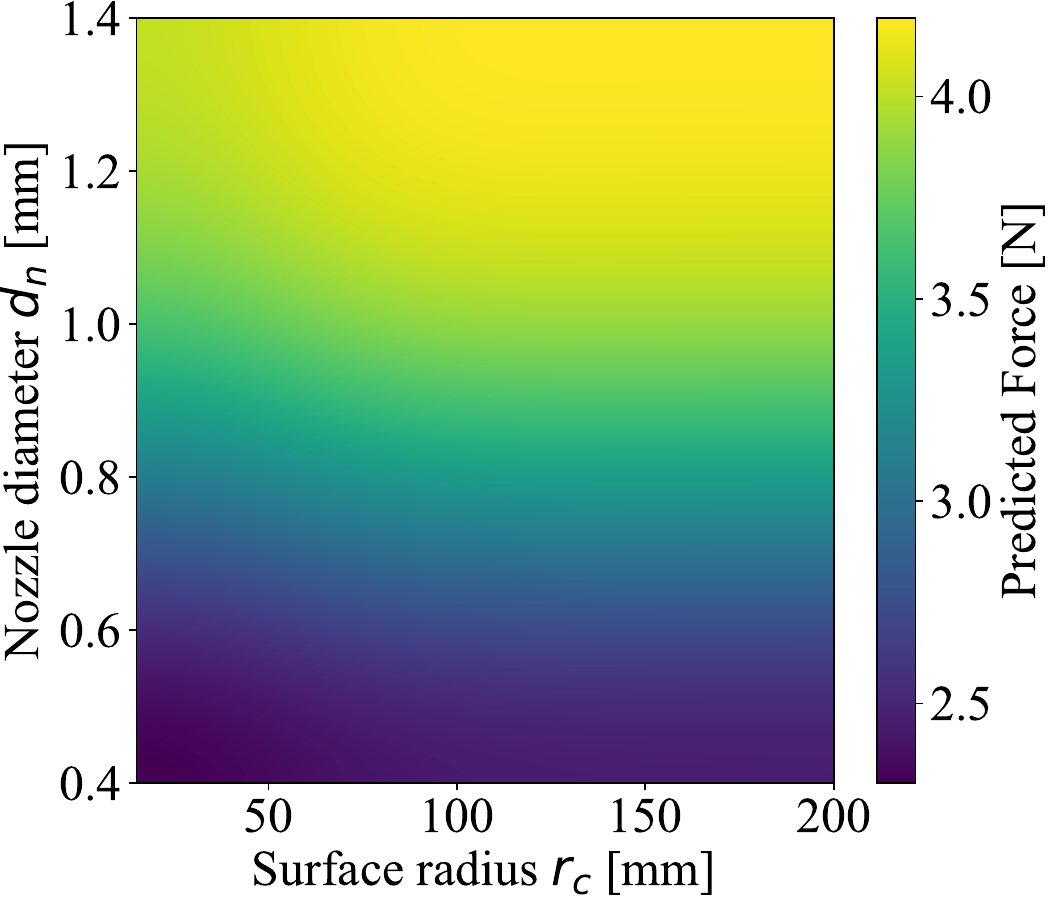}
         \caption{dome-concave \label{subfig12-3}}
	\end{subfigure}
	~
	\begin{subfigure}[b]{0.236\linewidth}
         \centering
         \includegraphics[width=\mywidth\linewidth, clip, trim=0pt 0pt 0pt 0pt]{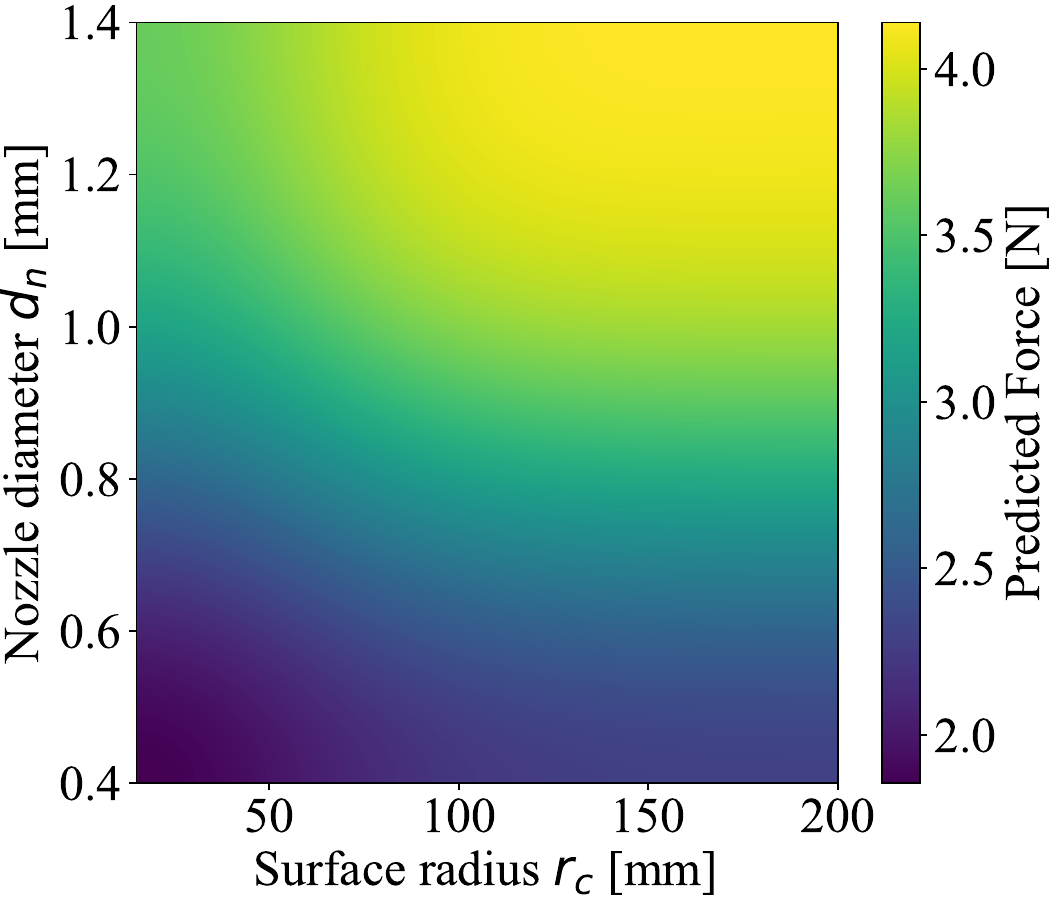}
         \caption{cylinder-concave \label{subfig12-4}}
	\end{subfigure}\\
 \caption{Machine learning prediction of the lifting force by changing the nozzle diameter, object roundness with supply pressure 400 kPa. \label{fig-12-1}}
\end{figure*}
%FFFFFFFFFFFFFFFFFFFFFFFFFFFFFFFFFFFFFFFFFFFFFF

%FFFFFFFFFFFFFFFFFFFFFFFFFFFFFFFFFFFFFFFFFFFFFF
\begin{figure*}[tb]
\newcommand{\mywidth}{1}
\centering
	\begin{subfigure}[b]{0.15\linewidth}
         \centering
         \includegraphics[width=\mywidth\linewidth, clip, trim=15pt 20pt 25pt 20pt]{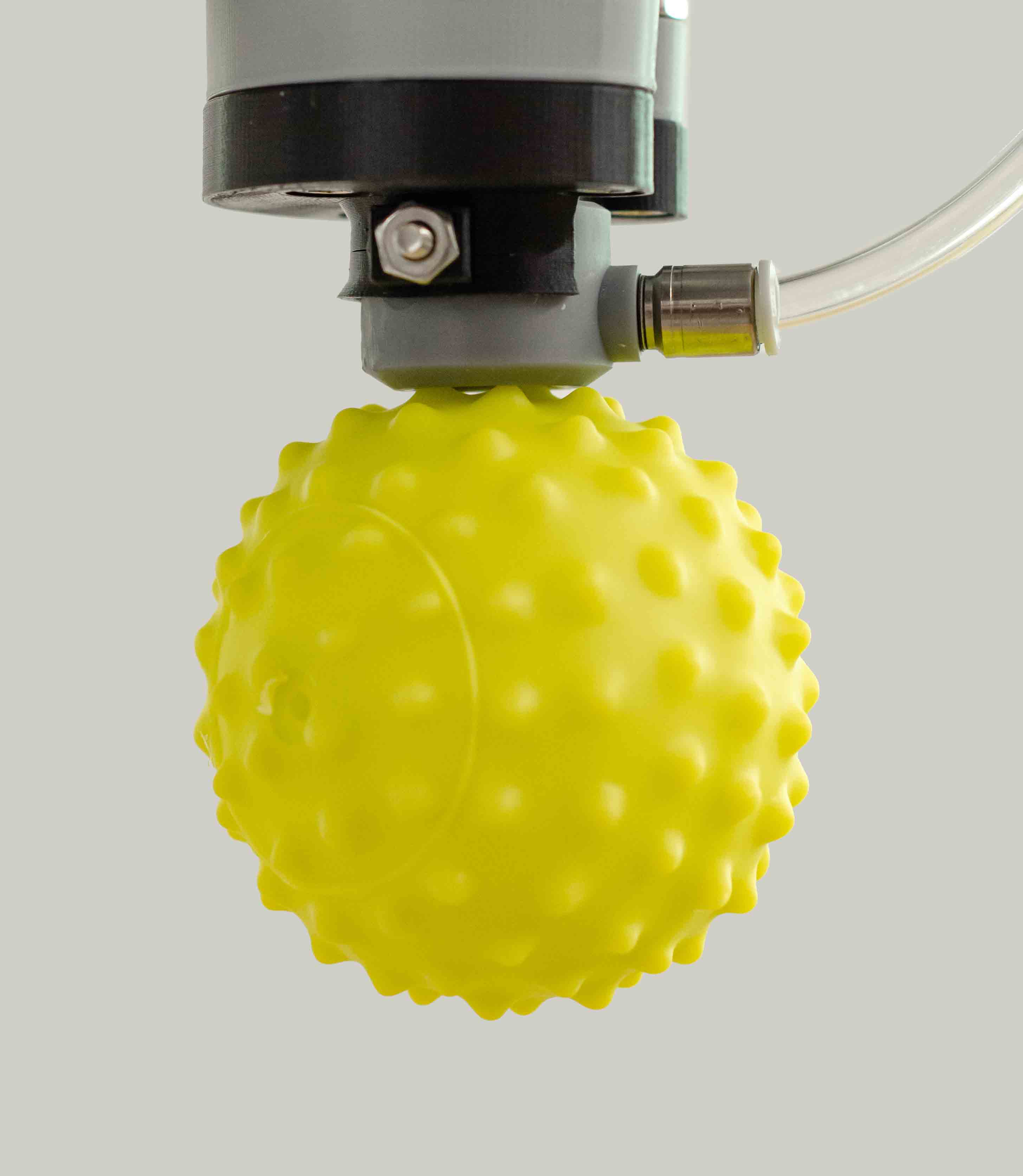}
        \caption{\label{subfig1}}
	\end{subfigure}
	~
	\begin{subfigure}[b]{0.15\linewidth}
         \centering
         \includegraphics[width=\mywidth\linewidth, clip, trim=20pt 20pt 20pt 20pt]{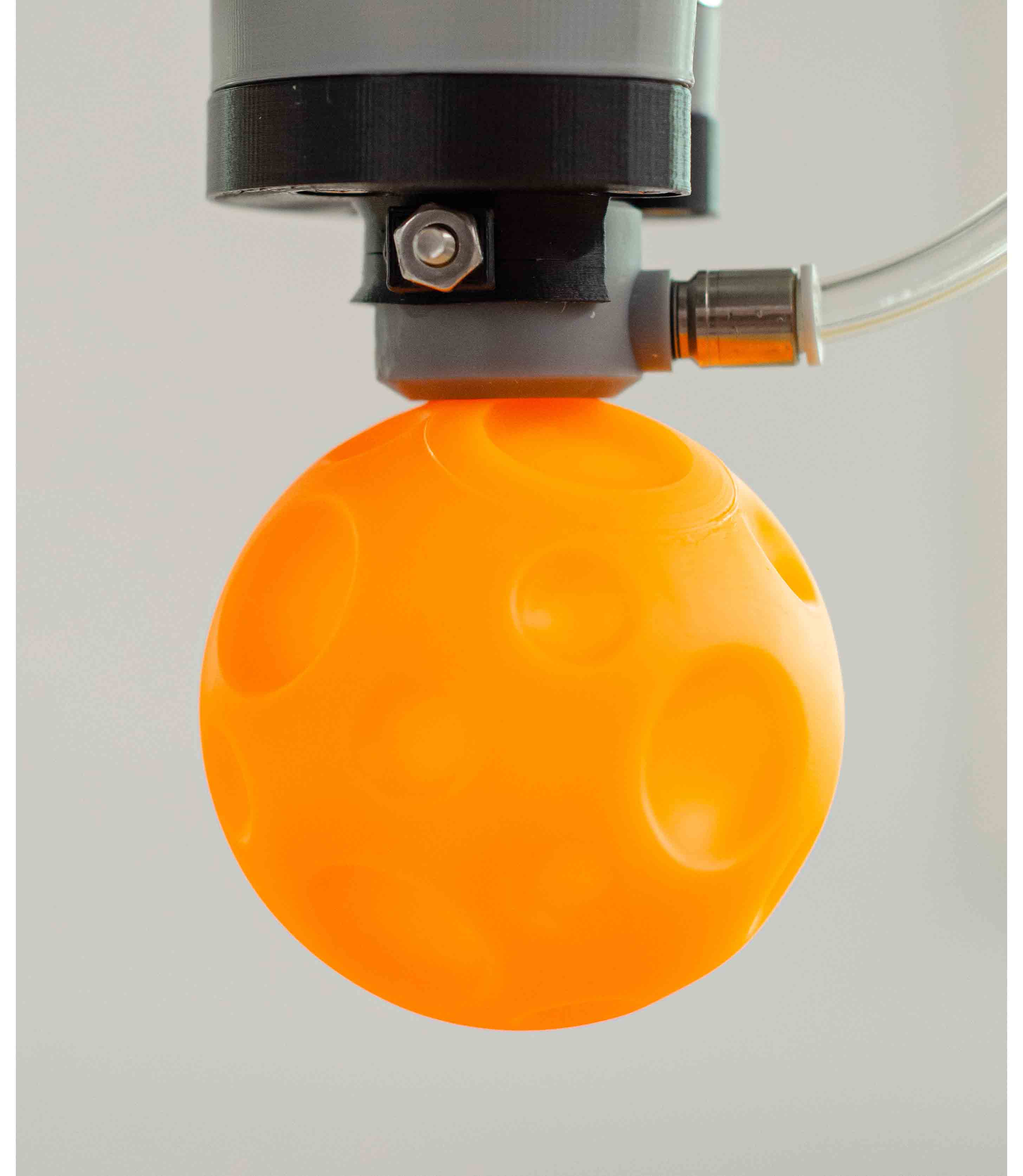}
         \caption{ \label{subfig2}}
	\end{subfigure}
        ~
	\begin{subfigure}[b]{0.15\linewidth}
         \centering
         \includegraphics[width=\mywidth\linewidth, clip, trim=25pt 35pt 25pt 17pt]{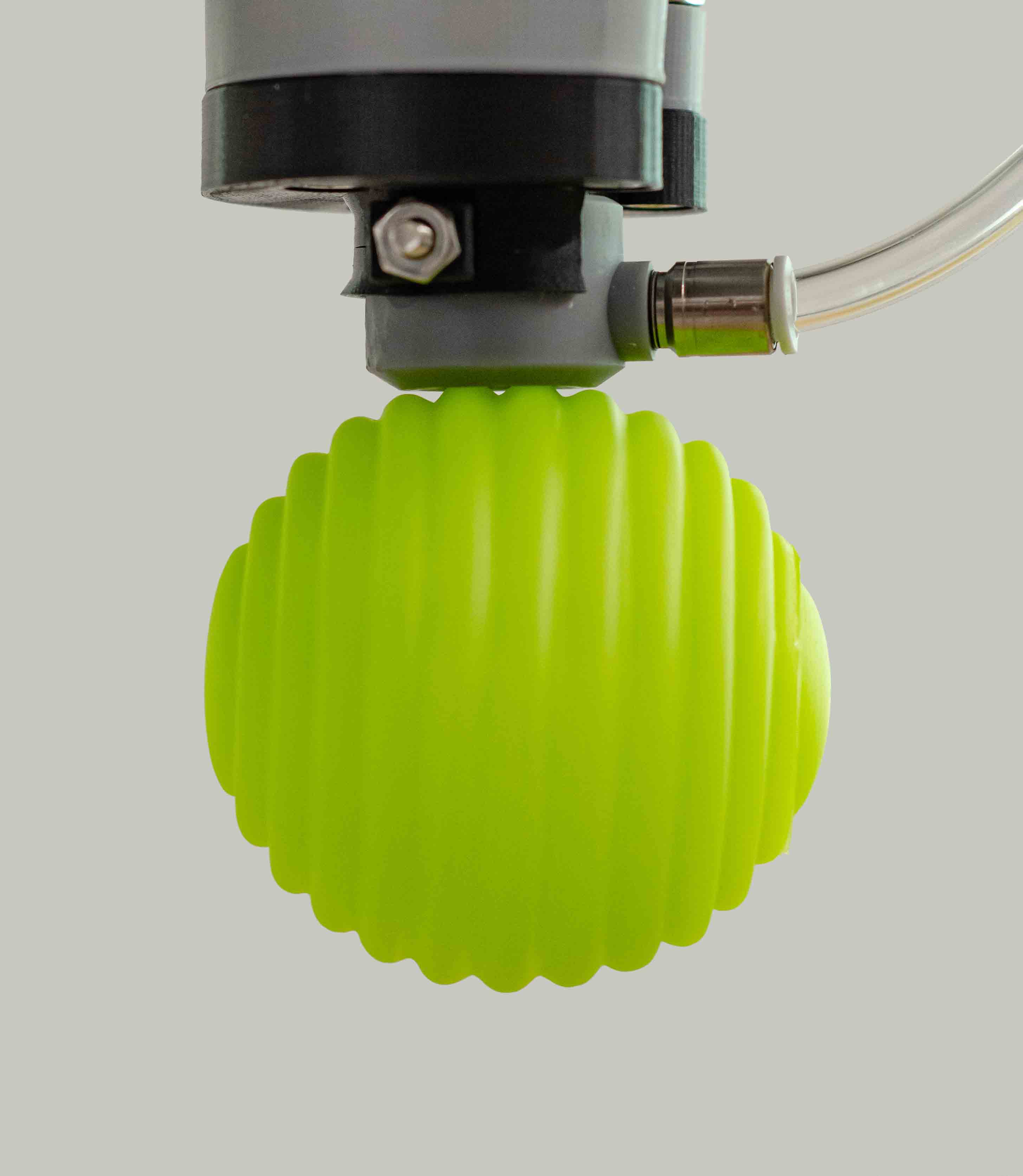}
         \caption{ \label{subfig3}}
	\end{subfigure}
	~
 	\begin{subfigure}[b]{0.15\linewidth}
         \centering
         \includegraphics[width=\mywidth\linewidth, clip, trim=15pt 20pt 25pt 20pt]{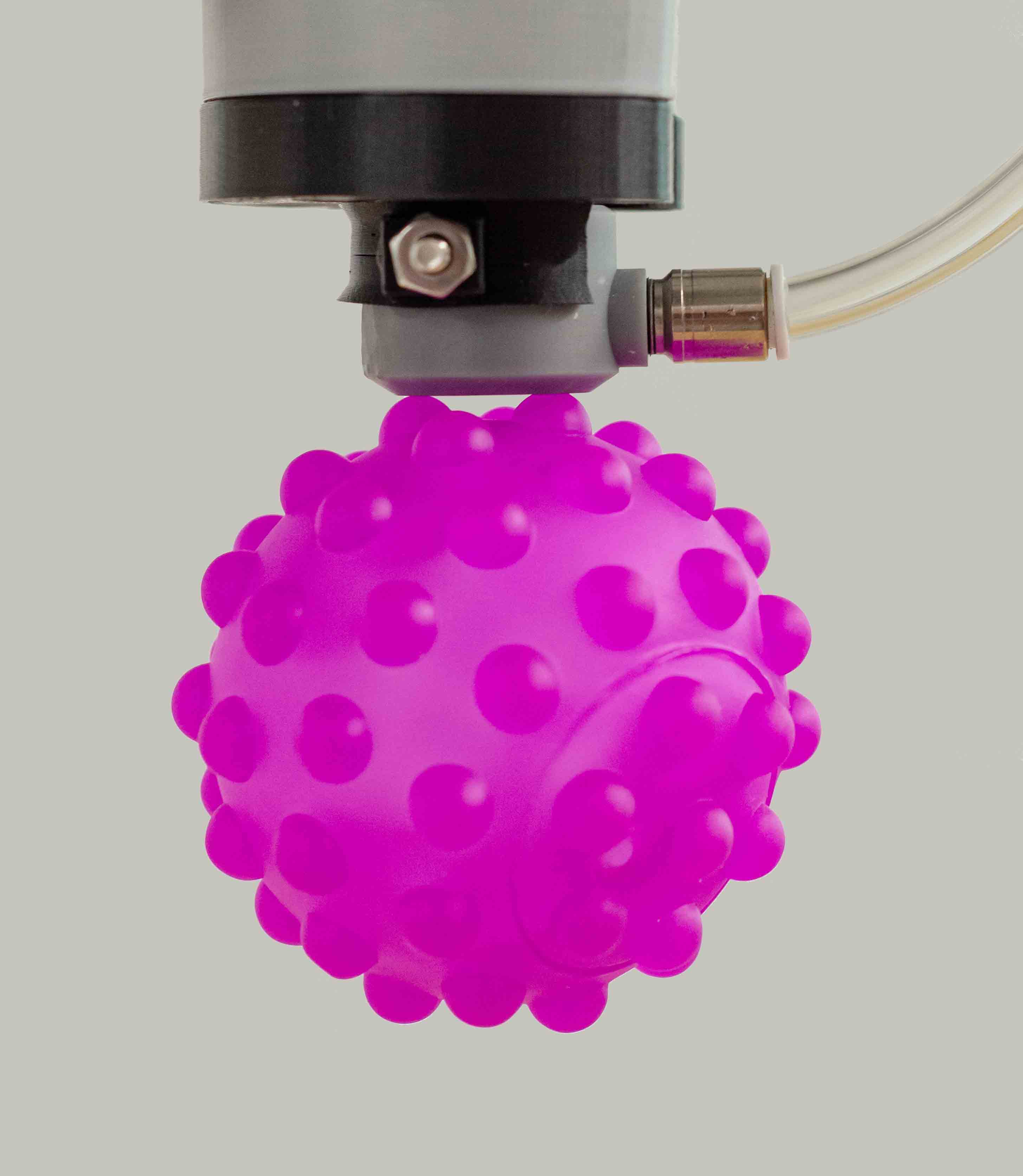}
         \caption{ \label{subfig4}}
	\end{subfigure}
	~
 	\begin{subfigure}[b]{0.15\linewidth}
         \centering
         \includegraphics[width=\mywidth\linewidth, clip, trim=25pt 35pt 25pt 17pt]{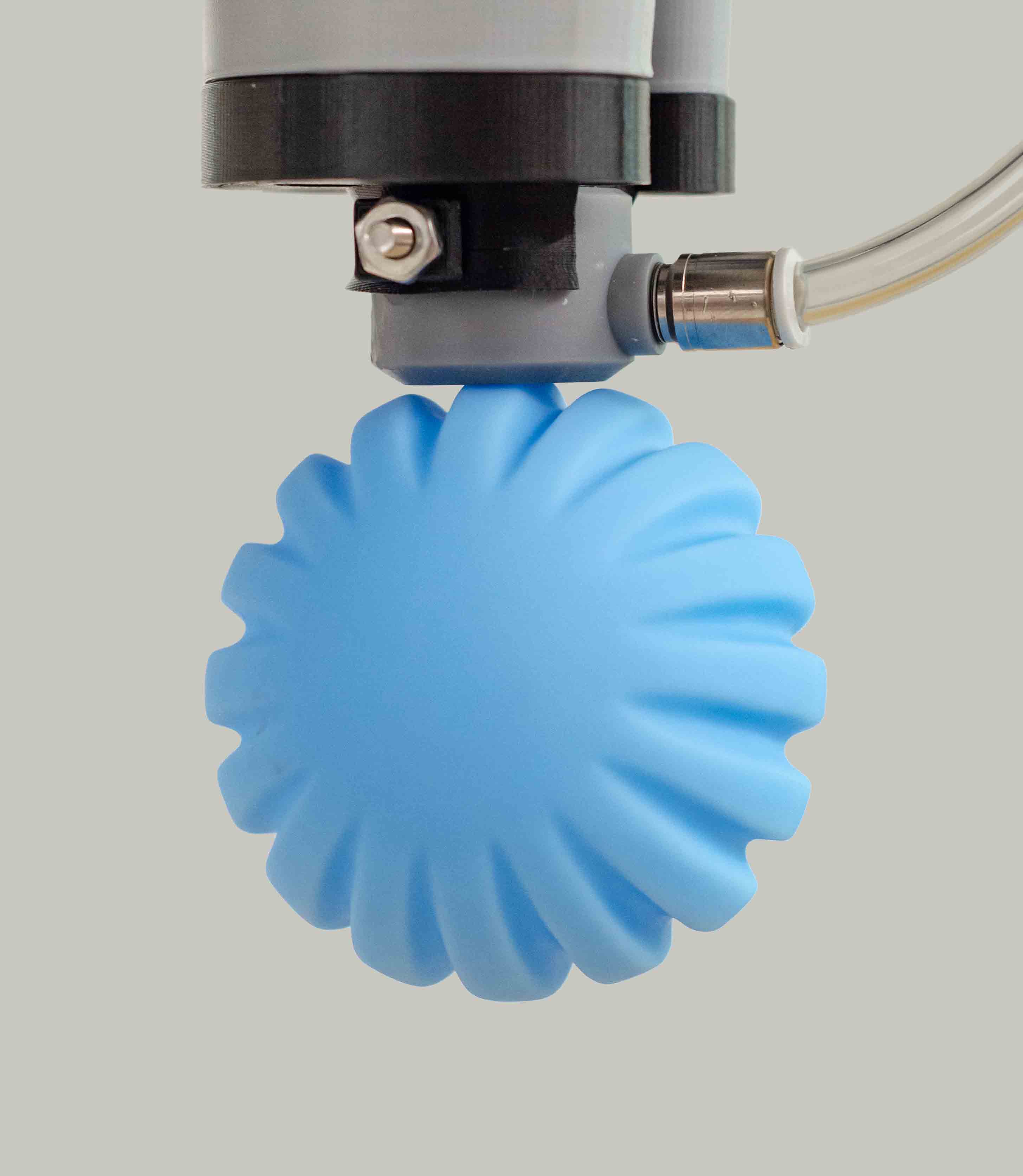}
         \caption{ \label{subfig5}}
	\end{subfigure}
	~
	\begin{subfigure}[b]{0.15\linewidth}
         \centering
         \includegraphics[width=\mywidth\linewidth, clip, trim=30pt 48pt 40pt 28pt]{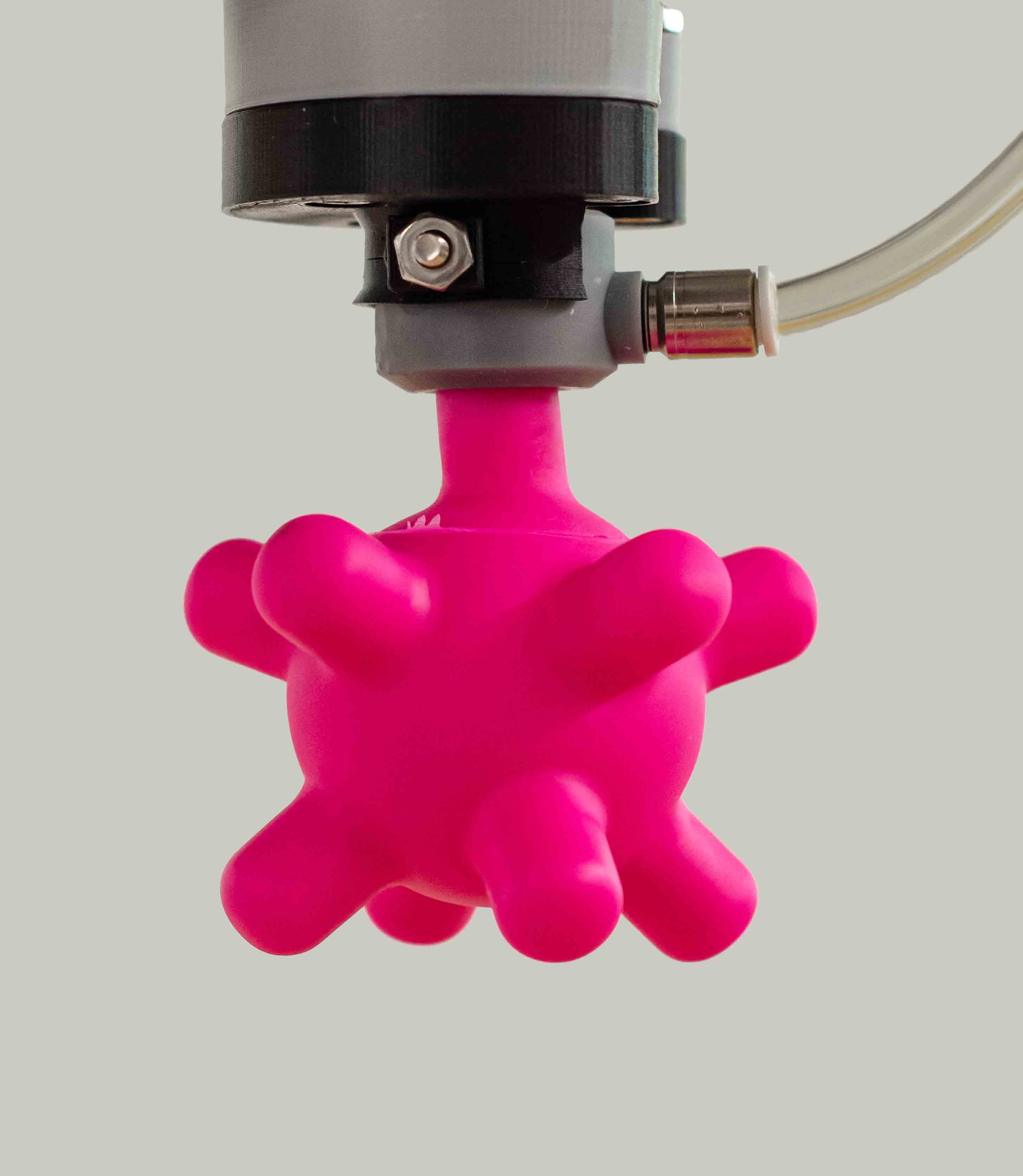}
         \caption{ \label{subfig6}}
	\end{subfigure}\\
 \caption{Grasping soft balls of various shapes using a vortex gripper with nozzle diameter $d_n=1$ mm and supply pressure of 200 kPa.\label{fig-12}}
\end{figure*}
%FFFFFFFFFFFFFFFFFFFFFFFFFFFFFFFFFFFFFFFFFFFFFF

The obtained general results for all vortex grippers and surfaces demonstrate the loss of lifting force due to the formation of a non-uniform gap between the surface of the soft object and the gripper (Fig.~\ref{fig-10-1}). An uneven gap between the surface of the soft object and the gripper is formed for cylindrical (Fig.~\ref{fig-10-1}b, f, j, n, d, h, l, p) or non-symmetrical surfaces. The obtained dataset allows us to perform machine learning to predict the performance of the gripper. We have added the obvious data to the dataset, that in case of an absence of supply pressure, there is no lifting force, and for a flat surface, we assume that the radius of rounding of the surface is $1 \cdot 10^6$ mm. To perform machine learning, we used a combined approach of RandomForest and AdaBoost regressor. Nozzle radius, object surface roundness, gripper supply pressure, and surface type were used as model factors. We divided the dataset into two parts where 80\% is the training dataset and 20\% is the testing dataset. We used a five-fold cross-validation of the training results to verify the prediction accuracy. In this case, the 20\% testing dataset is changed each time to cover the entire dataset 5 times. This allows us to make sure that the model is reliable and not sensitive to any particular part of the dataset. In the end, we get the combined average prediction of the two models: 

\begin{equation}
\label{deqn_ex5}
F_{\text{pred}} = \frac{F_{\text{pred\_RF}} + F_{\text{pred\_AB}}}{2},
\end{equation}

\noindent where $F_{\text{pred\_RF}}$ and $F_{\text{pred\_AB}}$ are the predictions from the Random Forest and AdaBoost models.

The proposed approach allowed us to combine the stability and robustness of RandomForest with the boosting capabilities of AdaBoost, and to generalize the reliability of the results obtained by cross-validation. As a result, we were able to obtain predicted gripper force data for a supply pressure of 400 kPa in an extended range of nozzle diameters and object roundness with a mean cross-validation score of 0.9461 (Fig.~\ref{fig-12-1}).  From the obtained results (Fig.~\ref{fig-12-1}) we see that the lifting force increases significantly even for cylindrical surfaces when the diameter of the nozzle $d_n$ of the vortex gripping device increases. As already discussed above, this is due to the increase in the mass flow of compressed air, which helps to interact with the soft surface at a great distance from the gripper. 

To demonstrate a significant increase in the ability to grasp soft surfaces of complex shapes, we grasping toys (soft balls) with different geometries (Fig.~\ref{fig-12}): a - small bumps, b - spherical conclave's, c - shallow waves, d - large bumps, e - waves of great depth, f - cylindrical bulges. Four soft balls (Fig.~\ref{subfig2}) can be grasped very easily in any position or orientation, which is a big challenge for other pneumatic grippers. From the point of view of the grasping process, the ball with spherical conclaves (Fig.~\ref{subfig2}) and the ball with cylindrical bulges (Fig.~\ref{subfig6}) turned out to be the most difficult. A ball with cylindrical bulges can be gripped using the developed vortex gripper in two ways: by the end of the cylinder (Fig.~\ref{subfig6}) and by the spherical part of the ball. Such a ball cannot be grabbed by a cylindrical surface, since with this position of the ball, the center of mass will move it to the side, which will cause the formation of a torque around the points of contact of the ball with the gripper and its subsequent fall. A ball with spherical concaves (Fig.~\ref{subfig2}) has the ability to scroll if the gripper is not directly above the concave or between the concaves. When the spherical concave is halfway under the gripper, a greater lifting force is generated in the zone without the concave, which scrolls the ball. However, this can be easily fixed by adding additional frictional elements to the structure, which will be the subject of further research.

%-------------------------------------------------
\subsection{Ex vivo and Bio-Inspired Evaluation}
%-------------------------------------------------

\begin{figure*}[t]
\centering
\subfloat[Grasping]{\includegraphics[width=0.165\linewidth,clip ,trim=0pt 0pt 0pt 120pt]{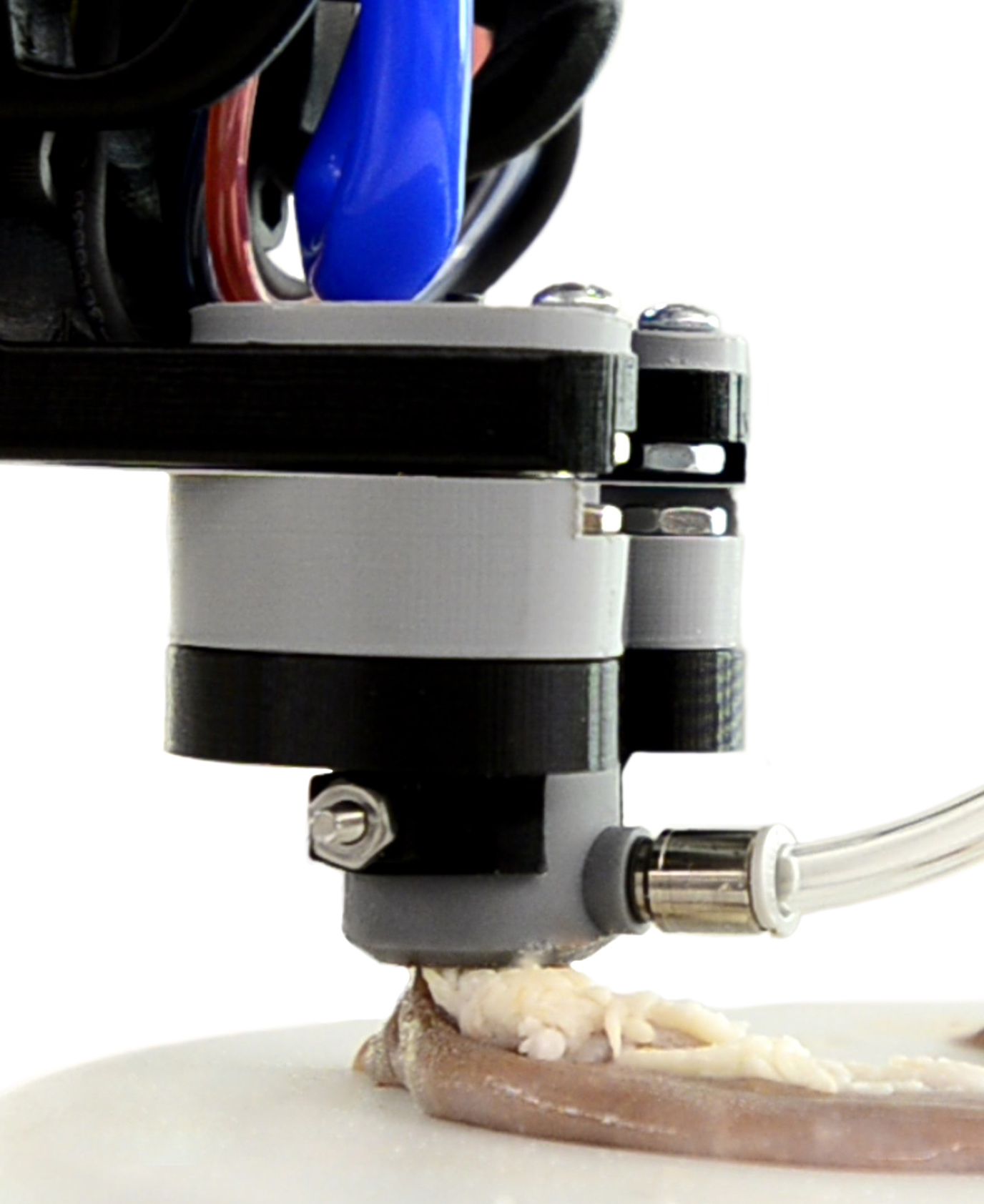}%
\label{fig13_1}}
\hfil
\subfloat[Lifting]{\includegraphics[width=0.165\linewidth,clip ,trim=0pt 0pt 0pt 120pt]{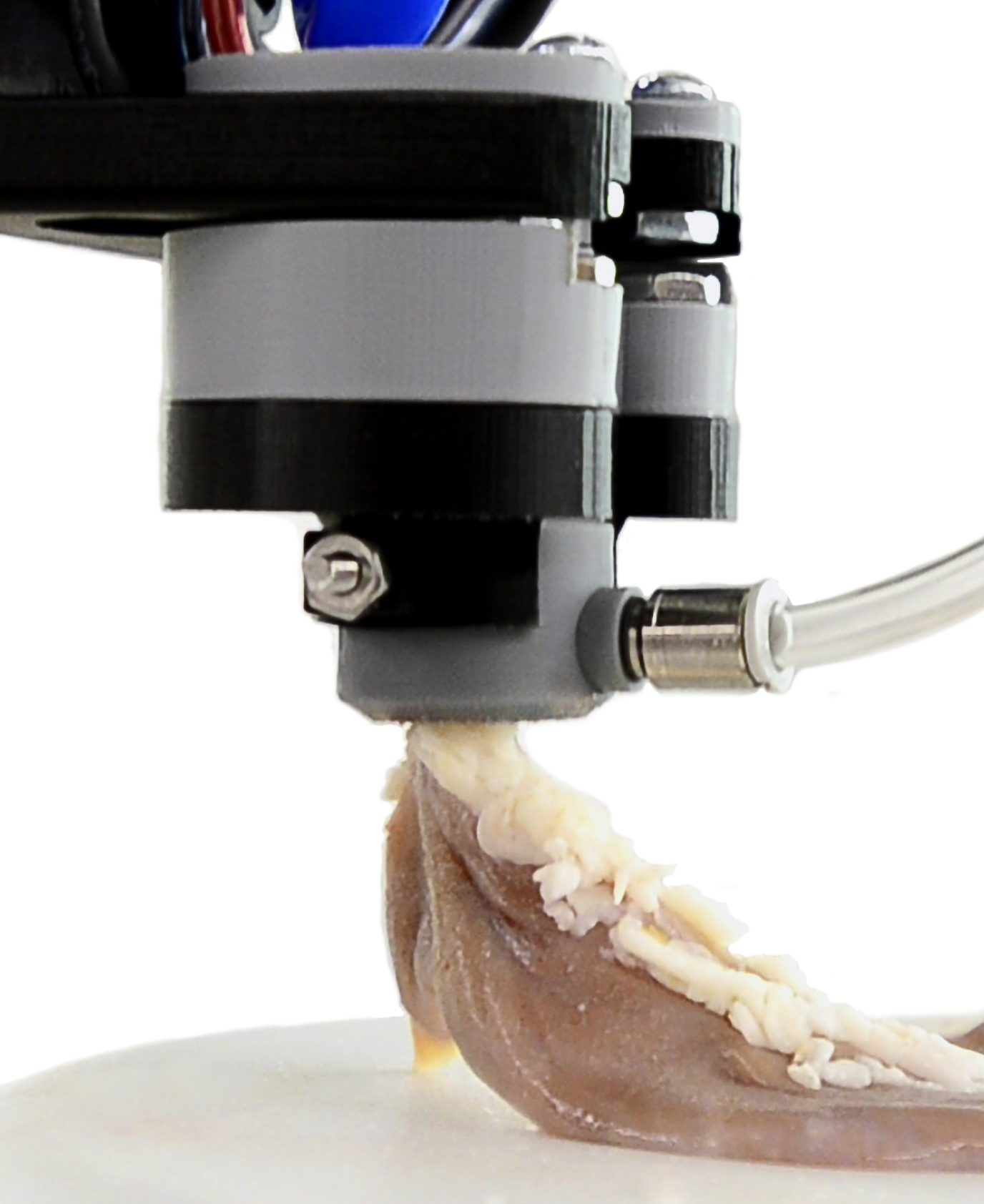}%
\label{fig13_2}}
\hfil
\subfloat[Releasing]{\includegraphics[width=0.165\linewidth,clip ,trim=0pt 0pt 0pt 120pt]{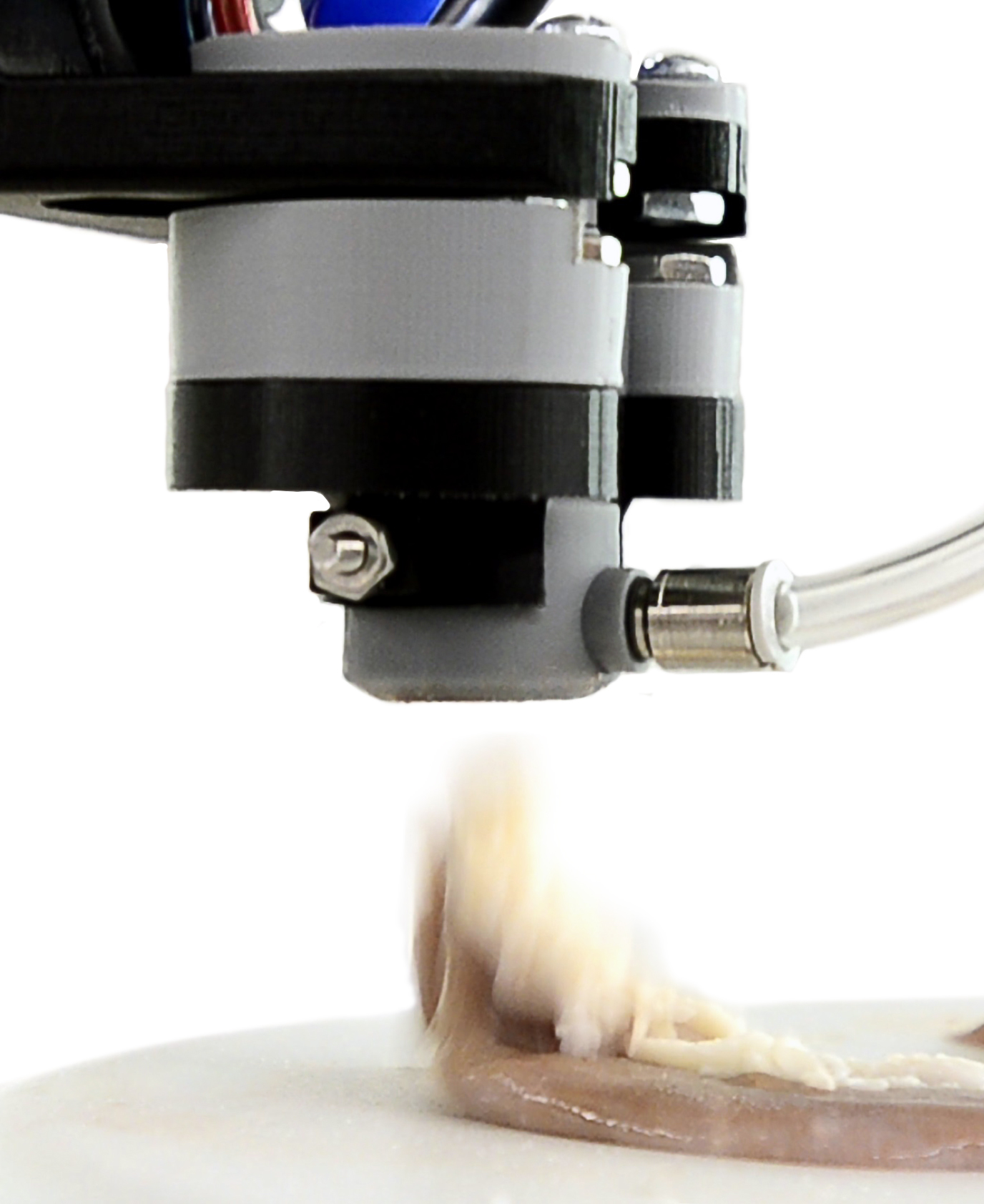}%
\label{fig13_3}}
\hfil
\subfloat[Grasping]{\includegraphics[width=0.165\linewidth,clip ,trim=0pt 0pt 0pt 120pt]{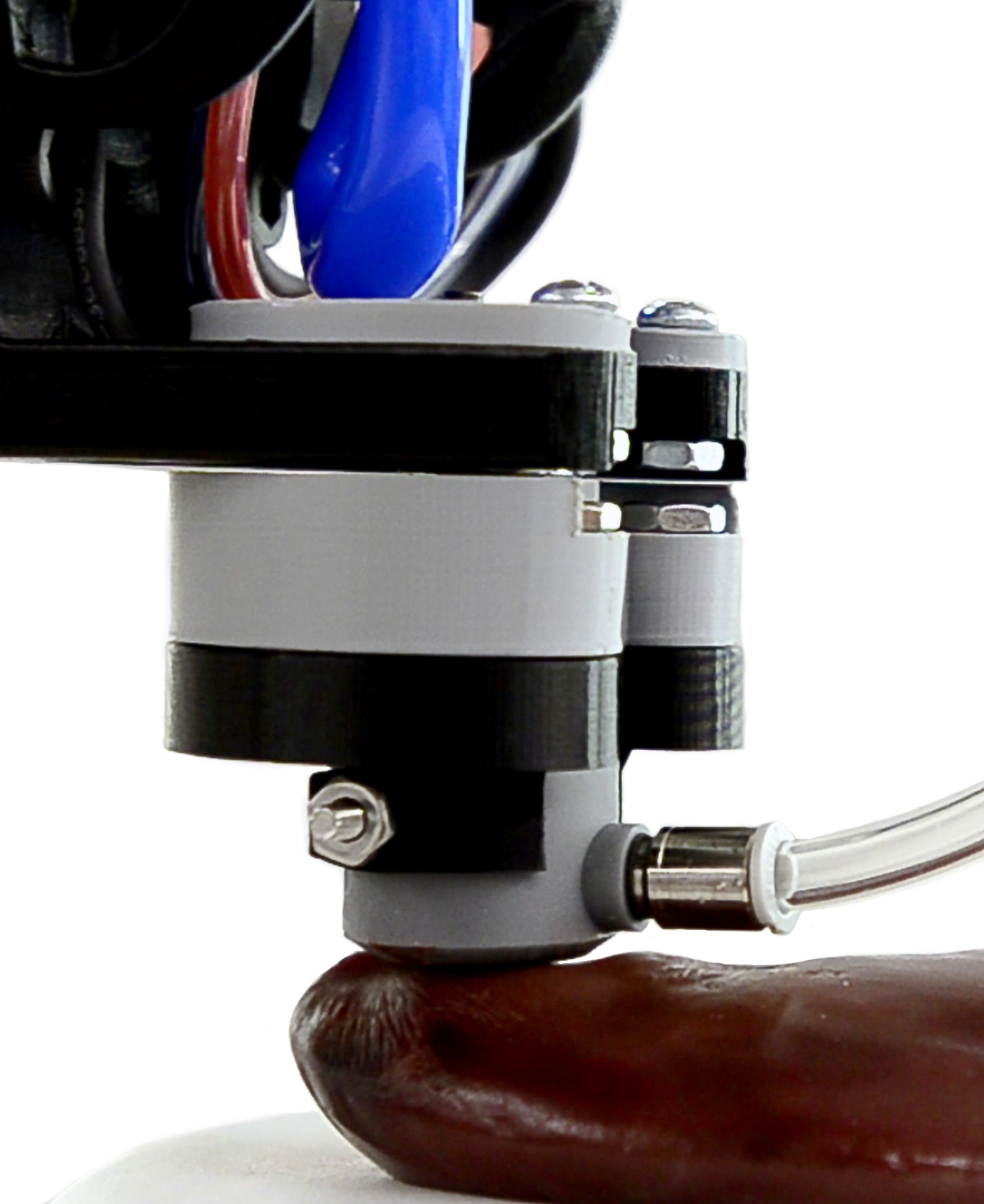}%
\label{fig13_4}}
\hfil
\subfloat[Lifting]{\includegraphics[width=0.165\linewidth,clip ,trim=0pt 0pt 0pt 120pt]{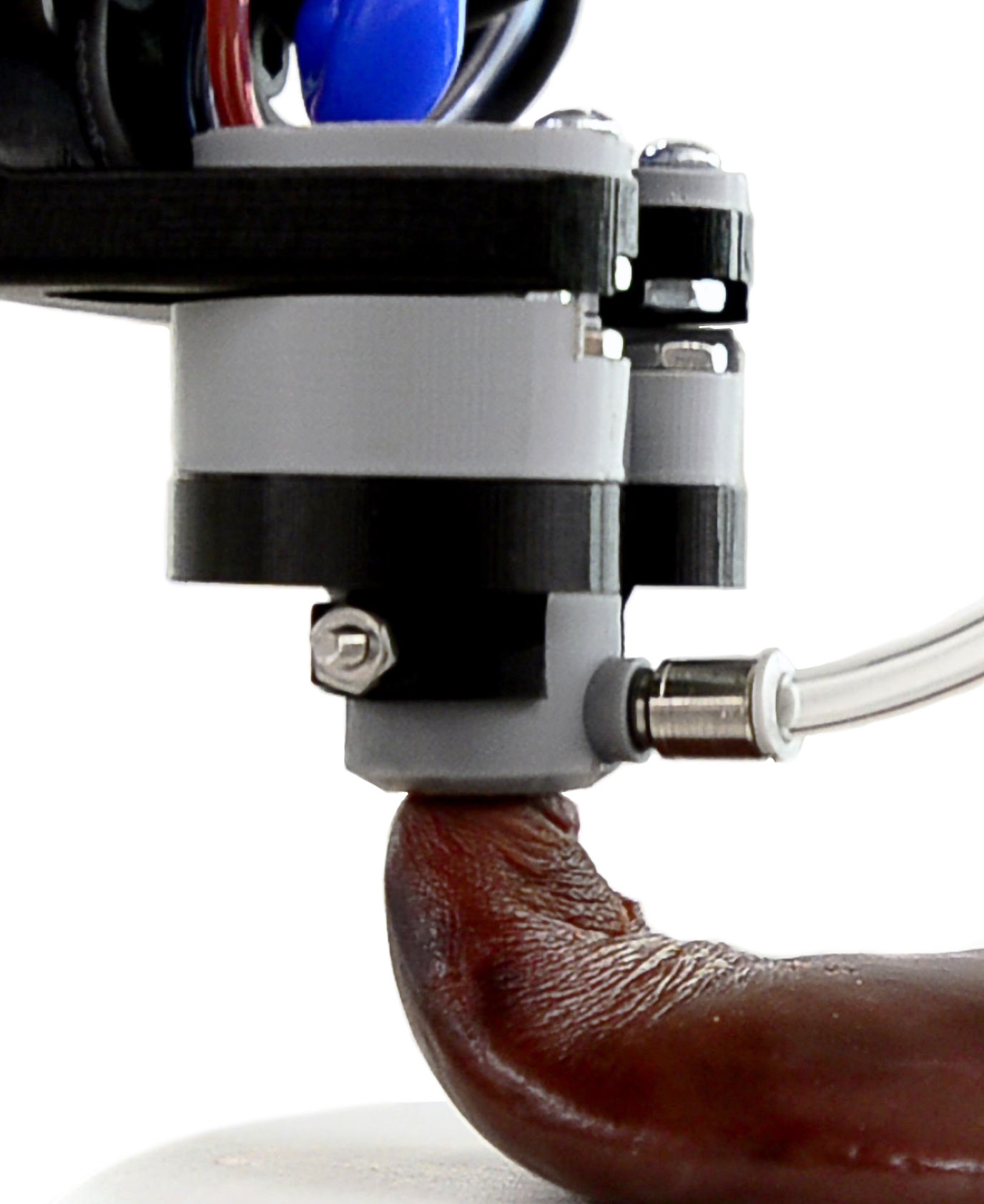}%
\label{fig13_5}}
\hfil
\subfloat[Releasing]{\includegraphics[width=0.165\linewidth,clip ,trim=0pt 0pt 0pt 120pt]{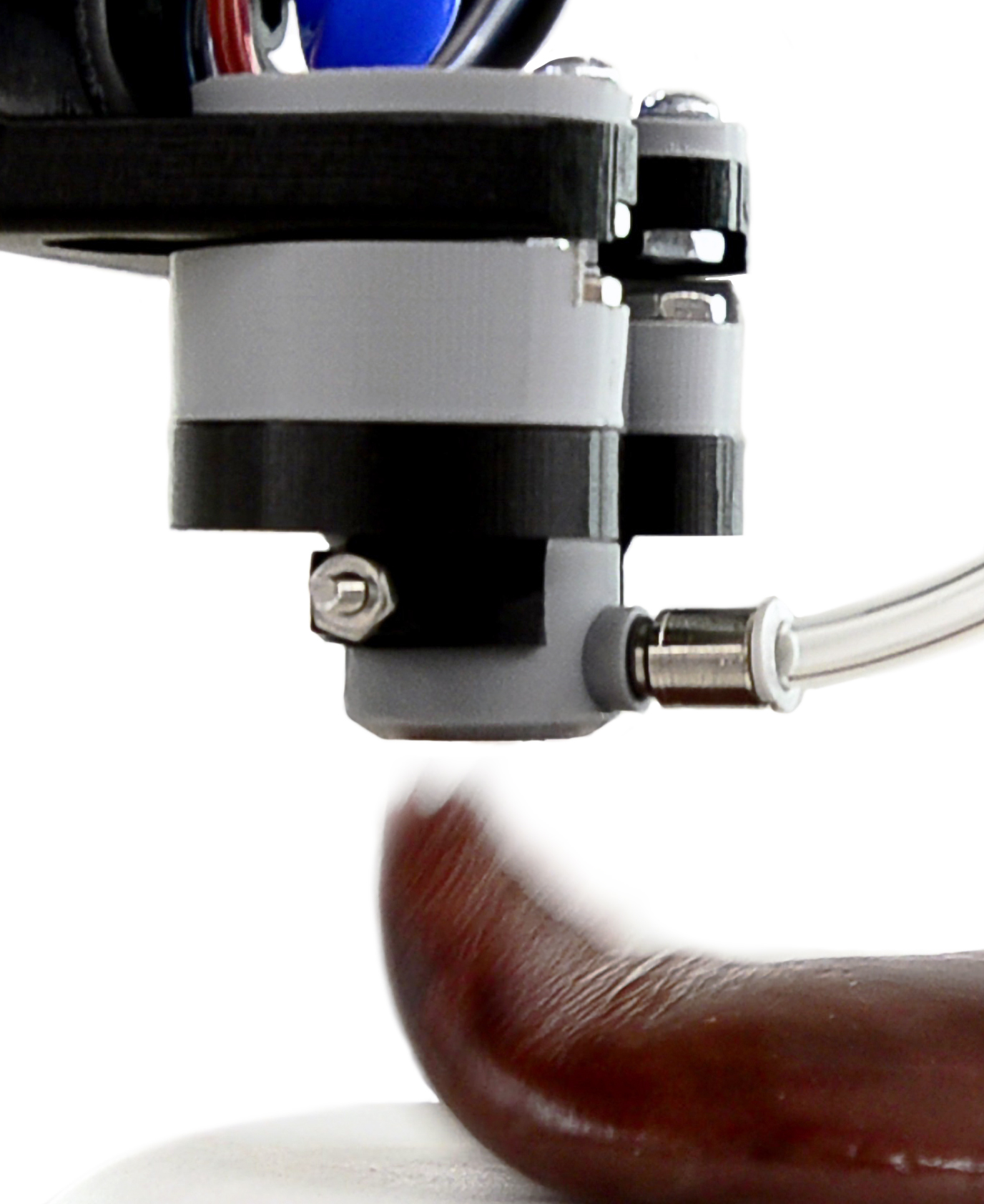}%
\label{fig13_6}}
\hfil
\caption{$Ex$ $vivo$ experiment of grasping biological soft tissues using a gripper G$_3$: (a, b, c) - bovine intestine, (d, e, f) - porcine kidney.}
\label{fig-13}
\end{figure*}

%FFFFFFFFFFFFFFFFFFFFFFFFFFFFFFFFFFFFFFFFFFFFFF
\begin{figure*}[t]
\newcommand{\mywidth}{1}
\centering
	\begin{subfigure}[b]{0.23\linewidth}
         \centering
         \includegraphics[width=\mywidth\linewidth, clip, trim=0pt 110pt 0pt 860pt]{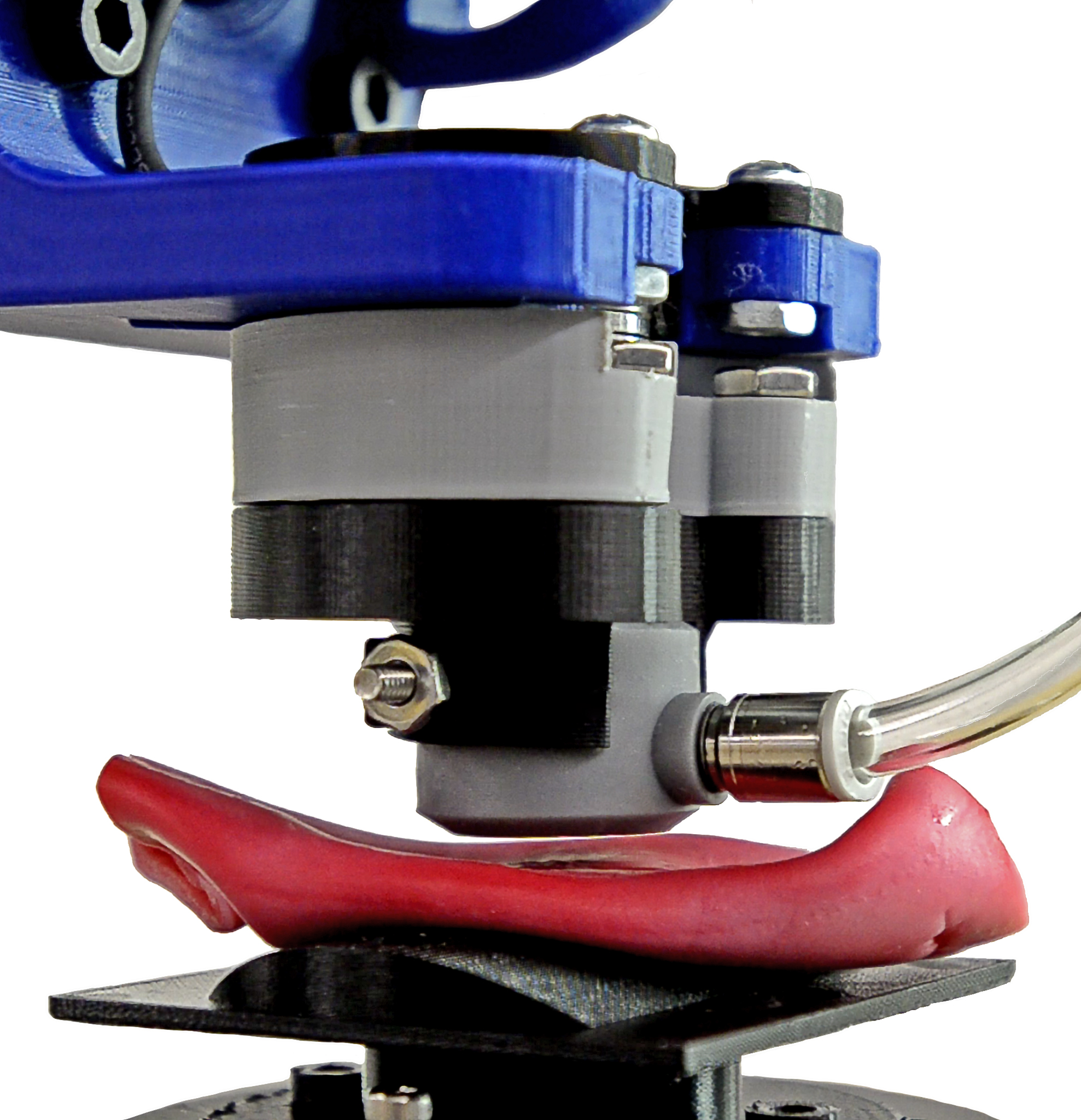}
        \caption{}\label{subfig1}
	\end{subfigure}
	~
	\begin{subfigure}[b]{0.23\linewidth}
         \centering
         \includegraphics[width=\mywidth\linewidth, clip, trim=0pt 110pt 0pt 860pt]{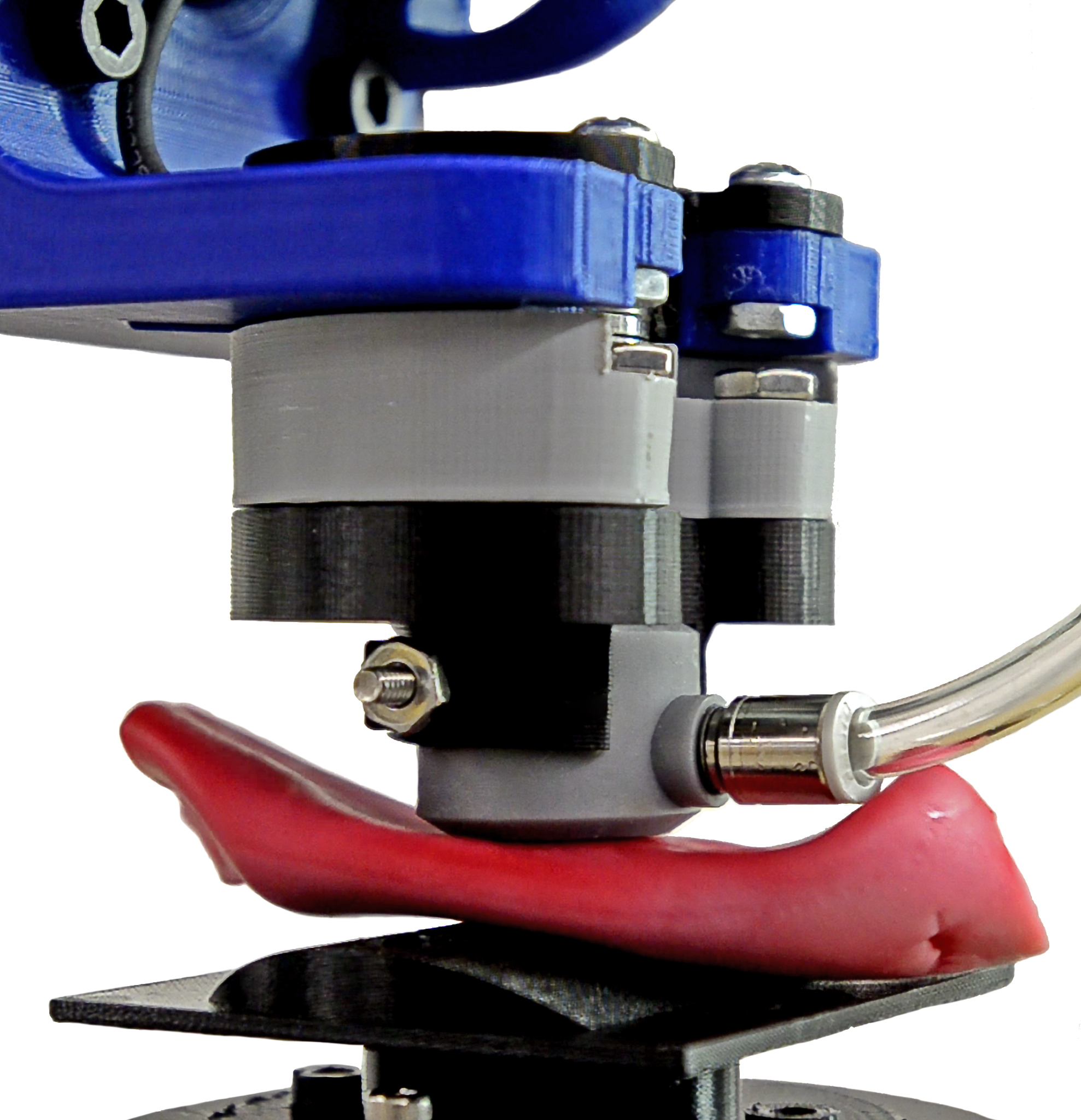}
         \caption{} \label{subfig2}
	\end{subfigure}
        ~
	\begin{subfigure}[b]{0.23\linewidth}
         \centering
         \includegraphics[width=\mywidth\linewidth, clip, trim=0pt 110pt 0pt 860pt]{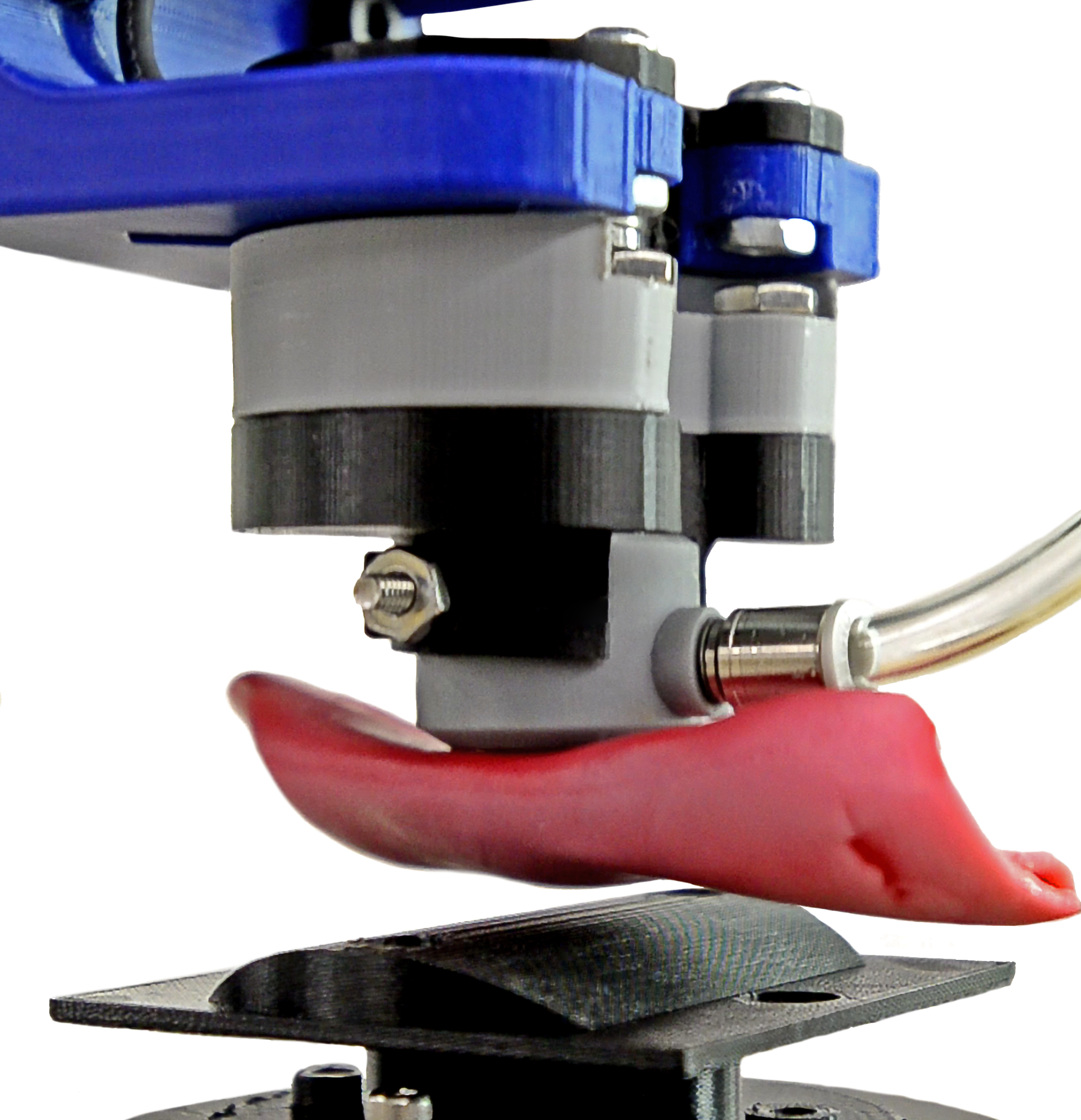}
         \caption{} \label{subfig3}
	\end{subfigure}
	~
 	\begin{subfigure}[b]{0.23\linewidth}
         \centering
         \includegraphics[width=\mywidth\linewidth, clip, trim=0pt 110pt 0pt 860pt]{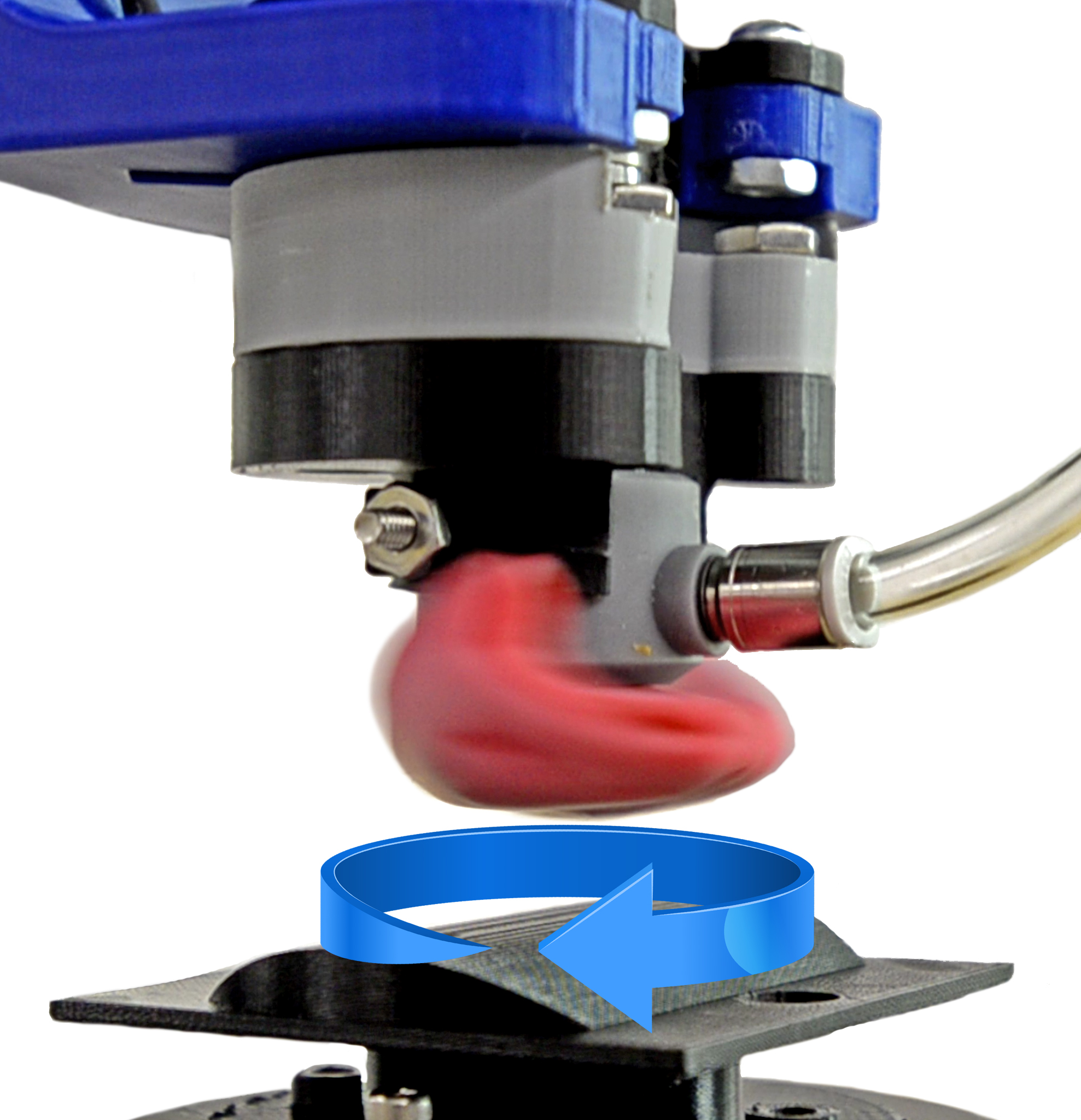}
         \caption{} \label{subfig4}
	\end{subfigure}\\ 
        ~ \\
 %%%%%%%%%%%%%%%%%%%%%%%%%%%%%%%%
	~
 	\begin{subfigure}[b]{0.23\linewidth}
         \centering
         \vspace{-2mm}
         \includegraphics[width=\mywidth\linewidth, clip, trim=0pt 110pt 0pt 860pt]{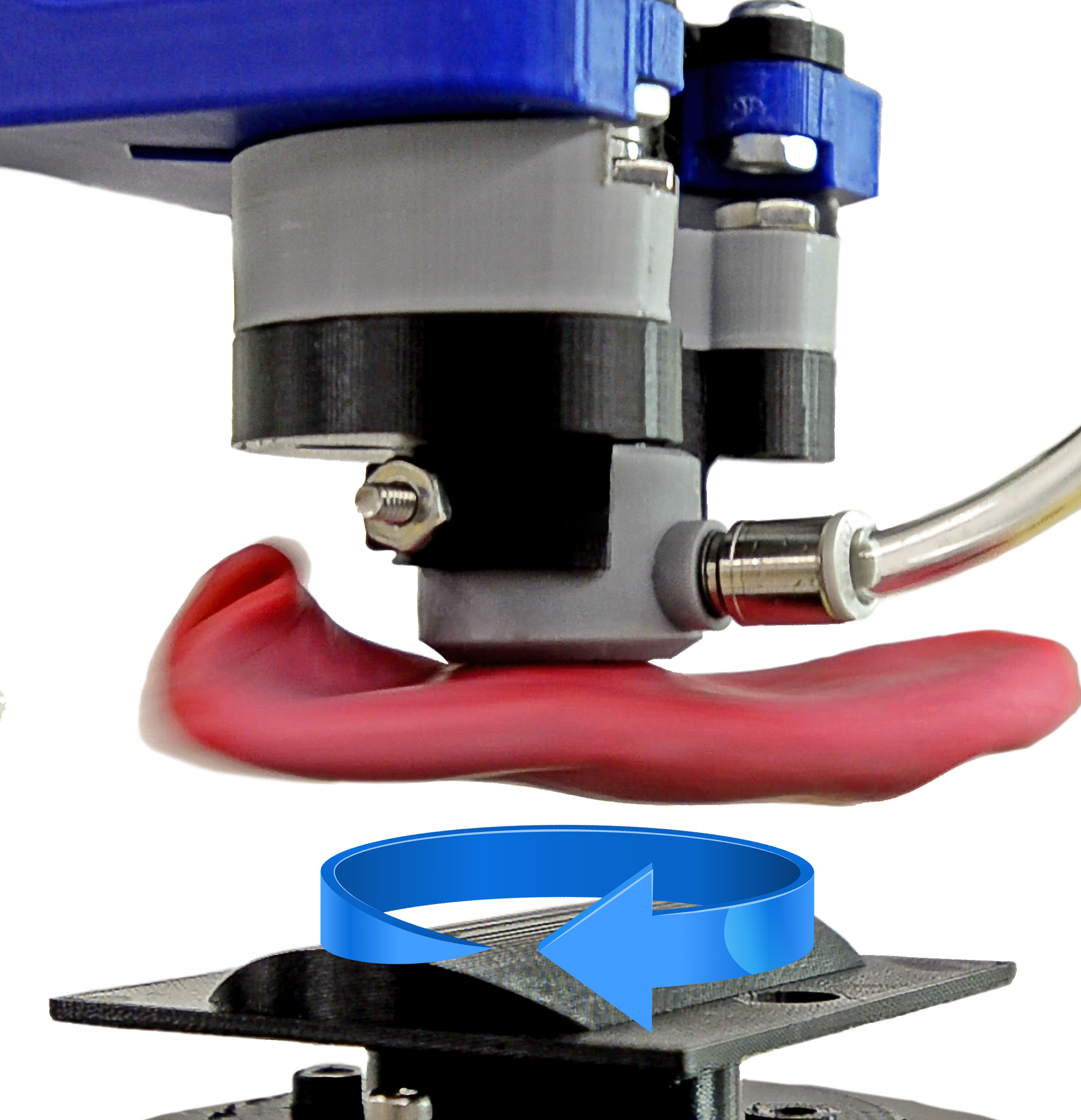}
         \caption{} \label{subfig5}
	\end{subfigure}
        ~
 	\begin{subfigure}[b]{0.23\linewidth}
         \centering
         \vspace{-2mm}
         \includegraphics[width=\mywidth\linewidth, clip, trim=0pt 110pt 0pt 860pt]{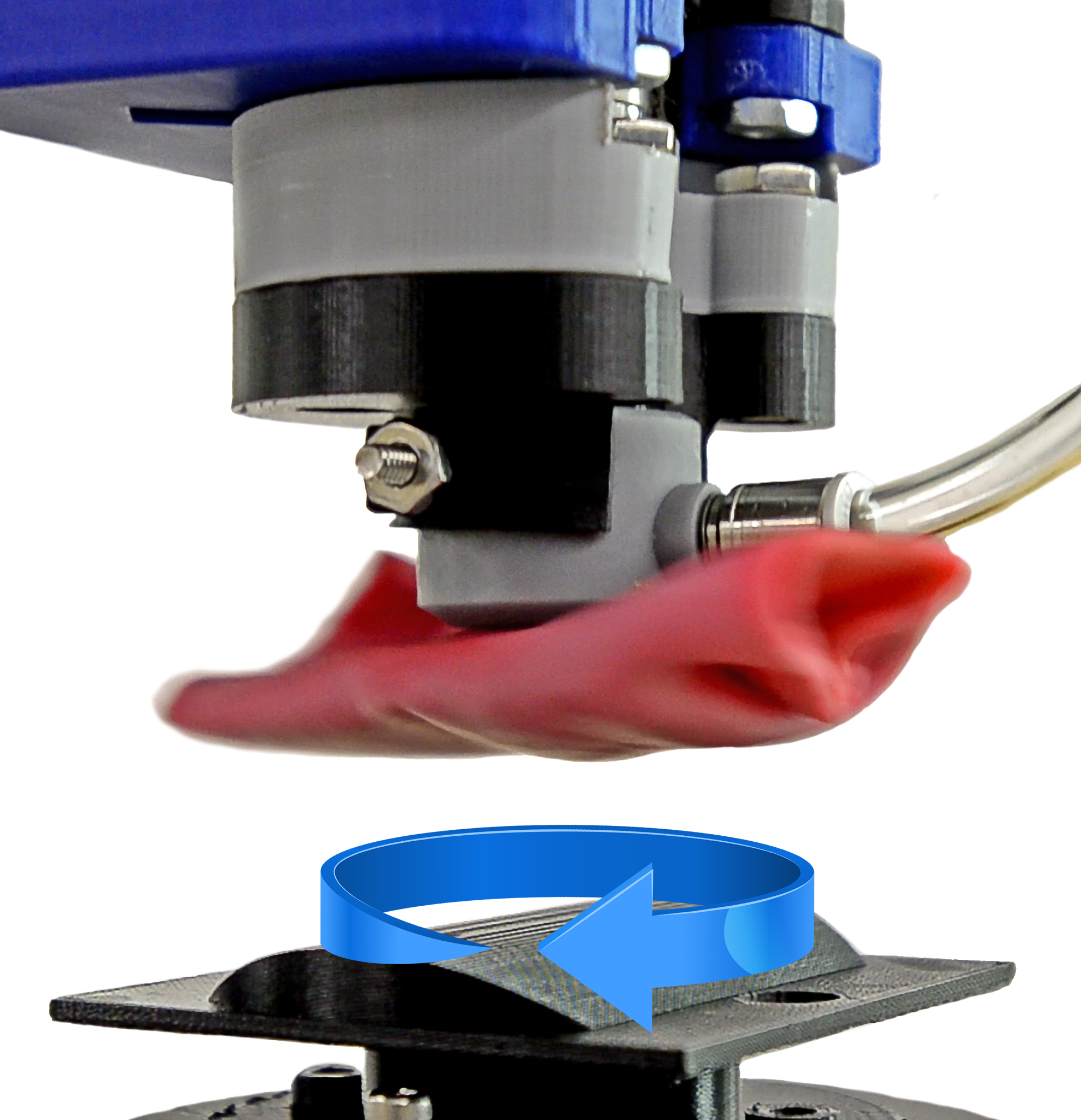}
         \caption{} \label{subfig5}
	\end{subfigure}
         ~
 	\begin{subfigure}[b]{0.23\linewidth}
         \centering
         \vspace{-2mm}
         \includegraphics[width=\mywidth\linewidth, clip, trim=0pt 110pt 0pt 860pt]{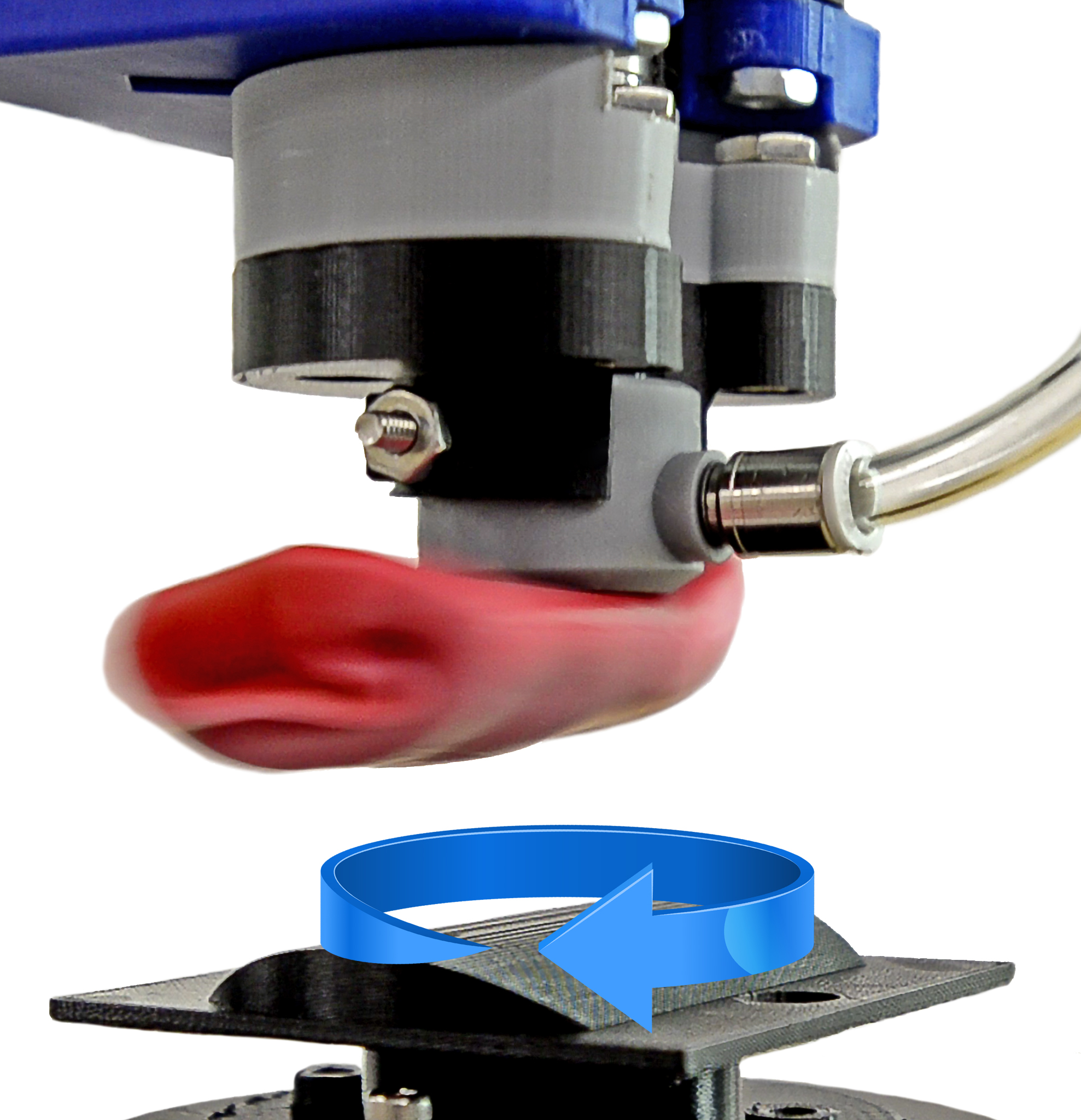}
         \caption{} \label{subfig5}
	\end{subfigure}
	~
	\begin{subfigure}[b]{0.23\linewidth}
         \centering
         \vspace{-2mm}
         \includegraphics[width=\mywidth\linewidth, clip, trim=0pt 110pt 0pt 860pt]{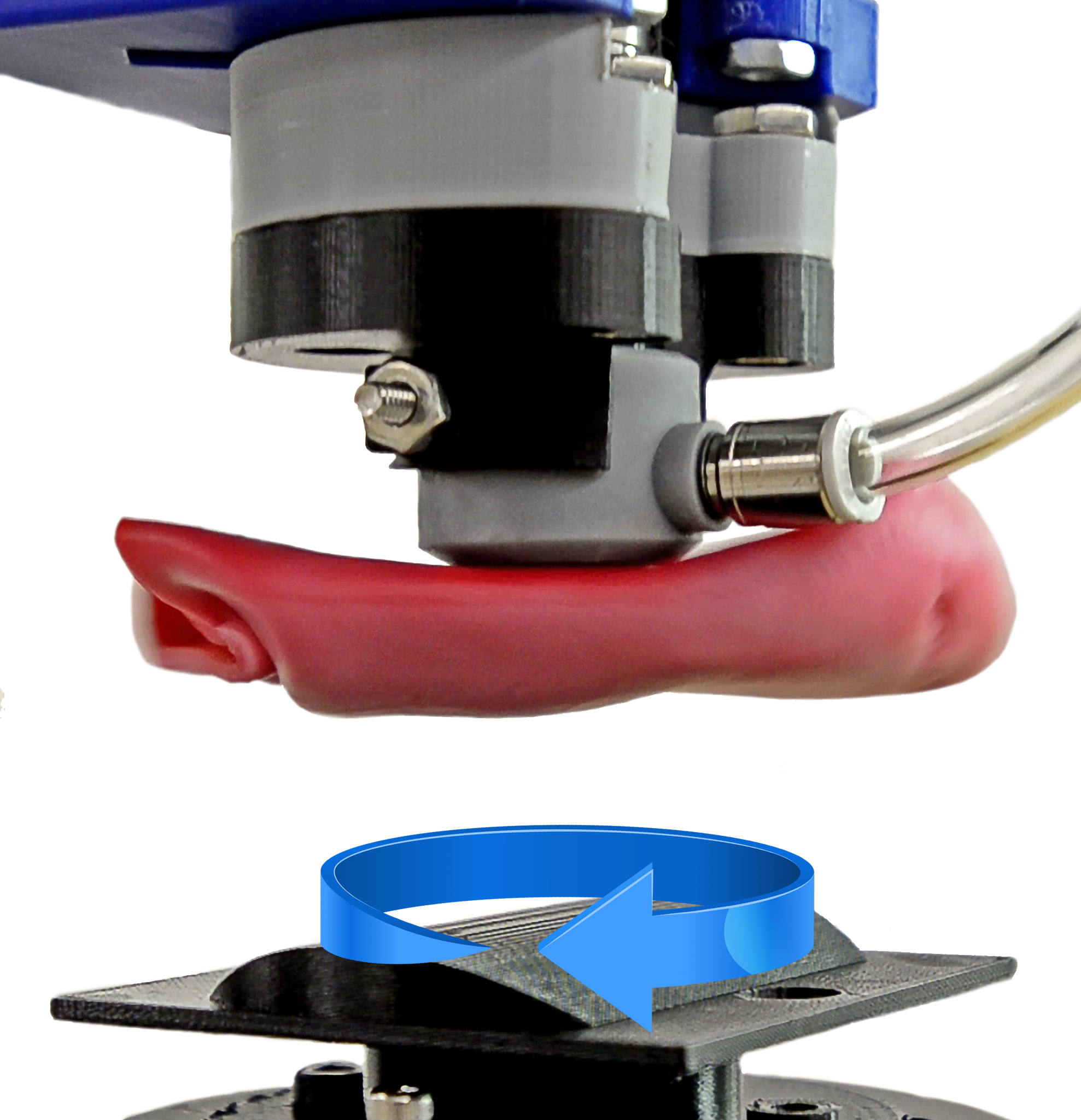}
         \caption{} \label{subfig6}
	\end{subfigure}\\
 \caption{Grasping (a-b) and lifting (c-h) bio-inspired intestines, which led to their rotation under supply pressure of 200 kPa.}\label{fig-14}
\end{figure*}
%FFFFFFFFFFFFFFFFFFFFFFFFFFFFFFFFFFFFFFFFFFFFFF

An $ex$ $vivo$ experiment was conducted on the grasping and lifting of two organs by the G$_3$ vortex gripper: bovine intestines (Fig.~\ref{fig13_1},~\ref{fig13_2},~\ref{fig13_3}) and porcine kidney (Fig.~\ref{fig13_4},~\ref{fig13_5},~\ref{fig13_6}). In Fig.~\ref{fig-13}, you can observe the stages of grasping, lifting and sliding, releasing, for both organs. The vortex grasping device has sufficient force to grasp and lift the organs, but due to the lack of friction elements in the design, the organs are sliding. Unlike soft balls (Fig.~\ref{fig-12}), organs have a very soft surface, which easily leads to the slipping. Therefore, in order to ensure the stable holding of the organs by the vortex gripper, we need to avoid slipping. For this, we chose a bio-inspired tissue for the experiment, which would behave similarly to the $ex$ $vivo$ experiment in which we see the vibration and sliding of the tissue.

Bio-inspired intestine with a diameter of 15 mm and a length of 120 mm was used to evaluate the effectiveness of vortex gripper grasping and lifting soft tissue. The gripping device was brought parallel to the bio-inspired intestine at a distance of 5 mm (Fig.~\ref{fig-14}a), after which a pressure of 200 kPa is supplied to the gripping cavity and the bio-inspired intestine is attracted to the gripper (Fig.~\ref {fig-14}b). Then the gripper moves up and the bio-inspired intestine lifts (Fig.~\ref{fig-14}c-h). During the lifting of the bio-inspired intestine, from the moment the intestine loses contact with the surface (Fig.~\ref{fig-14}c) on which it was lying, the intestine begins to rotate (Fig.~\ref{fig-14}d). The rotation of the intestines is due to the fact that the vortex air flows formed by the gripper easily rotate various objects during non-contact manipulation. Another reason for the rotation is the easy deformability of the bio-inspired intestine, which adapts to the air flow and prevents contact between the gripper and the intestine. Therefore, after the intestine loses contact with the surface on which it was lying, it begins to rotate until it slips off or releases from the gripper.

\begin{figure}[t]
\newcommand{\mywidth}{1}
\centering
         ~
 	\begin{subfigure}[b]{1\linewidth}
         \centering         \includegraphics[width=\mywidth\linewidth, clip, trim=0pt 40pt 0pt 100pt]{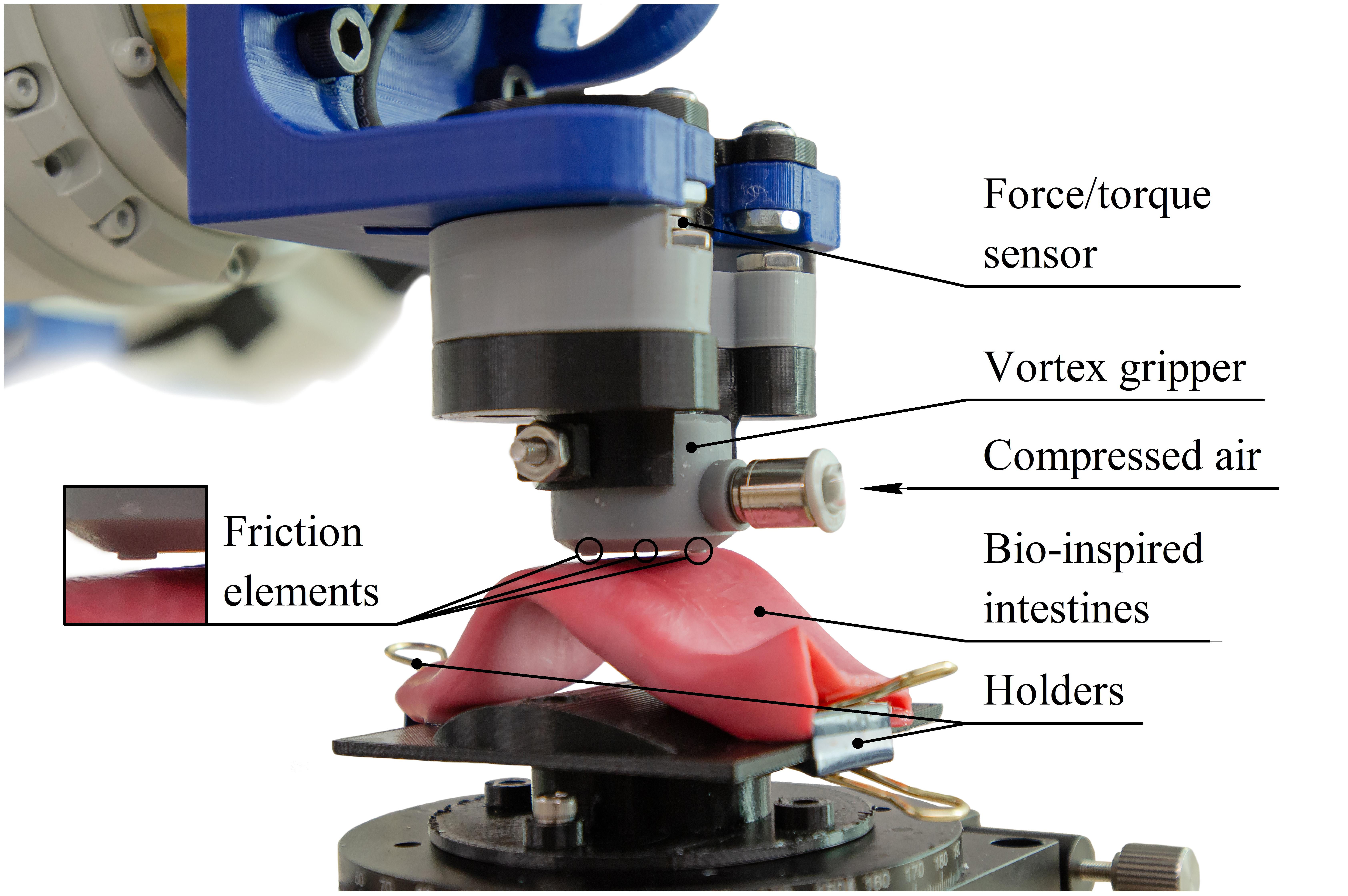}
         \caption{} \label{subfig12_2}
	\end{subfigure}
 	~ \\
 	\begin{subfigure}[b]{0.98\linewidth}
         \centering         \includegraphics[width=\mywidth\linewidth, clip, trim=35pt 35pt 45pt 35pt]{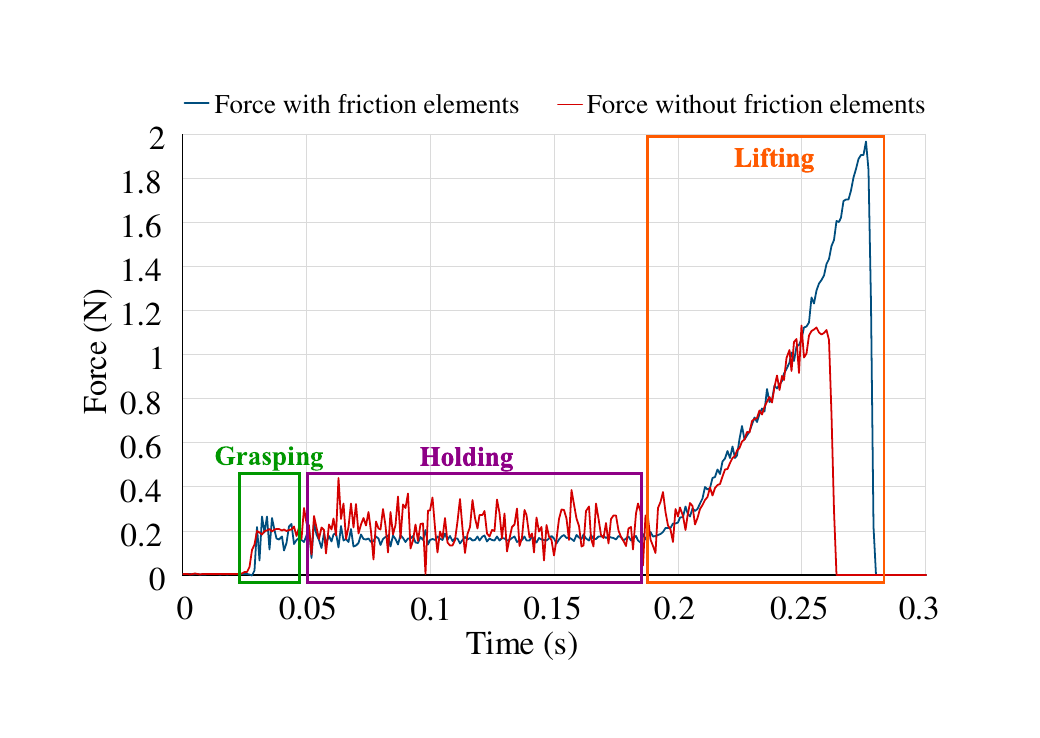}
         \caption{} \label{subfig12_3}
	\end{subfigure}
 \caption{Grasping, holding, and lifting of bio-inspired intestine fixed at two ends: (a) experimental setup and vortex griper with frictions elements (FE); (b) force parameters at a supply pressure of 300kPa.}
\label{fig-15}
\end{figure}

The classic and most stable method of using friction elements for jet gripping devices is the three-point contact of a cylindrical shape \cite{mykhailyshyn2021substantiation}. Therefore it was decided to add 3 friction elements (diameter 2 mm, 0.4 mm high, and 120 degrees between each) on the active surface of the gripper.  Also, conduct an experiment to determine the lifting force of the fixed bio-inspired intestine from two ends (Fig.~\ref{fig-15}a). For a comparative experiment, developed grippers with and without friction elements were used, which are brought to the intestine, grabbed, hold, and lifted vertically up the intestine until the contact is lost. The force parameters of the comparative experiment at a supply pressure of 300 kPa are presented in Figure~\ref{fig-15}b. As can be seen from Figure~\ref{fig-15}b, the force parameters for both grippers are practically identical during the grasping. However, the force parameters of the gripper without friction elements fluctuate during holding, which demonstrates the vibration of the intestine. Vibrations are caused by the fact that the intestine is deformed under the force of the gripper and is attracted to the gripper, but if the gap between the tissue and the gripper is too small, a significant increase in pressure occurs in this area and the tissue is deformed in the opposite direction, which creates tissue oscillations. At the same time, the vibration of the intestine can lead not only to the loss of the ability to hold the object but also to damage to soft tissues. There is practically no vibration when holding the intestine using a gripper with friction elements (Fig.~\ref{fig-15}b). This is determined by limiting the deformation of the soft tissue in the direction of the gripper with the help of friction elements. Another effect given by the presence of friction elements in the design of the gripper is to increase the lifting force due to the minimization of the sliding of the intestine during lifting. In the case of supply pressure of 300 kPa to the gripper chamber, the lifting force for the gripper with friction elements increases by 77\% from 1.1 to 1.95 N. 

\begin{figure}[t]
\newcommand{\mywidth}{0.90}
\centering
         \centering         \includegraphics[width=\mywidth\linewidth, clip, trim=0pt 0pt 30pt 0pt]{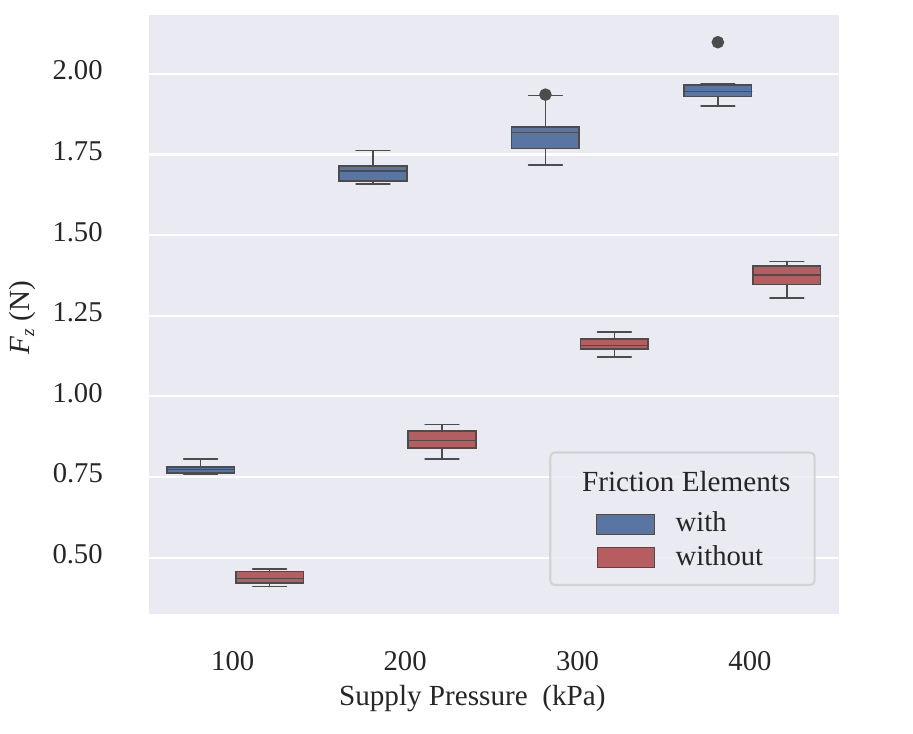}
 \caption{Distribution of force characteristics of vortex grippers when lifting bio-inspired intestine with and without friction elements at different supply pressures.}
\label{fig-16}
\end{figure}

For a more detailed assessment of the use of frictional elements in the design of the vortex gripper, an experimental study of the change in the lifting force during manipulation of bio-inspired intestine was conducted. The lifting force of the bio-inspired intestine for both grippers was determined by the method described above while changing the supply pressure in the range of 100-400 kPa with repeatability 10 times for each pressure (Fig.~\ref{fig-16}). From the obtained results, it is clear that the lifting force increases for both grippers with increasing supply pressure. However, due to the minimization of the vibration of the soft tissue and the avoidance of its slipping when using the design of the gripper with friction elements, it was possible to increase the lifting force by 68\% on average for all supply pressures. In addition, it is clearly visible from Fig.~\ref{fig-16} that for the gripper with friction elements, the lifting force does not increase significantly for supply pressures of 300 and 400 kPa. This is due to the fact that when the supply pressure increases, the optimal height of the friction elements changes, and since we made friction elements with a height of 0.4 mm, this limited the increase in the lifting force at higher pressures. It should also be noted that the cylindrical shape of the friction elements is not always optimal both from the point of view of the lifting force and the deformation properties of the grasping object. Therefore, optimizing the shape, position, and height of the friction elements can give a much better result both in terms of the force and the stability of holding soft tissues by the vortex gripper. What will be done in the following studies of the vortex gripper for medical applications.

%=================================================
\section{Conclusion}
In this paper, we have developed a vortex gripper design for grasping and manipulating soft tissues of various shapes. Thanks to the conducted experiments, we found out the strength characteristics of dome (convex), cylinder (convex), dome (concave), and cylinder (concave) surfaces. At a supply pressure of 400 kPa and dome (convex) surface, the vortex gripper can create a lifting force 35 times greater than its own weight. Even with a reduction of the lifting force for different radii of the cylindrical surface, the gripper provides the necessary force for lifting soft tissue such as bowls. The grasping of bio-inspired intestines using the developed vortex gripper was evaluated. The negative effects of vibration and sliding of the soft tissue during grasping and holding were revealed. These problems are solved due to the use of cylindrical friction elements in the gripper design that minimize vibration and sliding of soft tissue, which in turn allows stable manipulation of bio-inspired intestines and increases the lifting force by an average of 68\%. This study allows us to determine the limitations in the use of vortex technology and the next stages of its improvement for medical use. Experiments on grasping soft balls were conducted to demonstrate the improved capabilities of vortex grippers for grasping complex soft surfaces of various shapes. 

In future work, we aim to develop a soft vortex gripper designed for use in minimally invasive surgery. Additionally, we intend to determine the optimal position and shape of the friction elements within the vortex gripper design in upcoming studies.

% or


\section*{Acknowledgment}
The authors thank Yaskawa Motoman for access to a robotic system for this project.




%\bibliographystyle{IEEEtran}

%%%%%%%%%%%%%%%%%%%%%%%%%
%====================================
% References.
%====================================
%%%%%%%%%%%%%%%%%%%%%%%%%

\bibliographystyle{IEEEtran}
\IEEEtriggercmd{\enlargethispage{2in}}
%\IEEEtriggeratref{9}
%\IEEEtriggeratref{21}
\bibliography{Vortex}



\vfill



%\newpage

\begin{IEEEbiography}[{\includegraphics[width=1in,height=1.25in,clip,keepaspectratio]{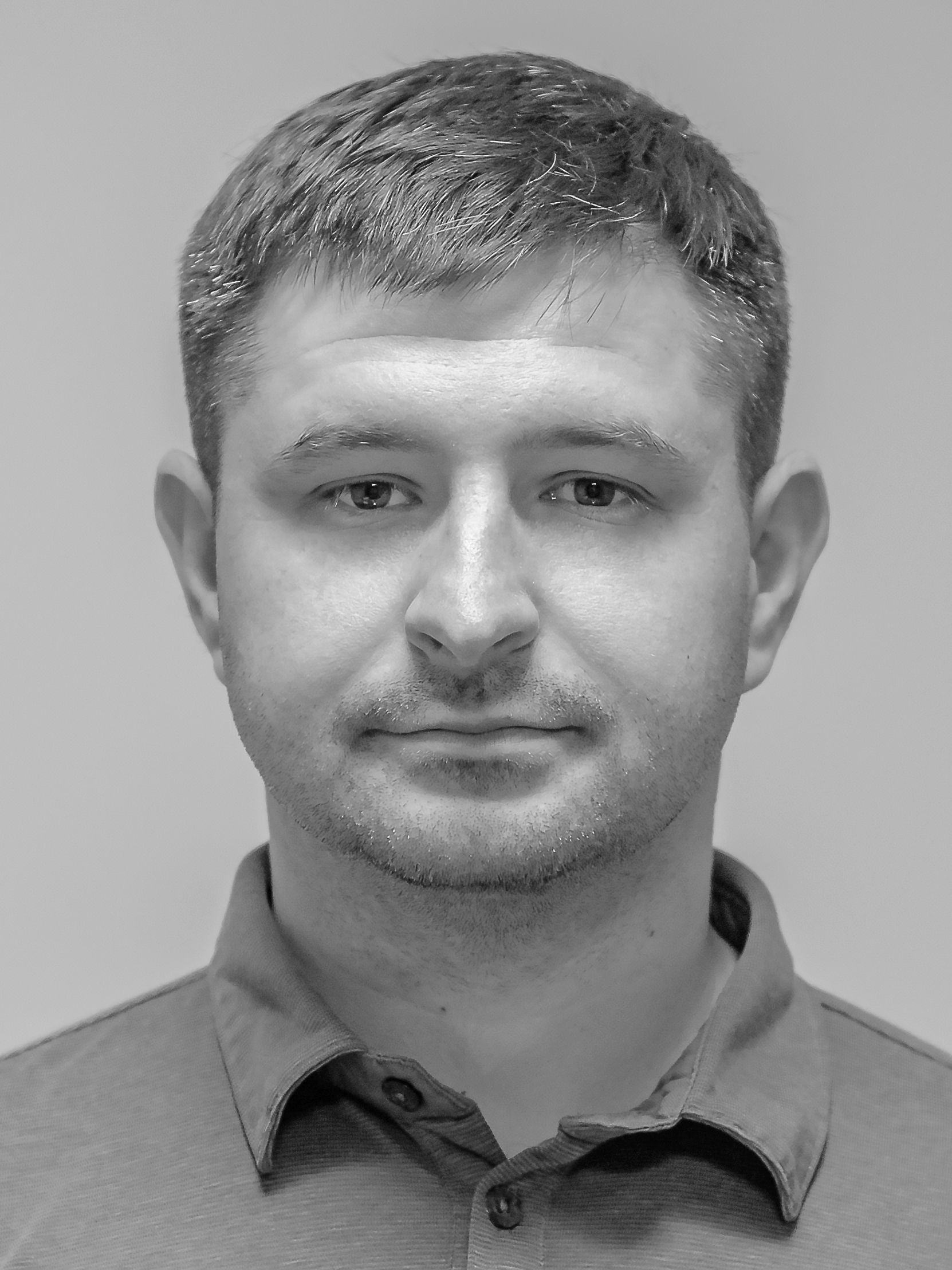}}]{Roman Mykhailyshyn}
(Member, IEEE) received the Ph.D. degree in Engineering Sciences from Ternopil Ivan Puluj National Technical University, Ternopil, Ukraine, in 2018.

He was an Associate Professor with the Department of Automation of Technological Processes and Manufacturing from 2019 to 2023. From 2021 to 2022, he was also a Fulbright Visiting Scholar with the Department of Robotics Engineering, Worcester Polytechnic Institute (WPI), USA.  From 2022 to 2024, he was a Research Fellow (Provost's Early Career Fellow) with the Texas Robotics and Walker Department of Mechanical Engineering, The University of Texas at Austin, USA. He is currently a Research Associate Professor at Osaka University with the transition to the National Institute of Advanced Industrial Science and Technology (AIST), Japan. His research interests include robotics, grasping, manipulation, haptics, gripper design and 3D printing.

Dr. Mykhailyshyn was the recipient 2020-2023 Fellowships of the Cabinet of Ministers of Ukraine for Young Scientists and the 2022 Machines Young Investigator Award.
\end{IEEEbiography}

\vspace{6pt}

\begin{IEEEbiography}[{\includegraphics[width=1in,height=1.25in,clip,keepaspectratio]{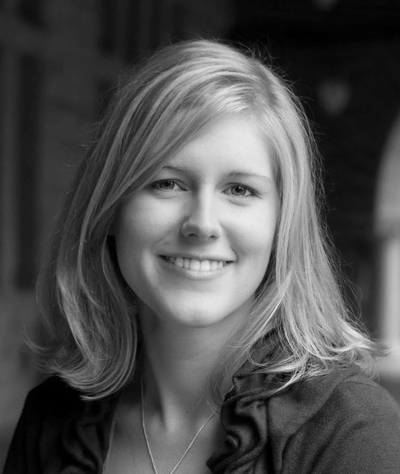}}]{Ann Majewicz Fey}
(Member, IEEE) received the B.S. degrees in Mechanical Engineering and Electrical Engineering from the University of St. Thomas, St. Paul, MN, in 2008, the M.S. degree in Mechanical Engineering from Johns Hopkins University, Baltimore, MD, in 2010, and the Ph.D. degree from Stanford University, Stanford, CA, in 2014, all in mechanical engineering.

She is currently an Associate Professor of mechanical engineering at the University of Texas at Austin, TX, where she holds a joint appointment in the Department of Surgery at UT Southwestern Medical Center, Dallas, TX. She directs the Human-Enabled Robotic Technology (HeRo) Laboratory where she is responsible for research projects in the areas of robot-assisted surgery, teleoperation, haptics, and human-centric modeling. 

Dr. Majewicz Fey received the 2015 National Science Foundation CISE Research Initiation Initiative (CRII) award and the 2019 National Science Foundation CAREER Award.
\end{IEEEbiography}

\end{document}